\documentclass[12pt,a4paper,twoside]{article}

\usepackage[rreport]{cmpcover}

\title{Two-view Matching with View Synthesis Revisited}
\author{Dmytro Mishkin, Michal Perdoch, Jiri Matas}

\CMPReportNo{CTU--CMP--2013--15} 

\CMPAcknowledgement{\centering The authors were supported by The Czech Science Foundation Project GACR
P103/12/G084 and by the Technology Agency of the Czech Republic
research program TE01020415 (V3C -- Visual Computing Competence
Center).}

\CMPEmail{ducha.aiki\{at\}gmail.com,perdom1,matas \{at\}(cmp.felk.cvut.cz)}  

\defRCSRevision{$Revision: 0.1 $}

\usepackage{times}
\usepackage{iccv}
\usepackage{epsfig}
\usepackage{graphicx}
\usepackage{amsmath}
\usepackage{amssymb}
\usepackage{gensymb}
\usepackage{rotating}
\usepackage{colortbl}
\usepackage{dcolumn}
\usepackage{algpseudocode}
\usepackage{algorithm}
\usepackage{tabu}
\usepackage{setspace}
\usepackage{appendix}
\usepackage{changepage}
\usepackage[margin=1.5cm]{geometry}

\definecolor{LightRed}{rgb}{1,0.9,0.9}

\newenvironment{itemize*}%
  {\begin{itemize}%
    \setlength{\itemsep}{1pt}%
    \setlength{\parskip}{1pt}}%
  {\end{itemize}}

\newenvironment{enumerate*}%
  {\begin{enumerate}%
    \setlength{\itemsep}{1pt}%
    \setlength{\parskip}{1pt}}%
  {\end{enumerate}}

\makeatletter
\renewcommand{\paragraph}{%
  \@startsection{paragraph}{4}%
  {\z@}{0.5ex \@plus 1ex \@minus .2ex}{-1em}%
  {\normalfont\normalsize\bfseries}%
}
\newcommand*{\@rowstyle}{}

\newcommand*{\rowstyle}[1]{
  \gdef\@rowstyle{#1}%
  \@rowstyle\ignorespaces%
}

\newcolumntype{=}{
  >{\gdef\@rowstyle{}}%
}

\newcolumntype{+}{
  >{\@rowstyle}%
}
\newcolumntype{B}[3]{>{\@rowstyle\DC@{#1}{#2}{#3}}c<{\DC@end}}

\newcolumntype{.}{B{.}{.}{1}}
\newcolumntype{,}{D{.}{.}{1}}
\makeatother

\newcommand{\footnoteremember}[2]{\footnote{#2} \newcounter{#1} \setcounter{#1}{\value{footnote}}}

\newcommand{\algrule}[1][.2pt]{\par\vskip.5\baselineskip\hrule height#1\par\vskip.5\baselineskip}

\newcommand{\sqbox}[1]{%
    \@begin@tempboxa\hbox{#1}%
    \@tempdima=\dimexpr\width-\totalheight\relax
    \ifdim\@tempdima<\z@
        \fbox{\hbox{\hspace{-.5\@tempdima}\usebox\@tempboxa\hspace{-.5\@tempdima}}}%
    \else
        \ht\@tempboxa=\dimexpr\ht\@tempboxa+.5\@tempdima\relax
        \dp\@tempboxa=\dimexpr\dp\@tempboxa+.5\@tempdima\relax
        \fbox{\usebox\@tempboxa}%
    \fi
    \@end@tempboxa
}

\begin{document}
\maketitle

\begin{abstract}
Wide-baseline matching focussing on problems with extreme
viewpoint change is considered. We introduce the use of view synthesis with
affine-covariant detectors to solve such problems and show that matching with
the Hessian-Affine or MSER detectors outperforms the state-of-the-art
ASIFT~\cite{Morel2009}.

To minimise the loss of speed caused by view synthesis, we propose the Matching
On Demand with view Synthesis algorithm (MODS) that uses progressively more
synthesized images and more (time-consuming) detectors until reliable estimation
of geometry is possible. We show experimentally that the MODS algorithm solves problems beyond the state-of-the-art and yet is comparable in speed to
standard wide-baseline matchers on simpler problems.

Minor contributions include an improved method for tentative correspondence
selection, applicable both with and without view synthesis and a view synthesis setup greatly improving MSER robustness to blur and scale
change that increase its running time by 10\% only.
\end{abstract}

\section{Introduction}
\label{sec:introduction}
The standard method for wide baseline matching involves detection of local
features, calculation of descriptors, generation of tentative
correspondences and their geometric verification using the homography or 
epipolar constraint.

It is well known~\cite{Mikolajczyk2005a,Fraundorfer2005,Dahl2011} that
performance of the pipeline decreases in the presence of viewpoint and scale
changes, blur, compression artefacts, etc. Lepetit and Fua~\cite{Lepetit2006}
showed that matching robustness is improved by synthesis of
additional views given a single, fronto-parallel view of an object.
Morel and Yu~\cite{Morel2009} combined viewpoint synthesis
with the similarity-covariant Difference-of-Gaussians detector (DoG) and
SIFT matching~\cite{Lowe2004}. The resulting image matching method, called
ASIFT, successfully matched challenging image pairs with significantly
different viewing angles.

We develop the idea of view synthesis for wide baseline matching and propose a
number of novelties that improve several stages of the matching pipeline. Some
of the improvements are also applicable to two-view matching without synthesis.
The proposed MODS wide-baseline matcher\footnote{Available at http://cmp.felk.cvut.cz/wbs/index.html}
outperforms ASIFT in terms of speed, the number and percentage of correct
matches generated as well as in the precision of the estimated geometry.
Performance was tested mainly on image pairs with extreme viewpoint
changes, but viewpoint synthesis also improves matching results in the presence
of phenomena like blur, occlusion and scale change.
\begin{figure}[tb]
\label{fig:hard-case}
\centering
\includegraphics[width=1\linewidth]{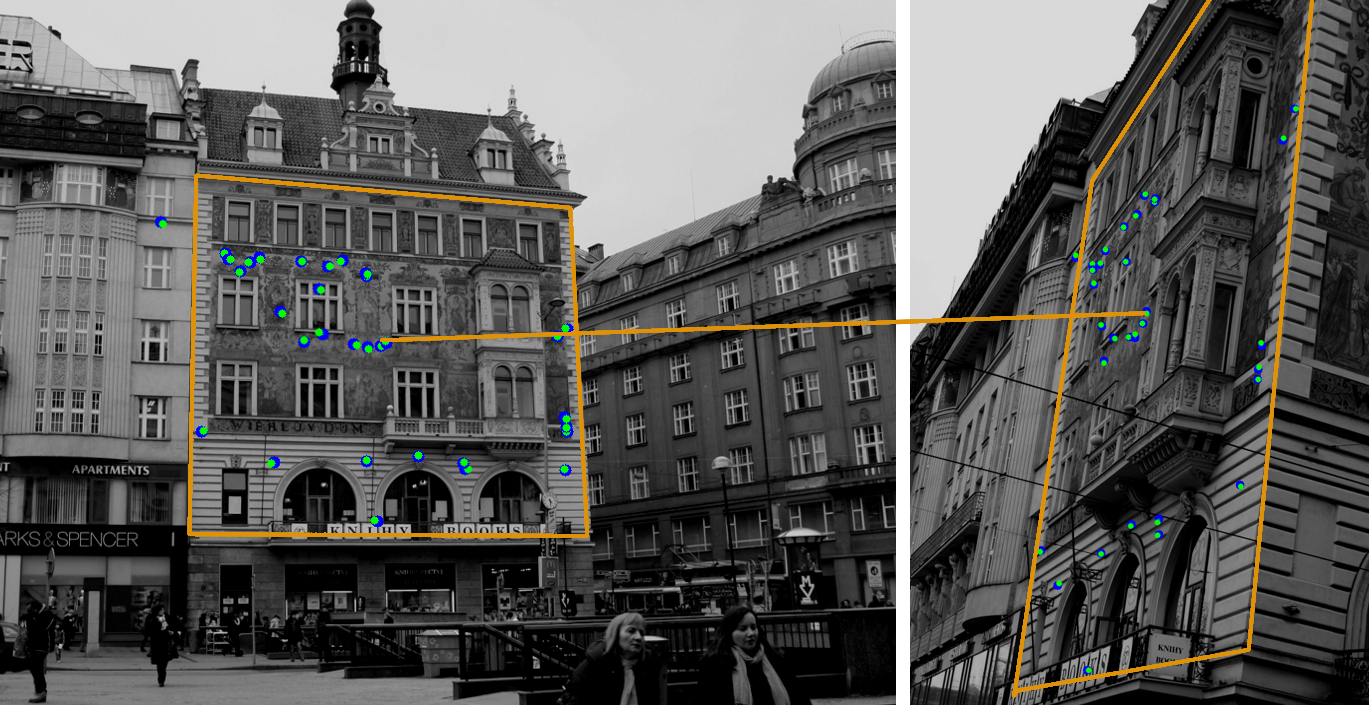}
\caption{Homography estimation with extreme viewpoint change. The proposed MODS
algorithm produces 32 matches, 25 are correct. The state-of-the-art
ASIFT~\cite{Morel2009} outputs 41 matches, 3 are correct. Blue
dots: centers of detected regions. Green dots: reprojected centers of
corresponding regions showing good alignment.} 
\end{figure}
The following contributions are made: first, we show that the seemingly
counter-intuitive synthesis of affine views for "affine-covariant" detectors
greatly improves their performance in wide baseline matching. With suitable
detector-specific configurations of synthesized viewpoints, found through
extensive experimentation, both the Hessian-Affine~\cite{Mikolajczyk2004} and
MSER~\cite{Matas2002} detectors clearly outperform DoG~\cite{Lowe2004}.

Second, we generalize the "first-to-second-closest SIFT distance ratio"
criterion for the selection of tentative correspondences. 
Depending on the image, the new criterion gives 5-20\% more true matches than
the standard at no extra computation cost. The proposed criterion
improves even matching performance without synthesis, especially in
images with local symmetries.

Third, we propose an adaptive algorithm for matching very challenging image
pairs which follows the "do only as much as needed" principle. The MODS
algorithm (Matching On Demand with view Synthesis) uses progressively more
detector types and more synthesized images until enough correspondences for reliable
estimation of two-view geometry are found. MODS is fast on easy
image pairs without compromising performance on the hardest~problems.
\subsection{Related work}
The use of view synthesis for image matching is
a recent development and the literature is limited and includes mainly
modifications of the ASIFT algorithm. Liu~\etal~\cite{Liu2012}
synthesised perspective warps rather than affine. Pang~\etal~\cite{Pang2012}
replaced SIFT by SURF~\cite{Bay2006} in the ASIFT algorithm to reduce the
computation time. Sadek~\etal~\cite{Sadek2012} present a new affine covariant
descriptor based on SIFT which can be used with or without view synthesis.
Detection of the MSERs on the scale space pyramid was proposed by Forss{\'e}n and Lowe~\cite{Forssen2007}. 

The rest of the paper is organised in a top-down manner. 
In Section~\ref{sec:method}, we introduce the adaptive MODS two-view matching
algorithm. Section~\ref{sec:parameters} studies view synthesis for affine-covariant detectors. 
Experiments are presented in Section~\ref{sec:main}. Full experimental data is in Appendix.
\vspace{-0.5em} 
\section{Matching with On Demand View Synthesis}
\label{sec:method}
The iterative MODS algorithm (see Alg.~\ref{alg:MODS}) repeats a sequences
of two-view matching procedures, until a required minimum number of geometrically
verified correspondences is found. In each iteration, a different
detector is used and a different set of views generated. The adopted
sequence is an outcome of extensive experimentation with the objective of
solving the most challenging problems while keeping speed comparable
to standard single-detector wide-baseline
matchers for simple problems. For instance, the first iteration of the MODS
algorithm runs the MSER detector with only a very coarse scale space
pyramid which is 10\% slower than standard MSER. Subsequent iterations
run complementary detectors with a higher number of synthesized views.
Details on the chose configuration and the selection process are given in
Section~\ref{sec:parameters}.
The rest of the section describes the steps employed in the iterations of
the MODS algorithm. 
\renewcommand\algorithmicindent{1em}
\begin{spacing}{0.7}
\setlength{\textfloatsep}{0pt}
\begin{algorithm}[htb]
\small
\caption{{\bf MODS}: Matching with On-Demand view Synthesis}
\label{alg:MODS}
\begin{algorithmic}
\Require{$I_1$, $I_2$ -- two images; $\theta_m$ -- minimum required number of matches; $S_{\text{max}}$ -- maximum number of iterations.  }
\Ensure{Fundamental of homography matrix F or H; \\list of corresponding points.}\\
\hskip-1em\textbf{Variables}:$N_{\text{matches}}$ -- detected correspondences, $\text{Iter}$ -- currect iteration.
\algrule
\vspace{-0.3em} 
\While {$(N_{\text{matches}} < \theta_{m})$ \textbf{and} $(\text{Iter} < S_{\text{max}})$} 
\For {$I_1$ and $I_2$ separately}
\State \textbf{1 Generate synthetic views} according to the\\
\hskip3em scale-tilt-rotation-detector setup for the Iter.
\State \textbf{2 Detect and describe local features}.
\State \textbf{3 Reproject local features} to original image. \\ 
\hskip3em Add described features to general list.
\EndFor
\State \textbf{4 Generate tentative correspondences} \\
\hskip2em using the first geom. inconsistent rule. 
\State \textbf{5 Filter duplicate matches}. 
\State \textbf{6 Geometrically verify tentative correspondences} \\
\hskip2em while estimating F or H.
\EndWhile
\end{algorithmic}
\end{algorithm}
\vspace{-1.7em} 
\end{spacing}
\subsection{Synthetic views generation}
It is well known that a homography $H$ can be approximated by
an affine transformation $A$ at a point using the first order Taylor expansion.
Further, an affine transformation can be uniquely decomposed by SVD into 
a rotation, skew, scale and rotation around the optical axis~\cite{Hartley2004}.
Morel and Yu~\cite{Morel2009} proposed to decompose the affine transformation
$A$ as 
%
\begin{equation}
\begin{array}{rcl}
\label{eq:affine-matrix-decomposition}
\footnotesize
A&=&H_\lambda{}R_1(\psi)T_t R_2 (\phi) = \\[3pt]
&=&\lambda{}%
\left(
\footnotesize
\begin{array}{@{\hspace{0pt}}c@{\hspace{6pt}}c@{\hspace{0pt}}}
\cos{\psi} & -\sin{\psi} \\
\sin{\psi} & \cos{\psi}\\
\end{array}\right)\!\left(%
\footnotesize
\begin{array}{
@{\hspace{0pt}}c@{\hspace{6pt}}c@{\hspace{0pt}}}
t & 0  \\
0 & 1 \\        
\end{array}\right)\!\left(%
\footnotesize
\begin{array}{
@{\hspace{0pt}}c@{\hspace{6pt}}c@{\hspace{0pt}}}
\cos{\phi} & -\sin{\phi}\\
\sin{\phi} & \cos{\phi}\\
\end{array}\right)
\end{array}
\end{equation}
where $\lambda{} > 0$, $R_1$ and $R_2$ are rotations, and $T_t$ is a
diagonal matrix with $t > 1$. Parameter $t$ is called the absolute tilt,
$\phi \in \langle 0,\pi)$ is the optical axis longitude and $\psi \in \langle
0 ,2\pi)$ is the rotation of the camera around the optical axis. Each synthesised
view is parametrised by the tilt, longitude and optionally the scale and represents
a sample of the view-sphere resp. view-volume around the original image.

The view synthesis proceeds in the following steps: at first, scale synthesis is
performed by building a Gaussian scale-space with Gaussian
$\sigma={\sigma_{\text{base}}}\cdot{S}$
and downsampling factor $S$ ($S<1$).
Second, each image in the scale-space is in-plane rotated by longitude $\phi$
with step $\Delta\phi={\Delta\phi_{\text{base}}}/{t}$. 
In the third step,
all rotated images are convolved with a Gaussian filter with $\sigma =
\sigma_{\text{base}}$ along vertical direction and $\sigma =
t\cdot{}\sigma_{\text{base}}$ along horizontal direction to eliminate aliasing
in the final tilting step. The tilt is applied by shrinking the image along the
horizontal direction by factor $t$. The parameters of the synthesis are: the set of
scales \{S\}, $\Delta\phi_{\text{base}}$ -- the step of longitude samples at tilt $t=1$,
and a set of simulated tilts \{t\}.
\vspace{-0.5em}
\subsection{Local feature detection and description}
The goal of the view synthesis procedure is to provide detectors with a sufficiently similar subset of all artificial views on the view-sphere that allows matching. For affine-covariant detectors, unlike the similarity-covariant
DoG of ASIFT, the number of necessary view samples is significantly decreased
while the performance for the most difficult image pairs gets improved.
Moreover, it is known that different detectors are suitable for different types
of images~\cite{Mikolajczyk2005a} and  that some detectors are complementary in
the feature points they detect~\cite{Aanaes2012}. Our experiments show (\cf
Section~\ref{sec:main}) that combining detectors improves the overall robustness
and speed of the matching procedure.

MODS uses the state-of-the-art affine covariant detectors MSER and
Hessian-Affine. 
The normalised patches are described by the recent modification of SIFT~\cite{Lowe2004} -- the RootSIFT~\cite{Arandjelovic2012}.
The local feature frames computed on the synthesised views are backprojected to
the coordinate system of the original image by a known affine matrix $A$
and associated with the descriptor and the originating synthetic view.
\vspace{-0.5em}
\subsection{Tentative correspondence generation}
\begin{figure}[tb]
\centering
\includegraphics[width=1\linewidth]{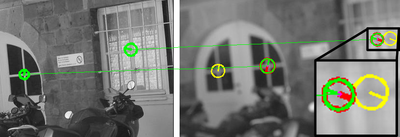}\\
\vspace{1 mm}
\includegraphics[width=1\linewidth]{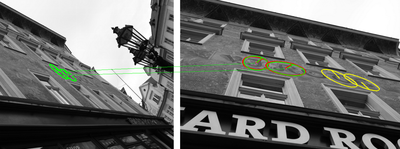}
\caption{Comparison of the proposed \emph{first to 1st inc. ratio} matching strategy and the standard \emph{first to second closest ratio} matching strategy.  Red regions are the second closest descriptors, yellow regions correspond to the closest geometrically inconsistent descriptors, green are the true corresponding regions. Upper pair -- rotationally symmetric DoG regions, lower pair -- affine covariant MSER regions.}
\vspace{-1.5em}
\label{fig:1st-inconsistent-illustration}
\end{figure}

Different strategies for computation of the tentative
correspondences in wide-baseline matching have been proposed.
The standard method for matching SIFT(-like) descriptors is based on the
distance ratio of the closest to the second closest descriptors in the
other image~\cite{Lowe2004}. Performance of this test in general very efficient
method degrades when multiple observations of the same feature are present. 
In this case, the similar descriptors will lead to the first to second SIFT ratio to be close to 1 and the correspondences will "annihilate" each other, despite the fact they all represent the same geometric constraints and are therefore not mutually contradictory (see Figure~\ref{fig:1st-inconsistent-illustration}). The problem of multiple detections is amplified in the matching by view synthesis since covariantly
detected local features have often a response in multiple synthetic views.
We propose to use, instead of comparing the first to the second closest
descriptor distance, the distance of the first descriptor and the
closest descriptor that is geometrically inconsistent with the first one (denoted
1st inc. in the following). We call descriptors in one image \emph{geometrically inconsistent} if the Euclidean distance between centers of the regions is $\geq$ 10 pixels. The difference of the first-to-second closest ratio
strategy and the closest-to-1st inc. strategy is illustrated
in~Figure~\ref{fig:1st-inconsistent-illustration}.

The kd-tree algorithm from FLANN library~\cite{Muja2009} effectively finds the
N-closest descriptors in the other image. The distance ratio thresholds of the
closest to 1st inc. were experimentally selected based on the CDFs of matching
and non-matching descriptors (see Appendix~\ref{sec:appendixA}). We recommend to use the same
values for SIFT and RootSIFT descriptors, but different thresholds for the
different local feature detectors: $R_{\text{MSER}} = 0.85$, $R_{\text{DoG}} =
0.85$ and $R_{\text{HA}} = 0.8$.
\subsection{Duplicate filtering} 
The redetection of covariant features in synthetic views results in duplicates
in tentative correspondences. The duplicate filtering is an optional step and
prunes correspondences with close spatial distance of local features in both
images. The number of pruned correspondences can be however used later for
evaluating the quality (probability of being correct) in PROSAC-like~\cite{Chum2005PROSAC} geometric verification.
\vspace{-0.5em} 
\subsection{Geometric verification}
The LO-RANSAC~\cite{Chum2003} algorithm searches for the maximal
set of geometrically consistent tentative correspondences. The model of the
transformation is set either to homography or epipolar geometry, or
automatically determined by a DegenSAC~\cite{Chum2005DEGENSAC} procedure.
\section{View synthesis for affine covariant detectors}
\label{sec:parameters}
%
\paragraph{Configurations.}
\begin{figure*}[htbp]
\includegraphics[width=0.34\linewidth]{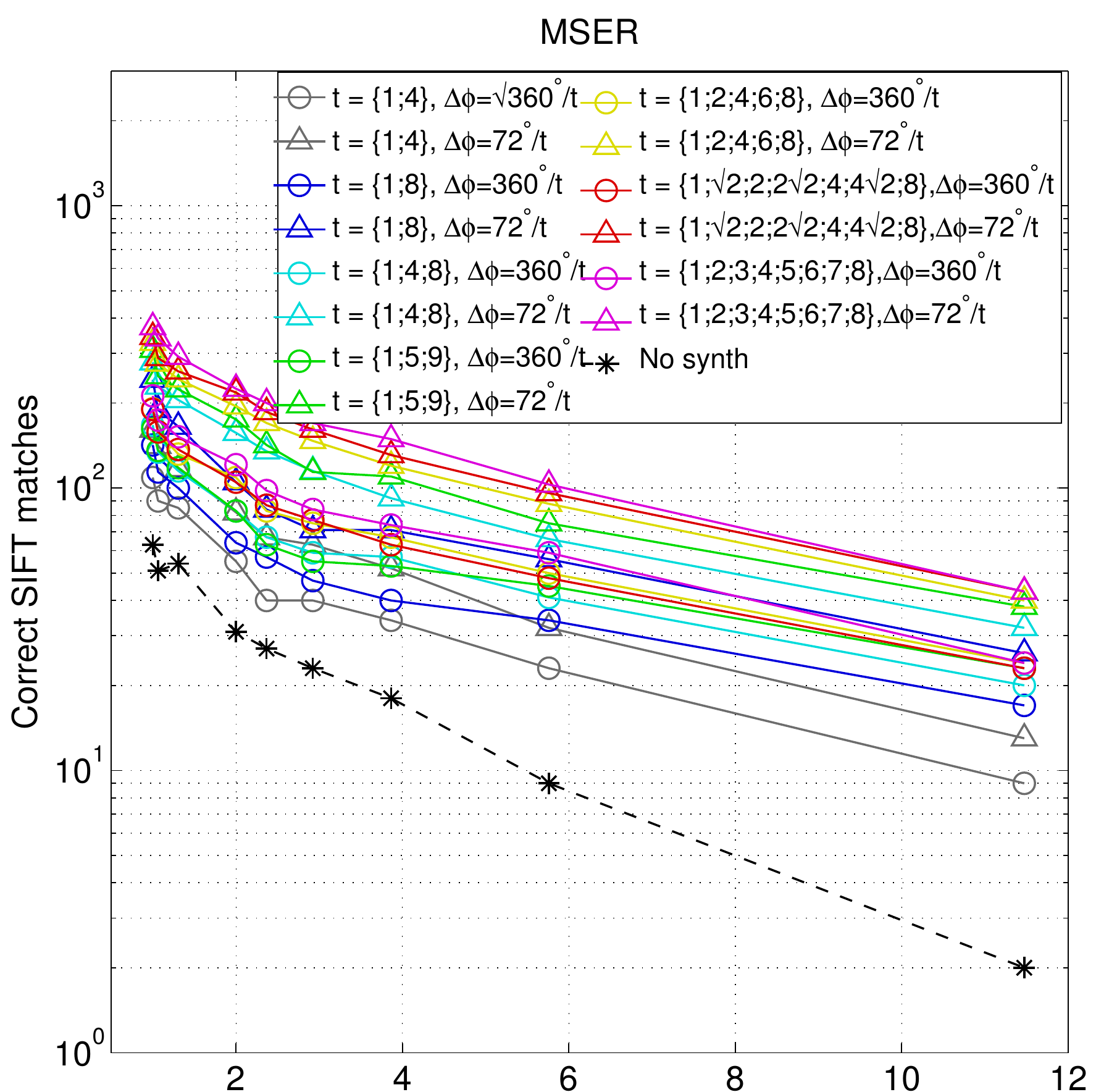}
\includegraphics[width=0.32\linewidth]{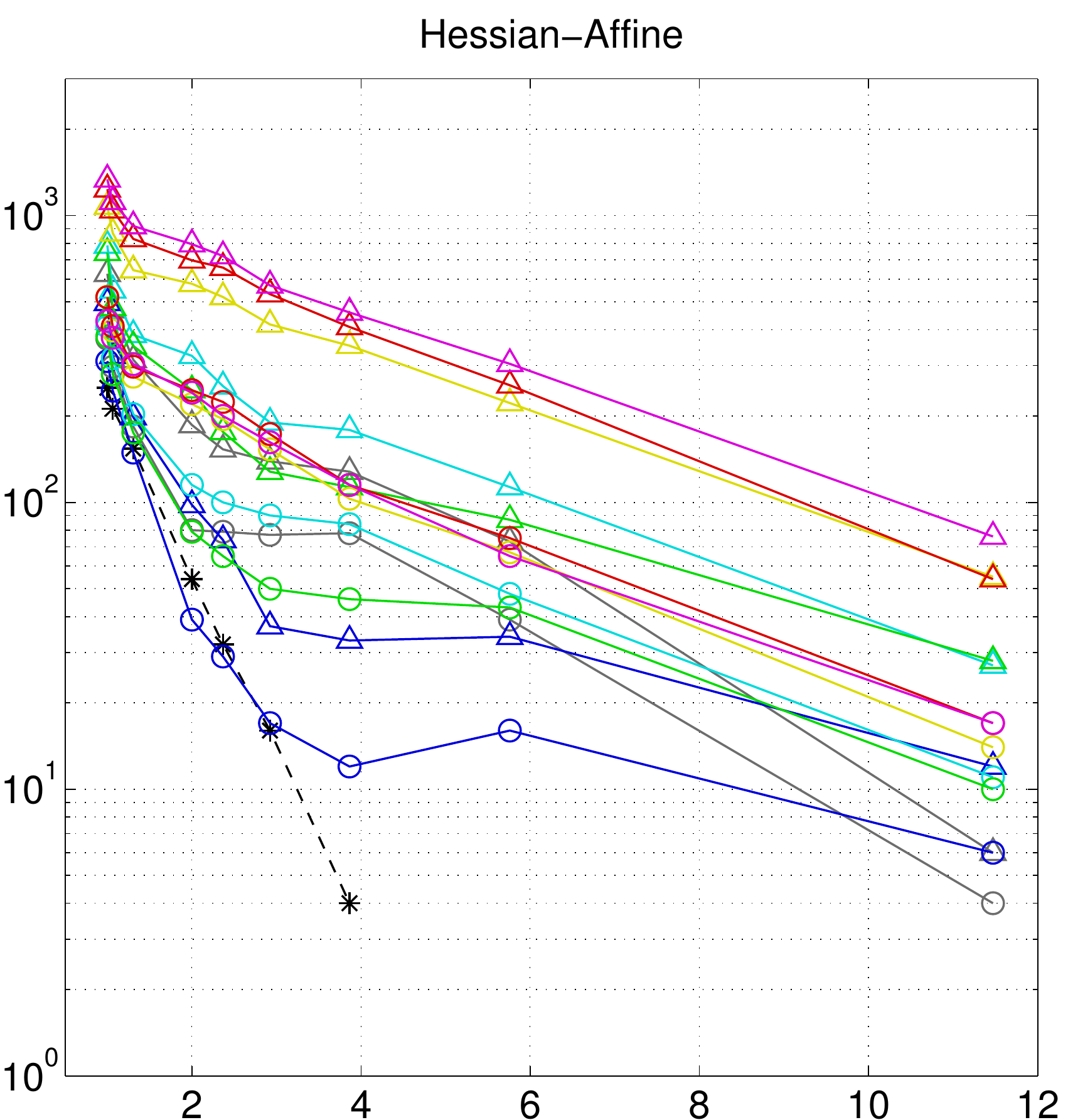}
\includegraphics[width=0.32\linewidth]{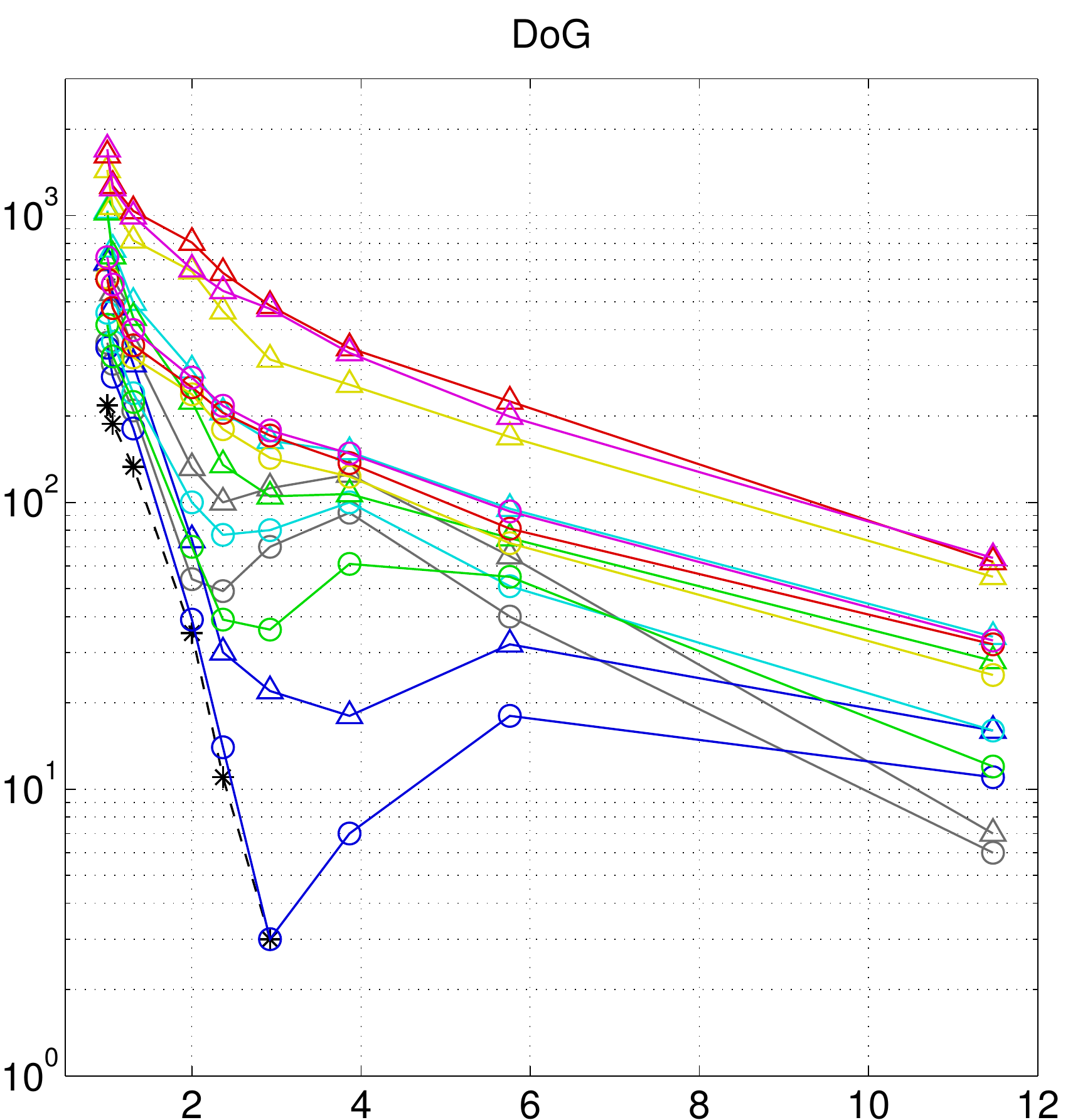}\\
\includegraphics[width=0.34\linewidth]{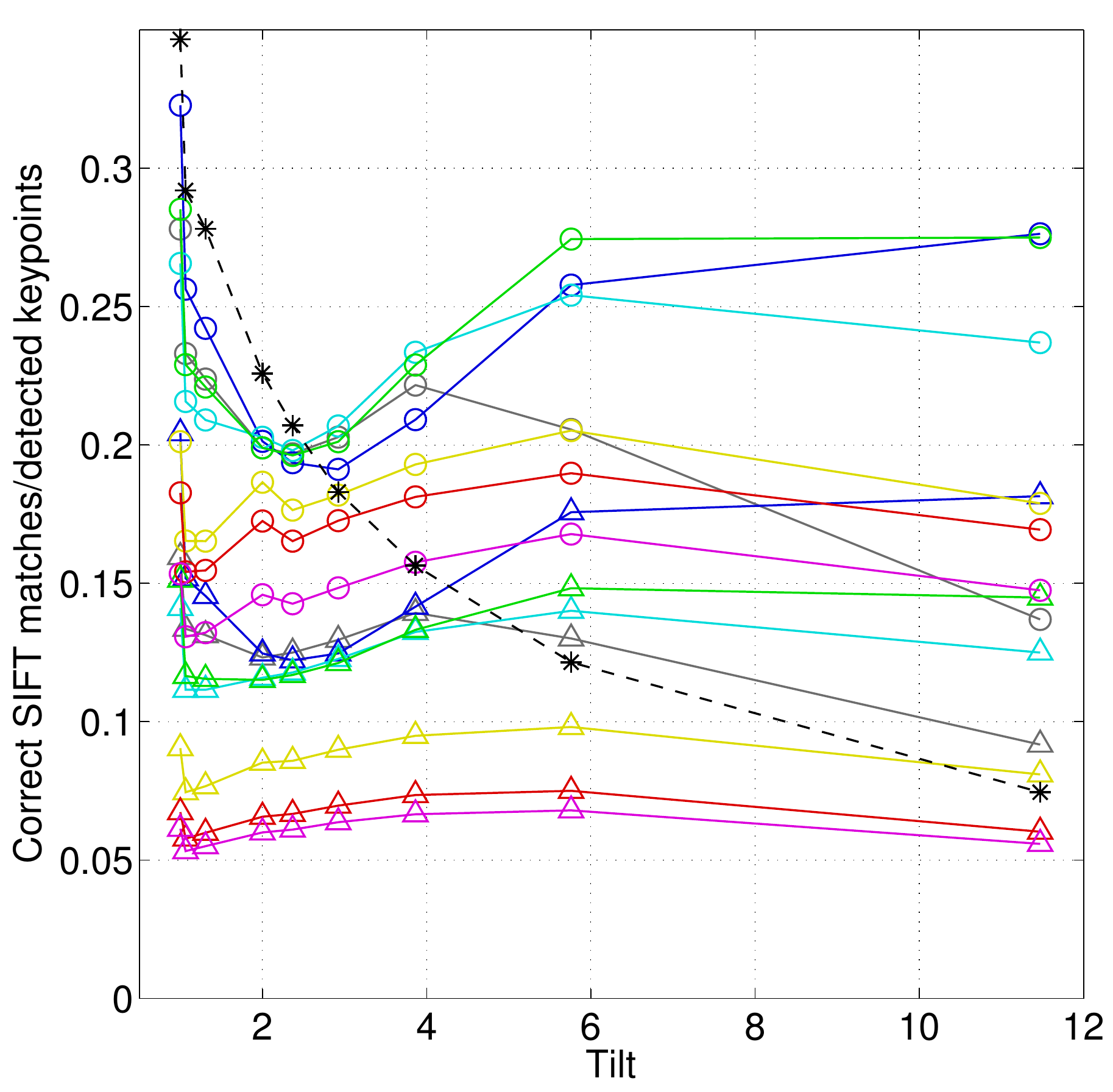}
\includegraphics[width=0.32\linewidth]{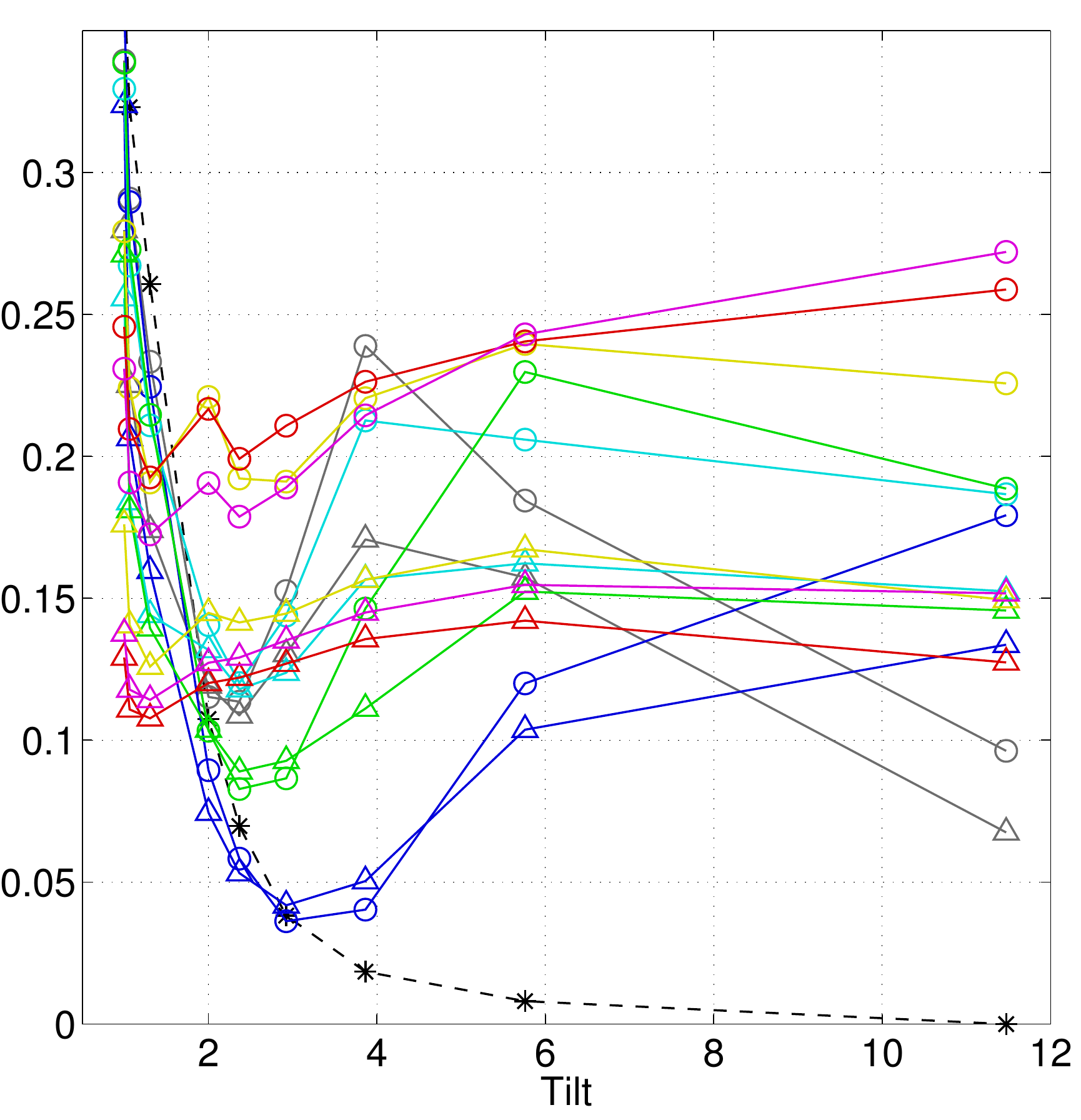}
\includegraphics[width=0.32\linewidth]{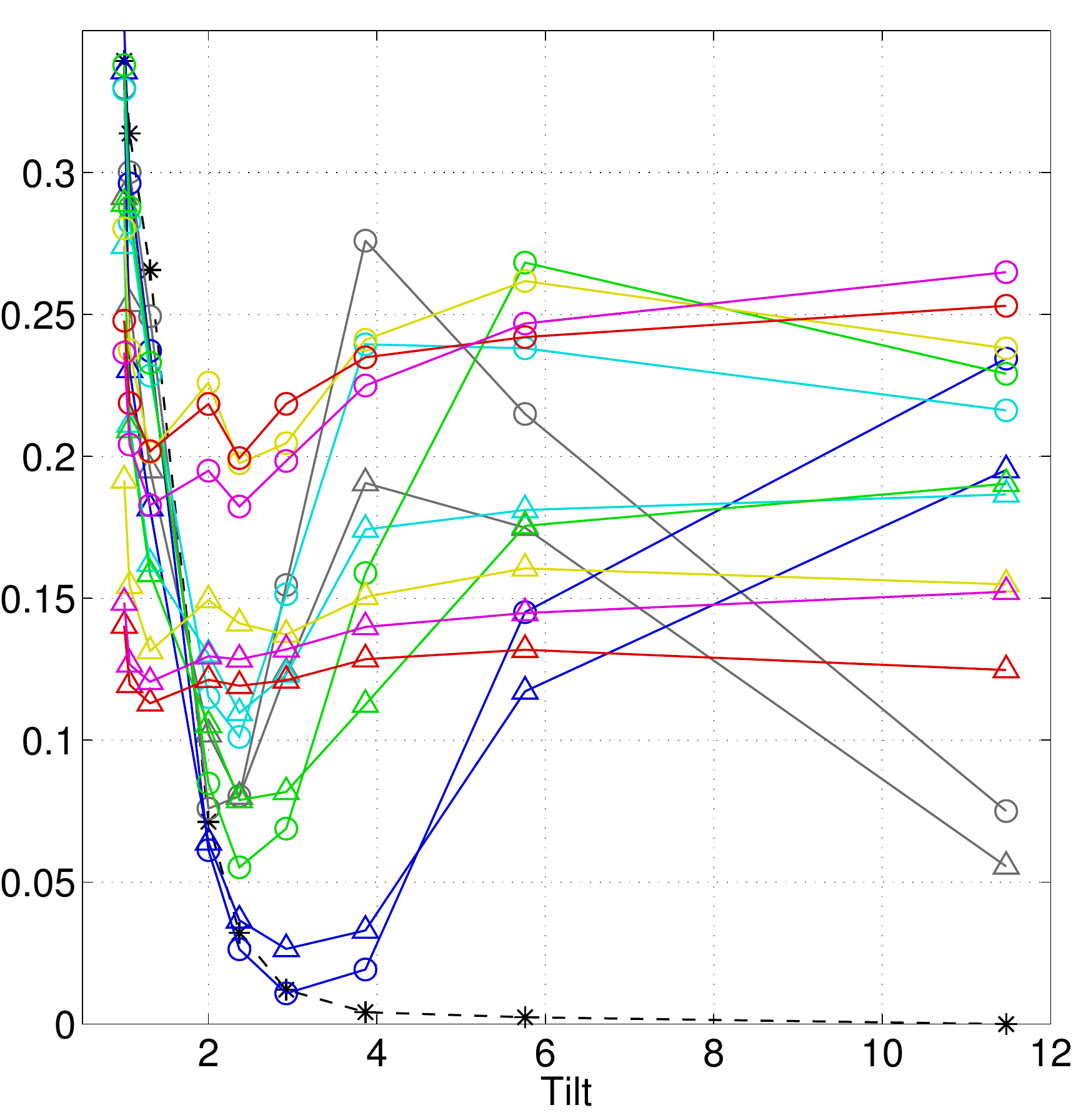}
\caption{Comparison of view synthesis configurations on the synthetic dataset.
First row: the number of correct SIFT matches a robust minimum (value 4\%
quantile) over 100 random images from~\cite{Philbin2007}). Second
row: the ratio of the number of correct matches to the number of detected
regions; the mean over 100 random images. Only selected configurations are
shown,  full version in Appendix.}
\label{fig:tilt-rotation}
\end{figure*} 
\begin{figure*}[htpb]
\includegraphics[width=0.34\linewidth]{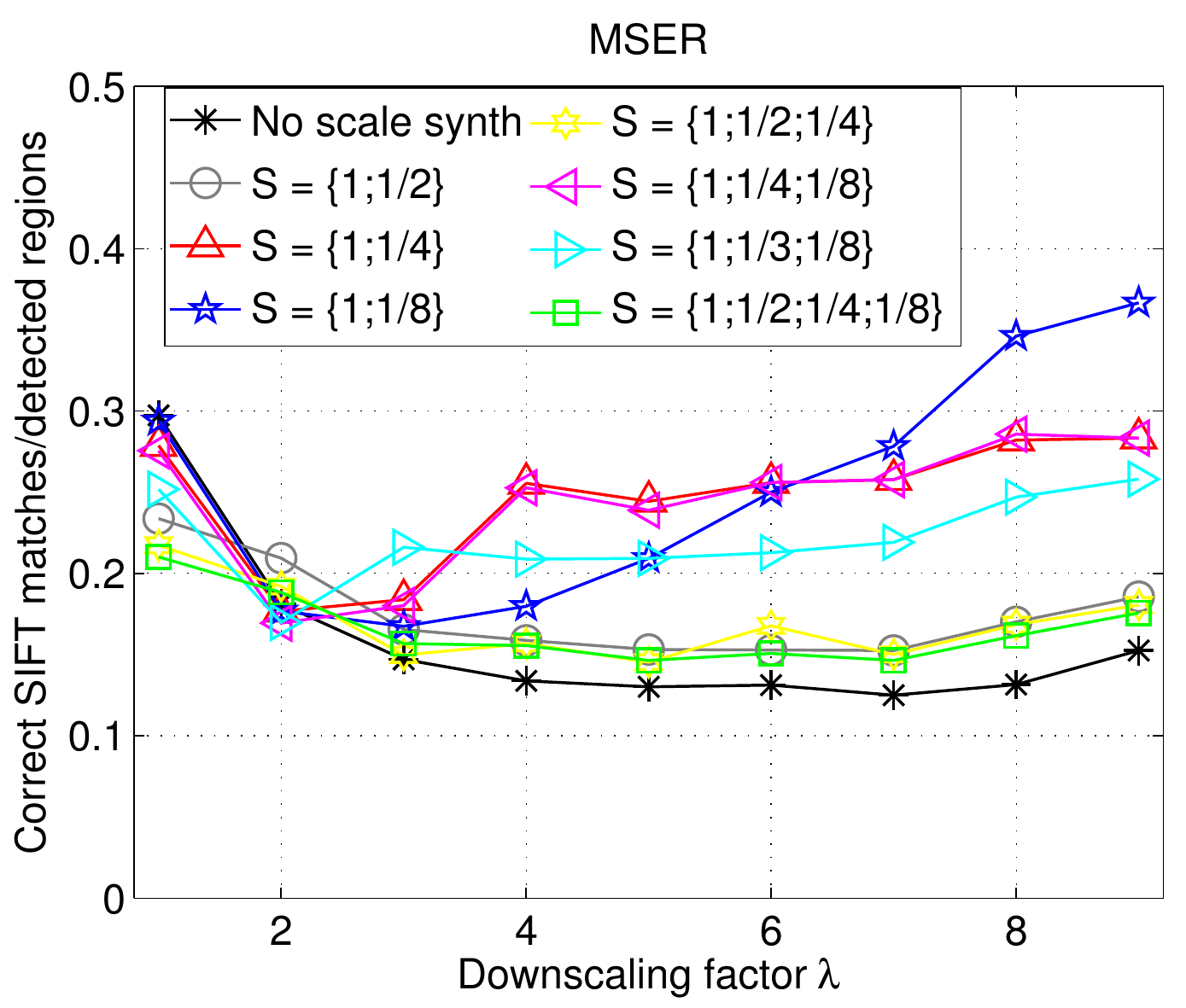}
\includegraphics[width=0.32\linewidth]{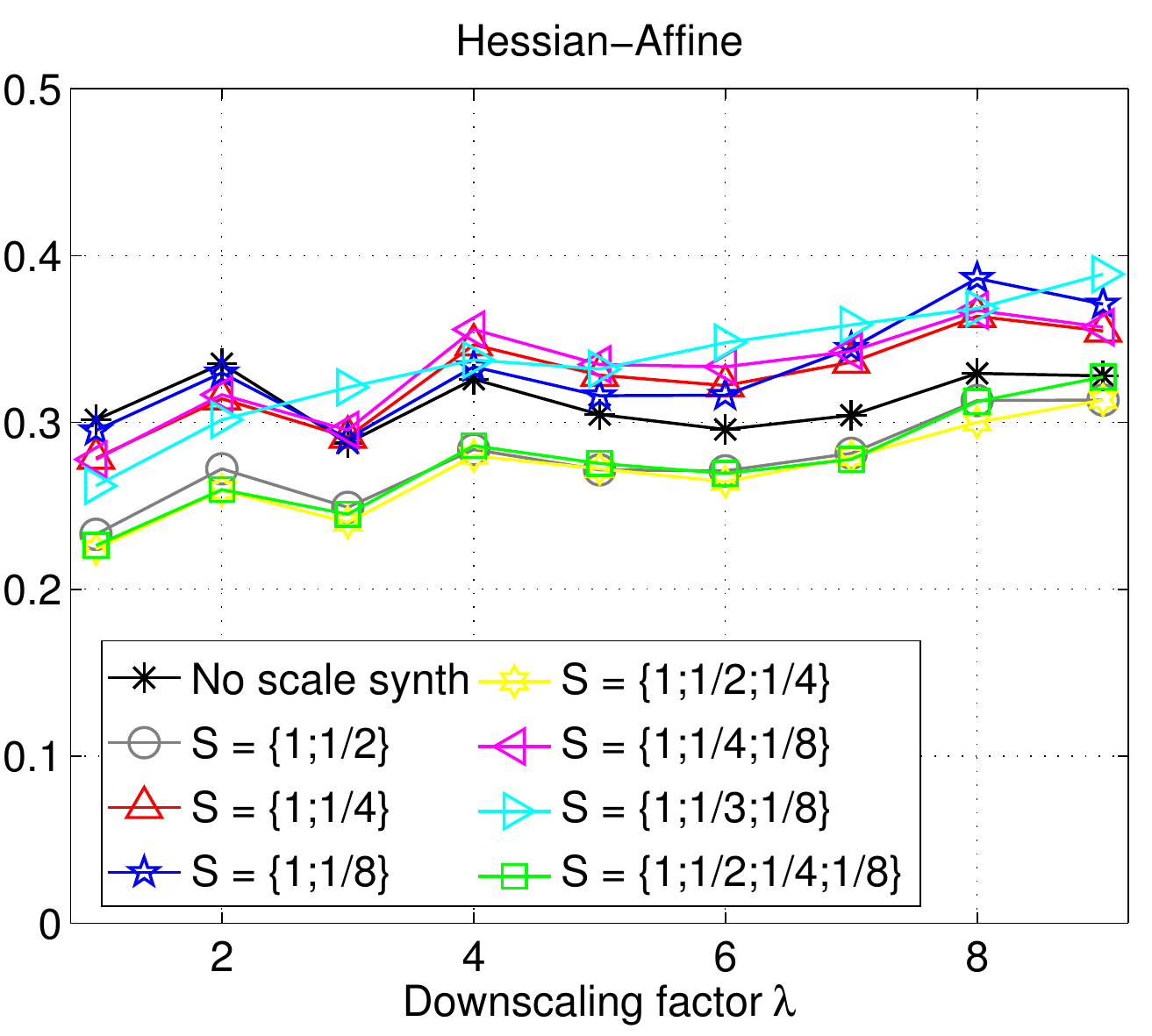} 
\includegraphics[width=0.32\linewidth]{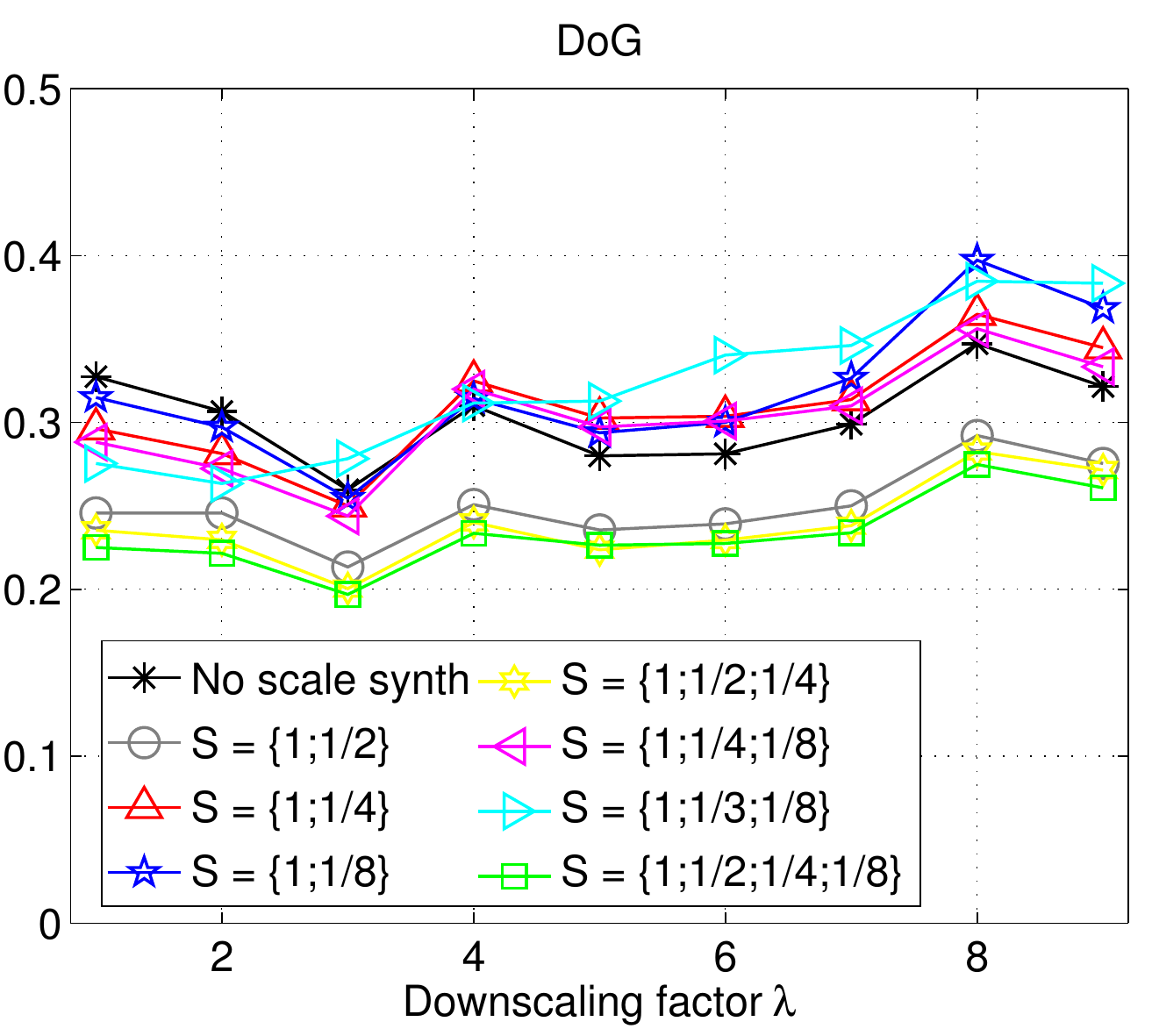} 
\caption{Estimation of the suitable scale synthesis configurations on the synthetic dataset. Ratio of the number of correct matches to the number of detected regions, mean over 100 random images from~\cite{Philbin2007}.}
\label{fig:scale-repeatability}
\end{figure*} 

The first two parameters of the view synthesis, tilt \{t\} sampling and latitude
step $\Delta\phi_{\text{base}}$, were explored in the following synthetic
experiment.
For each of 100 random images from  Oxford Building Dataset\footnote{available at http://www.robots.ox.ac.uk/$\sim$vgg/data/oxbuildings/}~\cite{Philbin2007},
a set of simulated views with latitudes angles $\theta=($0, 20, 40, 60, 65,
70, 75, 80, 85$) \degree$, corresponding to tilt series $t=($1.00, 1.06,
1.30, 2.00, 2.36, 2.92, 3.86,
5.75, 11.47$)$\footnote{assuming that the original image is in the
fronto-parallel view} was generated. The ground truth affine matrix $A$ was
computed for each synthetic view using
equation~(\ref{eq:affine-matrix-decomposition}) and used in the final
verification step of the MODS algorithm. 
The various configurations of the view synthesis were tested and results for the selected configurations are shown in Figure~\ref{fig:tilt-rotation}. Note that the view synthesis significantly increases the matching performance, however after reaching some density of the view-sphere sampling additional views does not bring more correspondences. MSER and Hessian-Affine need sparser view-sphere sampling than DoG. 

A similar experiment was performed to find the scale sampling set \{S\} of the
view synthesis. Instead of tilting and rotating the images, a synthetic
downsampling of the image by a factor $\lambda=1$ to 9 was
employed~(see~Figure~\ref{fig:scale-repeatability}).
It shows that MSER detector is prone to scale changes while the Hessian-Affine
and DoG detectors perform well even without view synthesis with scale
sampling. Consequently, the benefit of the scale sampling is higher for MSER
than for Hessian-Affine and DoG detectors. Tilting and rotation parameters
were not used in this experiment i.e. fixed to \{t\}~=~\{1\} and
$\Delta\phi_{\text{base}}=180$.

Two configurations, {\sc Sparse} and {\sc Dense}, were chosen for each detector
(see Table~\ref{tabular:view-synth-configuration}) using the following criteria:
efficiency -- the ratio of correct matches per detected region,  matching
performance -- the number of unique (non-duplicated) matches on the synthetic
image pairs with $85\degree$ out of plane rotation. The {\sc Sparse}
configuration is fast but still able to solve synthetic image pairs with
up to $85\degree$ out of plane rotation. The {\sc Dense} configuration generates sufficient number of correspondences for the most image pairs in the EVD dataset for each detector.

\begin{table}[tb] \caption{View synthesis configurations based on the
analysis of the algorithm on the synthetic dataset}
\label{tabular:view-synth-configuration} 
\footnotesize 
\centering 
\bgroup
\tabulinesep=0.1mm
\begin{tabu}spread 0pt{*{3}{|X[-1m,l]}|}
\hline
 & \multicolumn{2}{c|}{Configurations}\\
\hline
Detector&\centering {\sc Sparse} & \centering{\sc Dense}\\
\hline
MSER &  $\{S\} = \{1;0.25;0.125\}$, $\{t\} = \{1;5;9\}$, $\Delta\phi=360\degree{}/t$ &  $\{S\} = \{1;0.25;0.125\}$, $\{t\} = \{1;2;4;6;8\}$, $\Delta\phi=72\degree{}/t$ \\
\hline
HessAff & $\{S\} = \{1\}$, $\{t\} = \{1;\sqrt{2};$ $2;2\sqrt{2};4;4\sqrt{2};8\}$, $\Delta\phi=360\degree{}/t$& $\{S\} = \{1\}$, $\{t\} = \{1;2;4;6;8\}$, $\Delta\phi=72\degree{}/t$ \\
\hline
DoG & $\{S\} = \{1\}$, $\{t\} = \{1;2;4;6;8\}$, $\Delta\phi=120\degree{}/t$ & $\{S\} = \{1\}$, $\{t\} = \{1;\sqrt{2};$ $2;2\sqrt{2};4;4\sqrt{2};8\}$, $\Delta\phi=72\degree{}/t$ \\
\hline
\end{tabu}
\egroup
\end{table}
\paragraph{Image pre-smoothing.} Parameter $\sigma_{\text{base}}$, the amount of
image smoothing prior to view synthesis was set experimentally; it affects
matching performance significantly. Values too small fail to prevent aliasing,
values too high oversmooth the image reducing the number of detected regions.
Unlike MSER, the scale-space based detectors like DoG, Hessian-Affine apply
pre-smoothing as an initial step. This leads to different optimal values for
different detectors. We set $\sigma_{\text{base}}=0.8, 0.2$, and 0.4 for the
MSER, Hessian-Affine and DoG detectors, respectively.

\vspace{-1em}
\section{Experiments}
\label{sec:main}
\subsection{1st geometrically inconsistent vs. 2nd nearest neighbour
            correspondence selection strategy}
\label{main:1st-geom-criterion}
\begin{figure}[tb]
\includegraphics[width=0.51\linewidth]{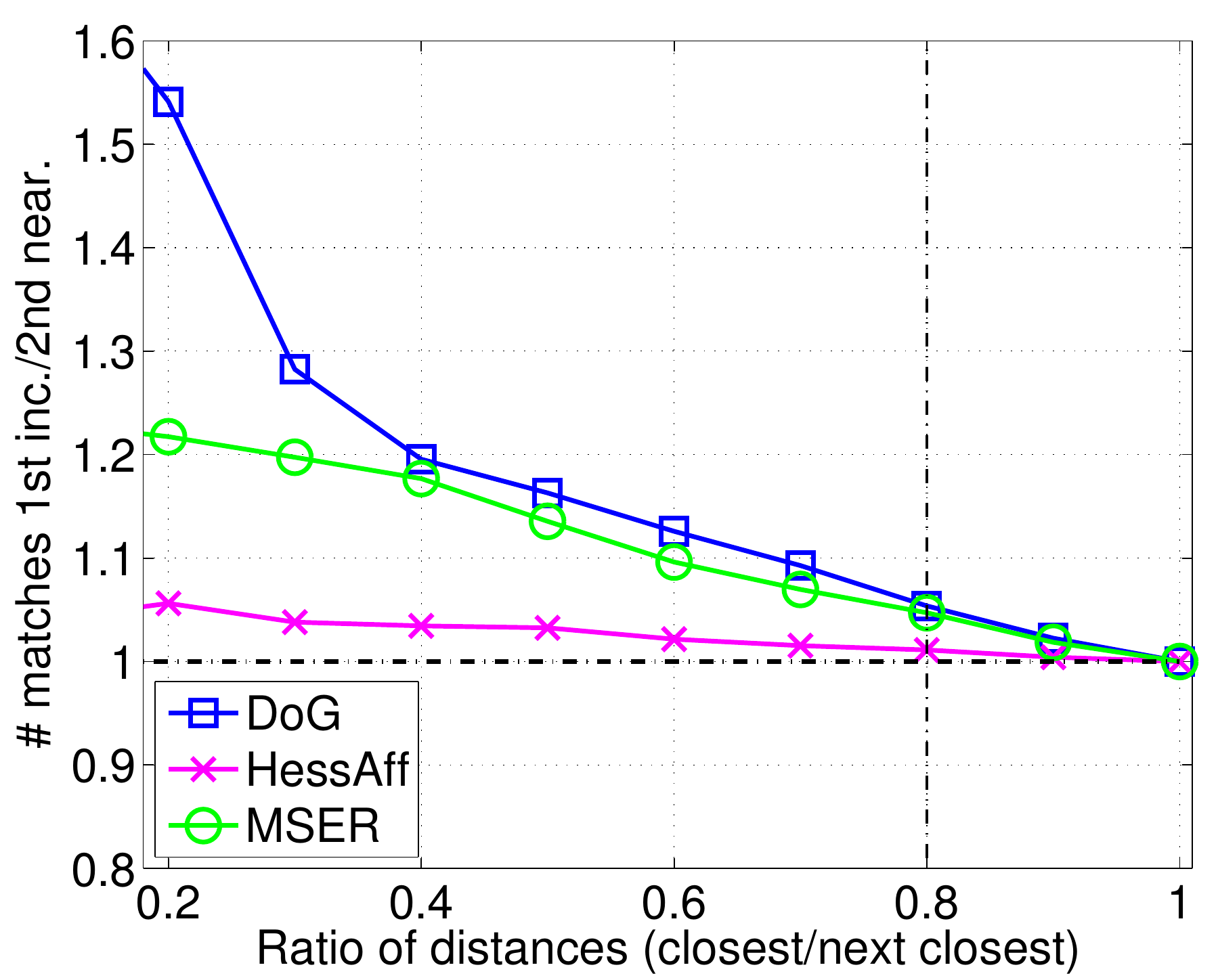}
\includegraphics[width=0.48\linewidth]{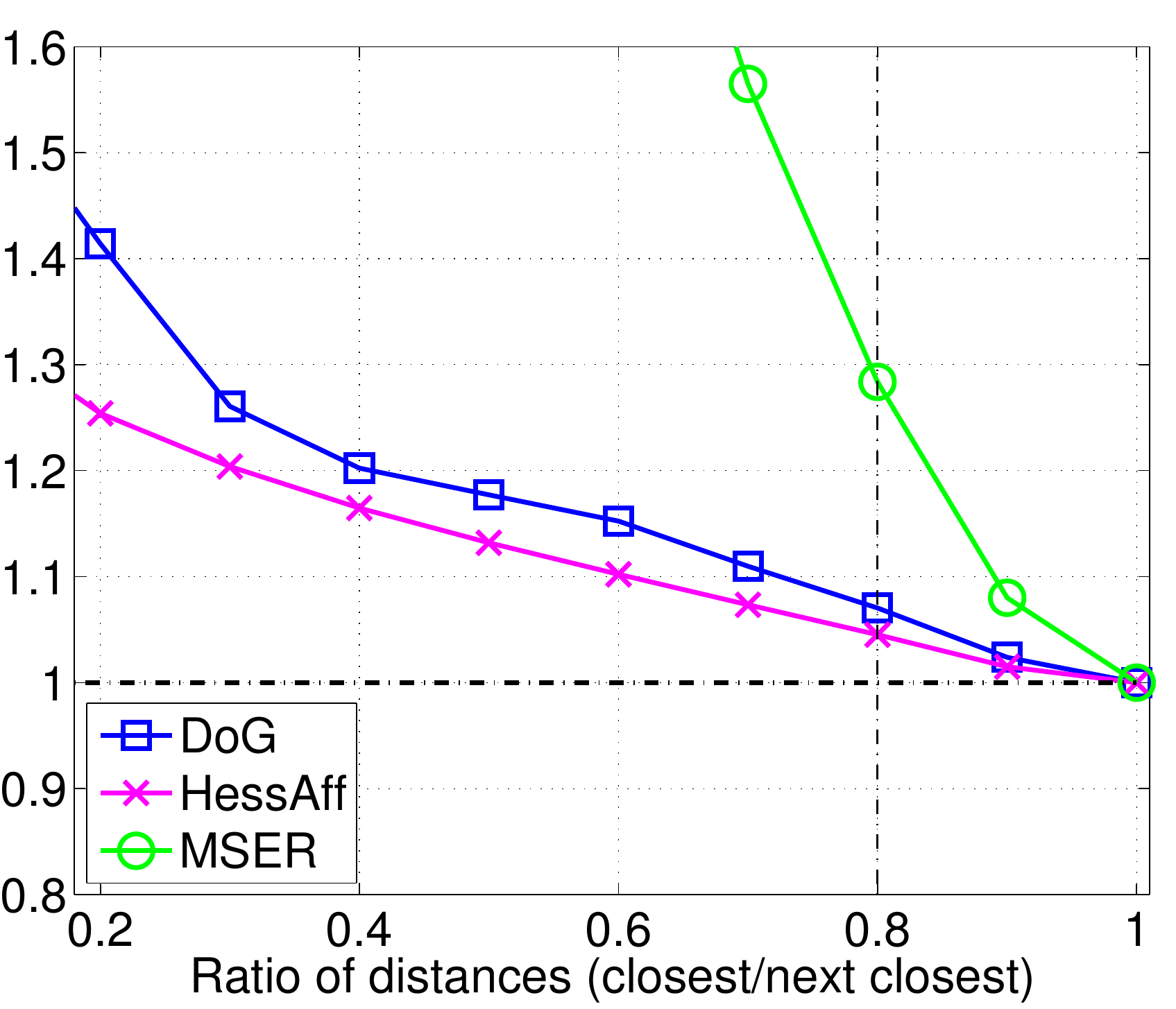}\\
\vspace{-1em} 
 \caption{The ratio of the number of correct matches obtained by the 1st
 inconsistent and 2nd nearest method, without (left) and with (right) view
 synthesis. The black dashed line denotes the widely used distance ratio threshold = 0.8.}
 \label{fig:1st-to-2nd-synth}
 \end{figure}

The \emph{first to first geometrically inconsistent} strategy was
evaluated on 50 image pairs of the publicly available datasets~\cite{Mikolajczyk2005a} and~\cite{Cordes2011}. The cumulative
distributions of the number of correct tentative correspondences
as functions of the descriptor distance ratio are used for comparison. The new
matching strategy improves the performance by up to 5\% for the matching without
view synthesis and up to 30\% (see Figure~\ref{fig:1st-to-2nd-synth}) for matching with view synthesis at almost no additional computational costs.
\subsection{Results on the Extreme Viewpoint Dataset}
\label{main:real-world-testing}
We introduce a two-view matching evaluation dataset\footnote{Available at http://cmp.felk.cvut.cz/wbs/index.html} with extreme viewpoint
changes, see Table~\ref{tabular:image-dataset}. The dataset includes image
pairs from publicly available datasets: {\sc adam} and {\sc
mag}~\cite{Morel2009}, {\sc graf}~\cite{Mikolajczyk2005a} and {\sc
there}~\cite{Cordes2011}. The ground truth homography matrices were estimated by
LO-RANSAC using correspondences from all three detectors in view synthesis
configuration $\{t\} = \{1;\sqrt{2};2;2\sqrt{2};4;4\sqrt{2};8\}$,
$\Delta\phi=72\degree{}/t$. The number of inliers for each image pair was $\geq$
50 and the homographies were manually inspected. For the image pairs {\sc graf}
and {\sc there} precise homographies are provided by
Cordes~\etal~\cite{Cordes2011}. Transition tilts $\tau$ were computed using
equation~(\ref{eq:affine-matrix-decomposition}) with SVD decomposition of the
linearised homography at center of the first image of the pair (see Table ~\ref{tabular:image-dataset}).

\begin{table*}[htbp]
\caption{The Extreme View Dataset -- EVD. Image sources: C --
Cordes~\etal~\cite{Cordes2011}, Ox -- Mikolajczyk~\etal~\cite{Mikolajczyk2005a},
M -- Morel and Yu~\cite{Morel2009}. }
\label{tabular:image-dataset}
\centering
\scriptsize
\bgroup
\def\arraystretch{1.1}
\setlength{\tabcolsep}{.54em}
\tabulinesep=0.5mm
\begin{tabu}spread 0pt{*{16}{|X[-1m,c]}|}
\hline
\# & 1 & 2& 3&4&5&6&7&8&9&10&11&12&13&14&15\\
\hline
Name & {\sc there} &{\sc graf}&{\sc adam}&{\sc mag}&{\sc grand}&{\sc pkk}&{\sc face}&{\sc girl}&{\sc shop}&{\sc dum}&{\sc index}&{\sc cafe}&{\sc fox}&{\sc cat}&{\sc vin}\\
Source & C & Ox &M & M& EVD &EVD & EVD& EVD& EVD & EVD &EVD&EVD &EVD & EVD&EVD \\
$\tau$ -- transitional tilt &6.3 &3.6 &4.8& 20& 2.9 & 7.1&6.9 &8.0&9.1 & 6.9&8.5&11.9&22.5&47&49.8\\
\hline
\shortstack{Resolution \\$[$pixels$]$ }&\shortstack{1536\\x\\1024} &\shortstack{800\\x\\640} & \shortstack{600\\x\\450}& \shortstack{600\\x\\450}& \shortstack{1000\\x\\667} & \shortstack{1000\\x\\750}&\shortstack{1000\\x\\750} &\shortstack{1000\\x\\750}&\shortstack{1000\\x\\562}&\shortstack{1000\\x\\729} & \shortstack{1000\\x\\750}&\shortstack{800\\x\\533}&\shortstack{1000\\x\\563}&\shortstack{1000\\x\\598}&\shortstack{1000\\x\\715}\\
\hline
\end{tabu}
\egroup
\\
\bgroup
\def\arraystretch{1.2}%
\setlength{\tabcolsep}{.16667em}
\tabulinesep=0.2mm
\begin{tabu}spread 0pt{*{9}{|X[-1m,c]}|}
\hline
\# & Image 1 & Image 2 &\# & Image 1 & Image 2&\# & Image 1 & Image 2 \\
\hline
1
& \includegraphics[height=1.5cm]{images/dataset/1/there.png} 
& \includegraphics[height=1.5cm]{images/dataset/2/there.png}
& 6 
& \includegraphics[height=1.5cm]{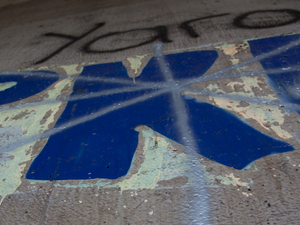} 
& \includegraphics[height=1.5cm]{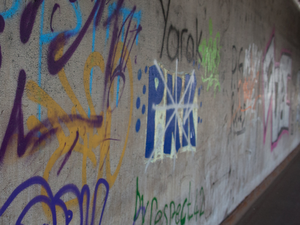} 
&11
& \includegraphics[height=1.5cm]{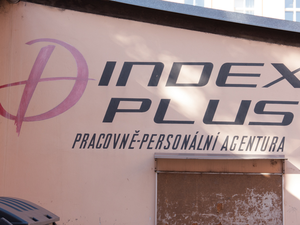} 
& \includegraphics[height=1.5cm]{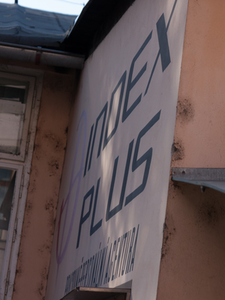} \\ 
\hline
2
& \includegraphics[height=1.5cm]{images/dataset/1/graf.png} 
& \includegraphics[height=1.5cm]{images/dataset/2/graf.png}
&7
& \includegraphics[height=1.5cm]{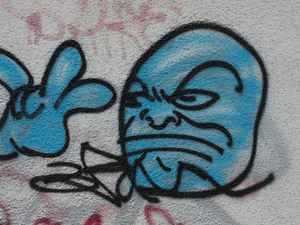} 
& \includegraphics[height=1.5cm]{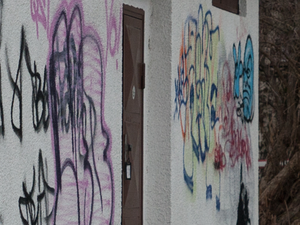}
&
12& \includegraphics[height=1.5cm]{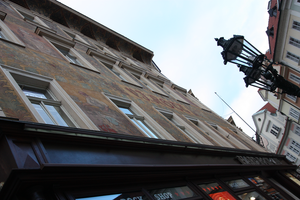} 
& \includegraphics[height=1.5cm]{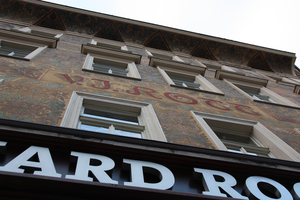} \\ 
\hline
3 
& \includegraphics[height=1.5cm]{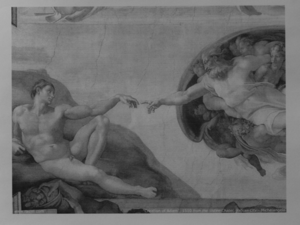} 
& \includegraphics[height=1.5cm]{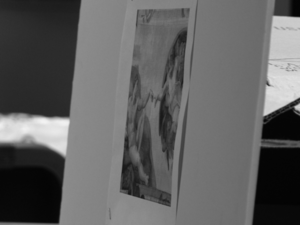} 
&8
& \includegraphics[height=1.5cm]{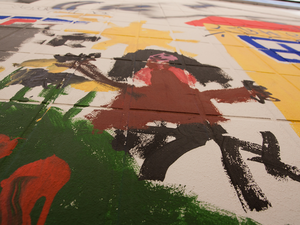} 
& \includegraphics[height=1.5cm]{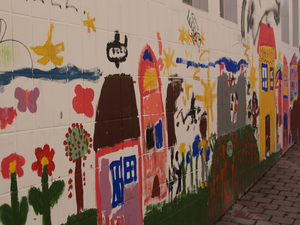}
& 13
& \includegraphics[height=1.5cm]{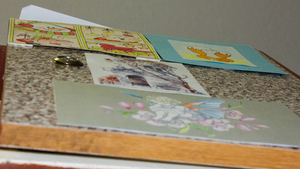} 
& \includegraphics[width=1.5cm, angle=90]{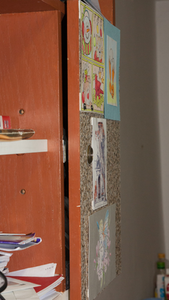}\\ 
\hline
4
& \includegraphics[height=1.5cm]{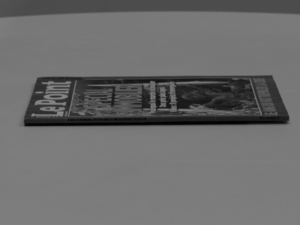} 
& \includegraphics[height=1.5cm]{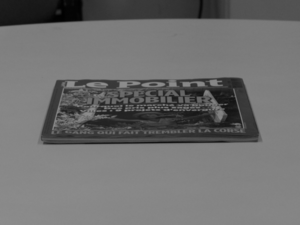}
&9 
& \includegraphics[width=1.5cm, angle=90]{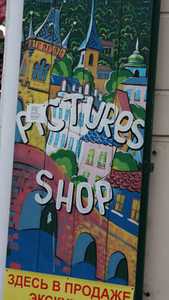} 
& \includegraphics[width=1.5cm, angle=90]{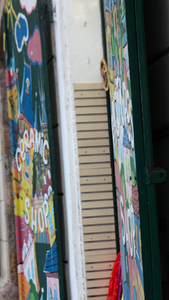}
&14 
&\includegraphics[width=1.5cm, angle=90]{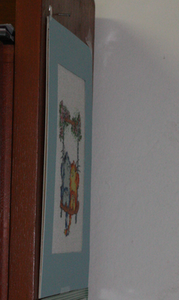} 
& \includegraphics[height=1.5cm]{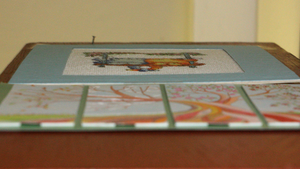} \\
\hline
5
&  \includegraphics[height=1.5cm]{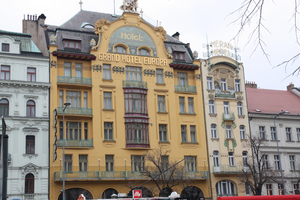} 
& \includegraphics[height=1.5cm]{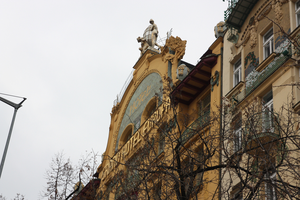}
&10
& \includegraphics[height=1.5cm]{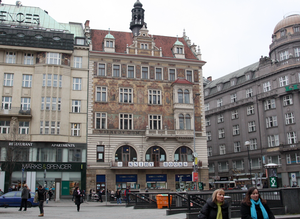} 
& \includegraphics[height=1.5cm]{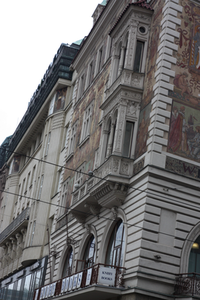}
&15
& \includegraphics[height=1.5cm]{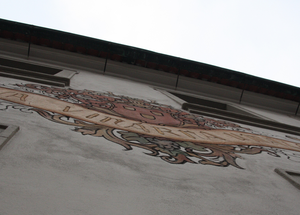}
& \includegraphics[height=1.5cm]{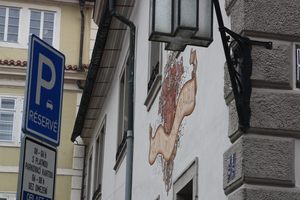} \\ 
\hline
\end{tabu}
\egroup
\end{table*}
%

The configurations evaluated are specified in Table~\ref{tabular:view-synth-configuration}. For comparison, ASIFT\footnoteremember{footnoteASIFT}{Reference code from
http://demo.ipol.im/demo/my\_affine\_sift} results are added. Computations were performed on Intel i5 CPU @ 2.6GHz with 4Gb RAM; results for
computation on one core are provided.
Based on results of the different configuration, we have chosen the following configuration for MODS w.r.t increasing computation time and performance of the configurations -- see Table~\ref{tabular:MODS-configuration}. Please note that only views complementary to the previous iterations are synthesised.

\begin{figure*}[htbp]
\vspace{-1em} 
\centering
\includegraphics[width=0.45\linewidth]{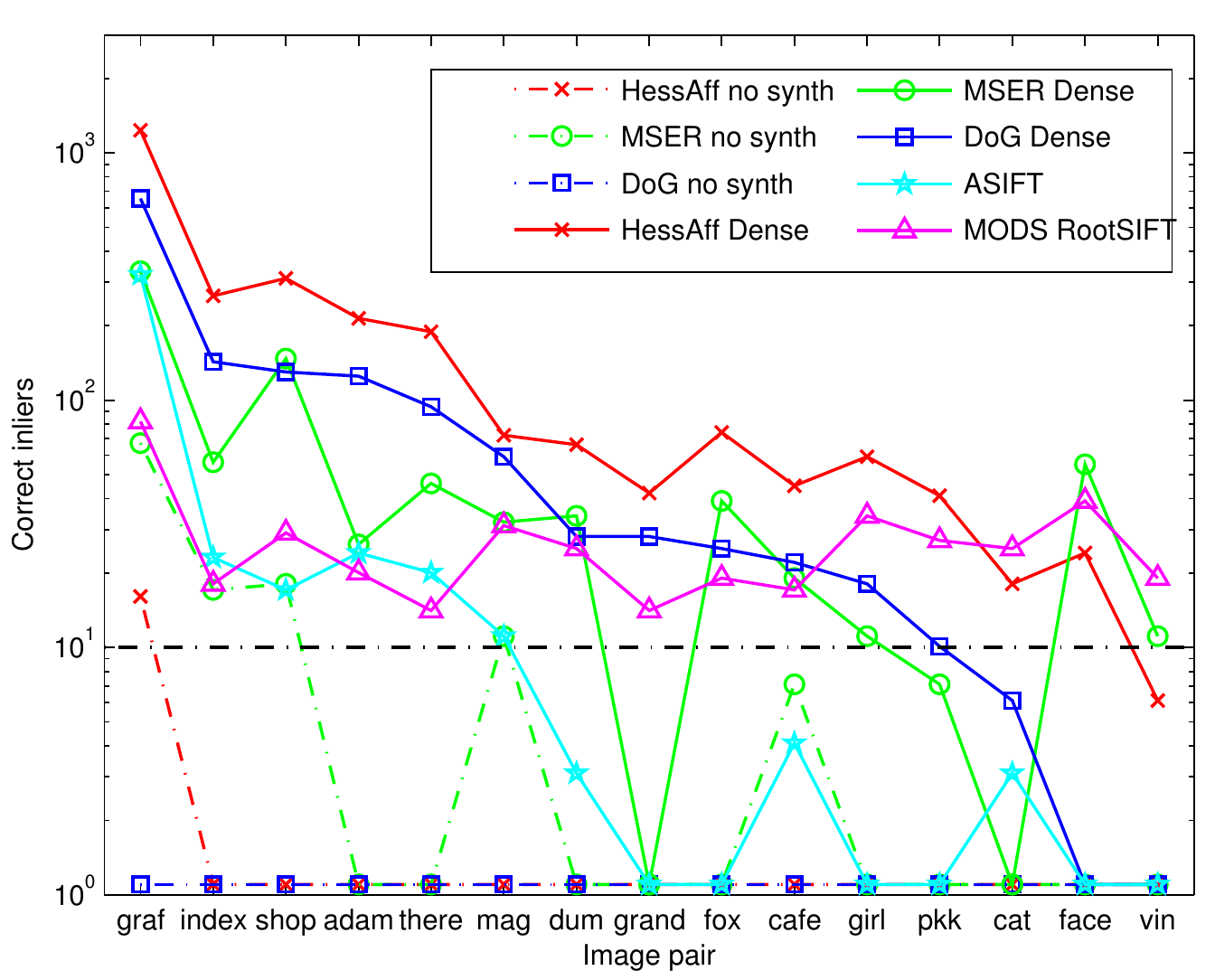}
\includegraphics[width=0.45\linewidth]{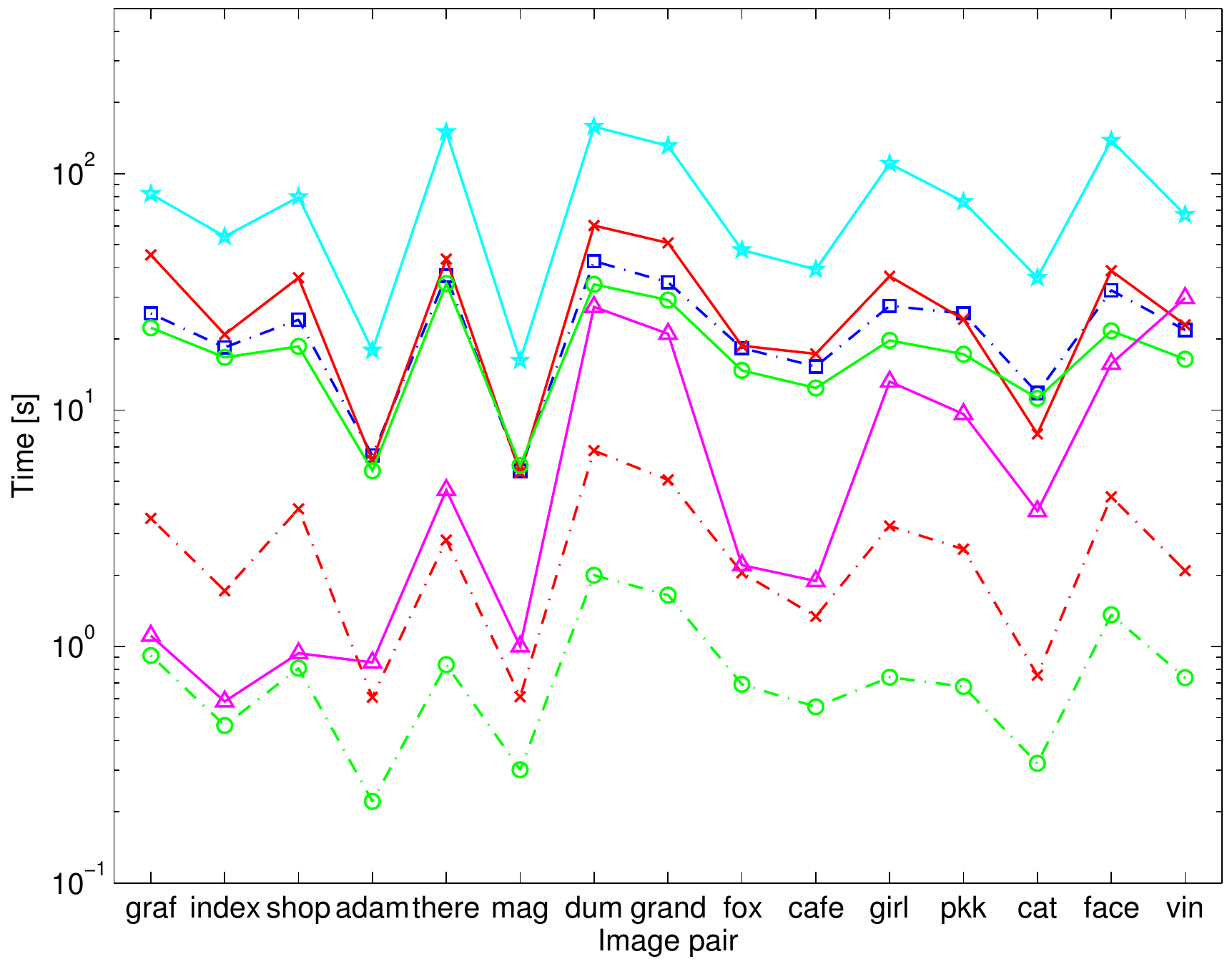}
\caption{Performance of the selected view synthesis configurations defined in
Table~\ref{tabular:view-synth-configuration}. MODS set to find $\geq$ 15 inliers. Left -- the number of
correct RANSAC inliers. 
The black dashed line marks the level of 10 correct inlier -- a minimum
for a reliable estimate of two-view geometry. Right -- runtime (1 core).}
\label{fig:inliers-bars}
\vspace{-1em}
\end{figure*}

The MODS algorithm allows to set the minimum desired number of inliers as a stopping criterion. The recommended value -- 15 inliers to the homography, have a very low probability to be a random result, but are few enough to show the time gain from the algorithm. To maximize the number of inliers for each of the detectors, we recommend to use {\sc Dense} configuration as a single step.
\begin{table}[tb]
\caption{Configurations for MODS steps}
\label{tabular:MODS-configuration}
\footnotesize 
\centering
\bgroup
\tabulinesep=0.4mm
\begin{tabu} spread 0pt{|X[-1m,c]|X[-1,m,l]|}
\hline
Iter. & \centering Setup\\
\hline
1 & MSER,$\{S\} = \{1;0.25;0.125\}$, $\{t\} = \{1\}$, $\Delta\phi=360\degree{}/t$ \\
\hline
2 & MSER,$\{S\} = \{1;0.25;0.125\}$, $\{t\} = \{1;5;9\}$, $\Delta\phi=360\degree{}/t$\\
\hline
3 & HessAff, $\{S\} = \{1\}$, $\{t\} = \{1;\sqrt{2};2;2\sqrt{2};4;4\sqrt{2};8\}$, $\Delta\phi=360\degree{}/t$\\
\hline
4& HessAff , $\{S\} = \{1\}$, $\{t\} = \{1;2;4;6;8\}$, $\Delta\phi=72\degree{}/t$ \\
\hline
\end{tabu}
\egroup
\end{table}
Figure~\ref{fig:inliers-bars} and Table~\ref{tabular:summary-test} compare the
different view synthesis configurations and the "affine-covariant"
detectors -- they generate more correct matches in a shorter
time than the DoG detector. The DoG based matching and ASIFT matching cannot
solve 3 resp. 9 out of the 15 image pairs. 
The ASIFT algorithm generates a lower number of correct inliers and works slower than our DoG {\sc Dense} configuration (which has the same tilt-rotation set). The main causes are elimination of "one-to-many", including correct, correspondences, the inferiority of the standard 2nd closest ratio and a simple bruteforce algorithm of matching used in ASIFT. 
 
No single detector solved all image pairs. The Hessian-Affine with {\sc Dense} configuration successfully solved 14 out of 15 image pairs and outperformed other detectors and configurations in the number of inliers, however, at the expense
of the highest computational cost. MSER with no synthesis and in the {\sc Sparse}
configuration is the fastest and could solve 10 out of 15 image pairs.
The MODS algorithm solves all image pairs and saves computational time on processing of the easy pairs at the cost of a small matching overhead on the hard cases. Also, MODS is the fastest algorithm in 7
cases, and in another 2 cases it is just $\sim10\%$ slower than the fastest
configuration. The difference results of MODS step 2 and MSER {\sc Sparse} are caused by randomization in RANSAC and kd-tree building.

\begin{figure}[htb]
\centering
\includegraphics[width=.5\linewidth]{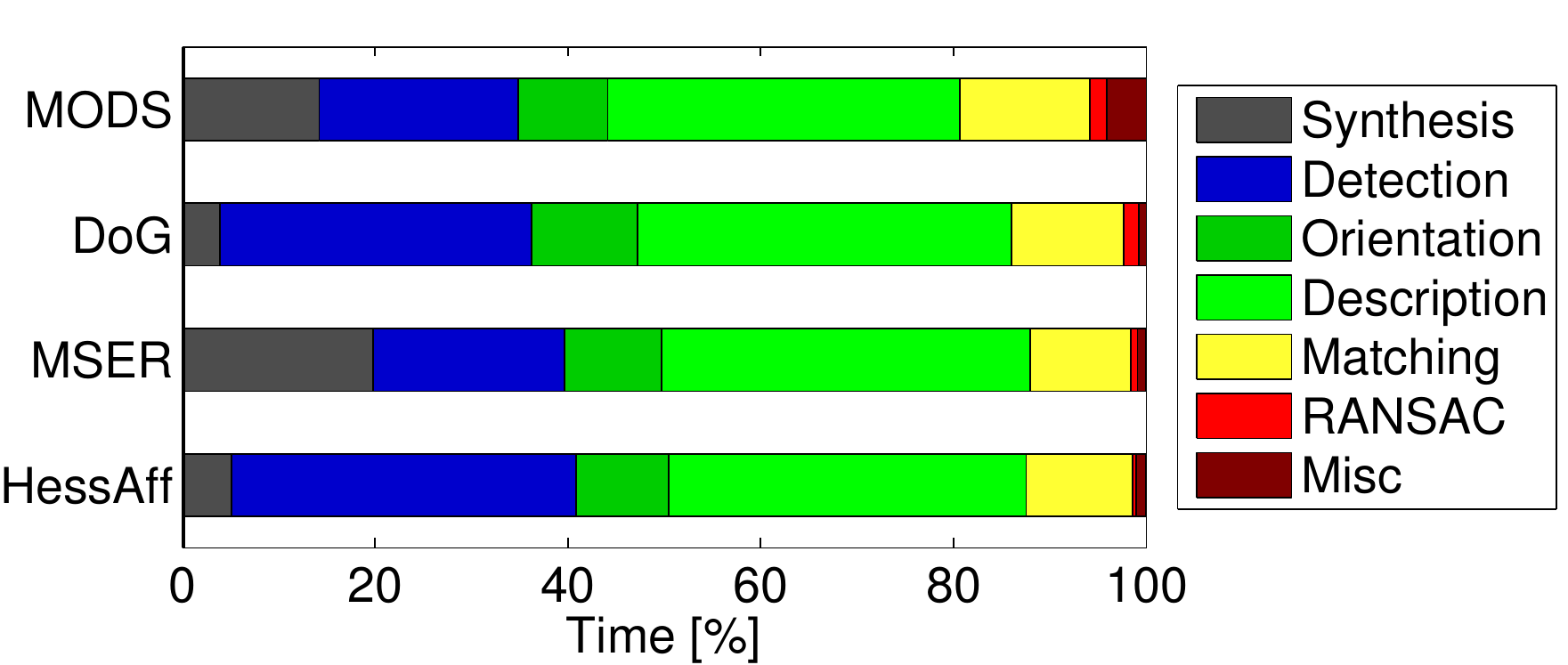} 
\caption{Percentage of time spent in the main stages of the matching with view
synthesis process on a single core, {\sc Dense} configuration. SIFT
description, i.e. the dominant gradient estimation and the descriptor
computation is the most time-consuming part.}
 \label{fig:time-consist}
\end{figure}

Fig.~\ref{fig:time-consist} shows the breakdown of the computational time. 
SIFT description with the dominant orientation estimation take 50\% of the time.
Note that the whole process is almost linear in the area of the synthesised
views. The only super-linear part, matching, takes only 10\% of the time.
\bgroup
\centering
\begin{table*}[htpb]
\centering
\begin{minipage}[h]{1\linewidth}
\centering
\caption{A comparison of different view synthesis and detector configurations
(with RootSIFT). Best results are highlighted by a grey background. MODS set to
find $\geq$ 15 inliers. Results with less than 8 correct inliers are in red.}
\label{tabular:summary-test}
\scriptsize 
\begin{tabular}{|=l|rrrrrr|,,,,,,|,,,,,,|}
\hline
Image& \multicolumn{6}{c|}{Correct inliers}
& \multicolumn{6}{c|}{Time, 1 core $[s]$}
& \multicolumn{6}{c|}{Correct inliers/sec}\\
\hline
&\begin{sideways}MODS, $\theta_m$ = 15\end{sideways}
&\begin{sideways}ASIFT\end{sideways} 
&\begin{sideways}MSER {\sc Sparse}\end{sideways}
&\begin{sideways}HessAff {\sc Sparse}\end{sideways}
&\begin{sideways}HessAff {\sc Dense}\end{sideways} 
&\begin{sideways}DoG {\sc Dense}\end{sideways} 
&\begin{sideways}MODS, $\theta_m$ = 15 \end{sideways}
&\begin{sideways}ASIFT\end{sideways} 
&\begin{sideways}MSER {\sc Sparse}\end{sideways}
&\begin{sideways}HessAff {\sc Sparse}\end{sideways}
&\begin{sideways}HessAff {\sc Dense}\end{sideways} 
&\begin{sideways}DoG {\sc Dense}\end{sideways} 
&\begin{sideways}MODS, $\theta_m$ = 15 \end{sideways}
&\begin{sideways}ASIFT\end{sideways} 
&\begin{sideways}MSER {\sc Sparse}\end{sideways}
&\begin{sideways}HessAff {\sc Sparse}\end{sideways}
&\begin{sideways}HessAff {\sc Dense}\end{sideways} 
&\begin{sideways}DoG {\sc Dense}\end{sideways}\\ 
\hline
 graf&82&322&165&375&\cellcolor[gray]{0.9}1235&653&\cellcolor[gray]{0.9}1.0&81.8&3.0&11.0&45.2&25.5&\cellcolor[gray]{0.9}83.9&3.9&55&34.1&27.3&25.6\\ 
index&18&23&24&34&\cellcolor[gray]{0.9}264&143&\cellcolor[gray]{0.9}0.5&54.1&2.2&5.4&20.8&18.3&\cellcolor[gray]{0.9}38.1&0.4&11.1&6.3&12.7&7.8\\ 
shop&29&17&73&133&\cellcolor[gray]{0.9}311&130&\cellcolor[gray]{0.9}0.8&79.5&2.5&10.1&36.2&24&\cellcolor[gray]{0.9}35.2&0.2&28.7&13.2&8.6&5.4\\ 
adam&20&24&18&86&\cellcolor[gray]{0.9}214&125&0.8&17.8&\cellcolor[gray]{0.9}0.7&1.6&6.0&6.3&26.7&1.3&24.3&\cellcolor[gray]{0.9}54.1&35.6&19.8\\ 
there&14&20&12&49&\cellcolor[gray]{0.9}189&94&\cellcolor[gray]{0.9}4.5&150.0&\cellcolor[gray]{0.9}4.5&10.1&43.4&36.9&3.1&0.1&2.7&\cellcolor[gray]{0.9}4.9&4.4&2.5\\ 
mag&31&11&28&54&\cellcolor[gray]{0.9}72&59&\cellcolor[gray]{0.9}0.8&16.1&\cellcolor[gray]{0.9}0.8&1.6&5.3&5.4&\cellcolor[gray]{0.9}37.3&0.7&34.4&33.5&13.5&10.9\\ 
dum&25&\rowstyle{\color{red}}\rowstyle{\color{red}}3&\rowstyle{\color{red}}\rowstyle{\color{red}}0&10&\cellcolor[gray]{0.9}66&28&29.4&158.0&4.8&\cellcolor[gray]{0.9}20.1&60.2&42.5&0.9&0.0&0.0&0.5&\cellcolor[gray]{0.9}1.1&0.7\\ 
grand&14&\rowstyle{\color{red}}\rowstyle{\color{red}}0&9&\rowstyle{\color{red}}\rowstyle{\color{red}}0&\cellcolor[gray]{0.9}42&28&21.9&131.0&\cellcolor[gray]{0.9}4.2&14.8&50.8&34.6&0.6&0.0&\cellcolor[gray]{0.9}2.1&0.0&0.8&0.8\\ 
fox&19&\rowstyle{\color{red}}\rowstyle{\color{red}}0&19&22&\cellcolor[gray]{0.9}74&25&\cellcolor[gray]{0.9}2.1&47.4&\cellcolor[gray]{0.9}2.1&5.8&18.6&18.2&9.0&0.0&\cellcolor[gray]{0.9}9.3&3.8&4&1.4\\ 
cafe&17&\rowstyle{\color{red}}\rowstyle{\color{red}}4&14&\rowstyle{\color{red}}\rowstyle{\color{red}}0&\cellcolor[gray]{0.9}45&22&1.8&39.2&\cellcolor[gray]{0.9}1.7&4.5&17.2&15.2&\cellcolor[gray]{0.9}9.3&0.1&8.2&0.0&2.6&1.4\\ 
girl&34&\rowstyle{\color{red}}\rowstyle{\color{red}}0&\rowstyle{\color{red}}\rowstyle{\color{red}}0&14&\cellcolor[gray]{0.9}59&18&13.1&110.0&2.7&\cellcolor[gray]{0.9}10.0&36.7&27.5&\cellcolor[gray]{0.9}2.6&0.0&0.0&1.4&1.6&0.7\\ 
pkk&27&\rowstyle{\color{red}}\rowstyle{\color{red}}0&\rowstyle{\color{red}}\rowstyle{\color{red}}6&12&\cellcolor[gray]{0.9}41&10&9.5&75.9&2.4&\cellcolor[gray]{0.9}6.8&24.1&25.5&\cellcolor[gray]{0.9}2.8&0.0&2.5&1.8&1.7&0.4\\ 
cat&\cellcolor[gray]{0.9}25&\rowstyle{\color{red}}\rowstyle{\color{red}}3&\rowstyle{\color{red}}\rowstyle{\color{red}}0&21&18&\rowstyle{\color{red}}\rowstyle{\color{red}}6&3.9&36.2&1.4&\cellcolor[gray]{0.9}2.2&7.8&11.7&6.3&0.1&0.0&\cellcolor[gray]{0.9}9.6&2.3&0.5\\ 
face&\cellcolor[gray]{0.9}39&\rowstyle{\color{red}}\rowstyle{\color{red}}0&9&17&24&\rowstyle{\color{red}}\rowstyle{\color{red}}0&15.6&138.0&\cellcolor[gray]{0.9}3.4&11.3&38.8&32.0&2.5&0.0&\cellcolor[gray]{0.9}2.7&1.5&0.6&0.0\\ 
vin&\cellcolor[gray]{0.9}19&\rowstyle{\color{red}}\rowstyle{\color{red}}0&\rowstyle{\color{red}}\rowstyle{\color{red}}0&\rowstyle{\color{red}}\rowstyle{\color{red}}0&\rowstyle{\color{red}}\rowstyle{\color{red}}6&\rowstyle{\color{red}}\rowstyle{\color{red}}0&\cellcolor[gray]{0.9}30.3&66.9&2.3&6.3&22.8&21.7&\cellcolor[gray]{0.9}0.6&0.0&0.0&0.0&0.3&0.0\\ 

\hline
\end{tabular}
\end{minipage}
\end{table*}
\egroup
\bgroup
\centering
\begin{table*}[htpb]
\centering
\begin{minipage}[h]{0.90\linewidth}
\caption{MODS ($\theta_m$ = 15) performance on the EVD dataset.
The k-th iteration includes regions from all previous steps.}
\label{tabular:test-MODS-sift}
\end{minipage}
\scriptsize
\centering
\begin{minipage}[h]{0.45\linewidth}
\setlength{\tabcolsep}{.15em}
\begin{tabular}{|=l|=c...|=r+r|=r+r.|=r+r|}
\hline
Image 
& \multicolumn{11}{c|}{\shortstack{MODS (SIFT) 4 steps. \\1. MSER Scale only. 2. MSER {\sc Sparse}.\\ 3. HessAff {\sc Sparse}. 4. HessAff  {\sc Max} }} \\
\hline
& \multicolumn{4}{c|}{Time}
& \multicolumn{2}{c|}{RANSAC}
& \multicolumn{3}{c|}{\shortstack{Tentatives \\quality} }
& \multicolumn{2}{c|}{Regions}\\
\hline
&&&&& \multicolumn{2}{c|}{SIFT}
& \multicolumn{3}{c|}{SIFT}
& &\\
&\begin{sideways}Iteration\end{sideways}
&\begin{sideways}1 core $[s]$\end{sideways}
&\begin{sideways}2 cores $[s]$\end{sideways}
&\begin{sideways}4 cores $[s]$\end{sideways}
&\begin{sideways}Correct inliers\end{sideways} 
&\begin{sideways} Inliers\end{sideways}
&\begin{sideways}Correct matches\end{sideways} 
&\begin{sideways}Tentatives\end{sideways}
&\begin{sideways}Correct  / all $[\%]$\end{sideways}
&\begin{sideways}Image 1\end{sideways}
&\begin{sideways}Image 2\end{sideways}\\
\hline
 graf&1&1&0.8&0.8&81&85&82&160&51.2&1018&1674\\ 
index&1&0.5&0.4&0.4&19&20&19&56&33.9&411&246\\ 
shop&1&0.8&0.7&0.7&28&30&28&84&33.3&1321&711\\ 
adam&2&0.8&0.5&0.3&19&22&21&48&43.8&357&164\\ 
there&2&4.5&2.8&2&10&18&15&66&22.7&571&2833\\ 
mag&2&0.8&0.5&0.4&30&31&30&52&57.7&393&509\\ 
dum&3&29.4&18.9&15.5&24&29&30&1136&2.6&33342&23711\\ 
grand&3&21.9&13.7&11.3&17&25&21&754&2.8&24731&20297\\ 
fox&2&2.1&1.4&1.1&16&19&19&76&25&1717&1011\\ 
cafe&2&1.8&1.2&0.9&18&20&18&142&12.7&1402&1319\\ 
girl&3&13.1&7.4&5.3&35&46&38&549&6.9&10460&16105\\ 
pkk&2&2.5&1.6&1.3&\rowstyle{\color{red}}7&\rowstyle{\color{red}}15&10&81&12.3&1229&1267\\ 
cat&3&3.9&2&1.4&35&38&35&143&24.5&1262&3279\\ 
face&2&3.6&2.4&2&9&15&11&118&9.3&3411&2371\\ 
vin&4&30.3&17.8&12.5&18&38&22&657&3.4&18956&31984\\ 

\hline
\end{tabular}
\end{minipage}
\begin{minipage}[h]{0.45\linewidth}
\setlength{\tabcolsep}{.15em}
\begin{tabular}{|=l|=c...|=r+r|=r+r.|=r+r|}
\hline
Image 
& \multicolumn{11}{c|}{\shortstack{MODS (RootSIFT), 4 steps. \\1. MSER Scale only. 2. MSER {\sc Sparse}.\\3. HessAff {\sc Sparse}. 4. HessAff  {\sc Max} }} \\
\hline
& \multicolumn{4}{c|}{Time}
& \multicolumn{2}{c|}{RANSAC}
& \multicolumn{3}{c|}{\shortstack{Tentatives\\ quality}}
& \multicolumn{2}{c|}{Regions}\\
\hline
&&&&& \multicolumn{2}{c|}{RootSIFT}
& \multicolumn{3}{c|}{RootSIFT}
& &\\
&\begin{sideways}Iteration\end{sideways}
&\begin{sideways}1 core $[s]$\end{sideways}
&\begin{sideways}2 cores $[s]$\end{sideways}
&\begin{sideways}4 cores $[s]$\end{sideways}
&\begin{sideways}Correct inliers\end{sideways} 
&\begin{sideways} Inliers\end{sideways}
&\begin{sideways}Correct matches\end{sideways} 
&\begin{sideways}Tentatives\end{sideways}
&\begin{sideways}Correct  / all $[\%]$\end{sideways}
&\begin{sideways}Image 1\end{sideways}
&\begin{sideways}Image 2\end{sideways}\\
\hline
 graf&1&1&0.8&0.8&82&87&83&154&53.9&1018&1674\\ 
index&1&0.5&0.4&0.4&18&20&18&42&42.9&411&246\\ 
shop&1&0.8&0.7&0.7&29&31&29&61&47.5&1321&711\\ 
adam&2&0.8&0.5&0.4&20&23&22&47&46.8&357&164\\ 
there&2&4.5&2.8&2&14&17&16&60&26.7&571&2833\\ 
mag&2&0.9&0.5&0.4&31&31&31&44&70.5&393&509\\ 
dum&3&27.2&16.9&13.5&25&32&29&850&3.4&33342&23711\\ 
grand&3&20.9&12.5&10&14&24&19&468&4.1&24731&20297\\ 
fox&2&2.1&1.4&1.1&19&20&20&62&32.3&1717&1011\\ 
cafe&2&1.8&1.2&0.9&17&21&18&117&15.4&1402&1319\\ 
girl&3&13.1&7.3&5.2&34&44&38&436&8.7&10460&16105\\ 
pkk&3&9.5&5.3&4&27&37&33&344&9.6&10686&7085\\ 
cat&3&3.6&2.1&1.5&25&34&30&149&20.1&1262&3279\\ 
face&3&15.6&8.9&7.1&39&44&42&534&7.9&18857&13271\\ 
vin&4&29.7&17.1&11.8&19&32&21&455&4.6&18956&31984\\ 

\hline
\end{tabular}
\end{minipage}
\end{table*}
\egroup
\subsection{MSER vs. blur and scale change}
\label{main:mser-blur}
We have tested performance of recommended scale synthesis configuration for MSER on the image pairs most distorted by blur and scale change from the Oxford~\cite{Mikolajczyk2005a} dataset. To allow comparison with~\cite{Mikolajczyk2005a}, the standard SIFT was used instead of RootSIFT in this experiment. Note that the results are not fully compatible as we use NN-distance ratio matching threshold = 0.8 (In~\cite{Mikolajczyk2005a} no ratio threshold has been used, so the absolute number of the matches differs a lot. But relative ratio between detectors performance remains the same). We have also performed the duplicate filtering procedure, which reduces the number of correspondences (\cf Section~\ref{sec:method}). 
 
\begin{figure}[htb]
\includegraphics[width=0.49\linewidth]{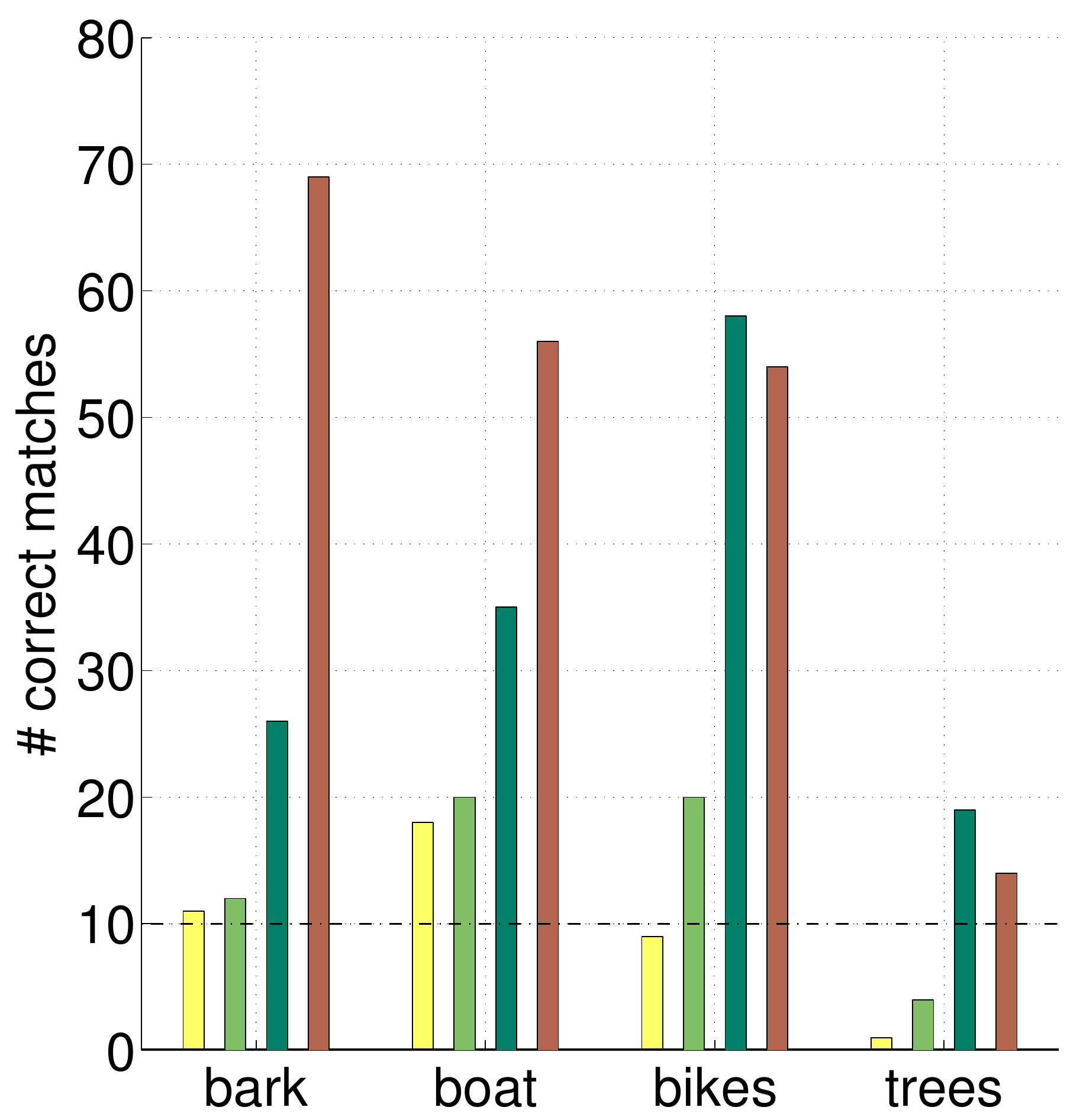} 
\includegraphics[width=0.49\linewidth]{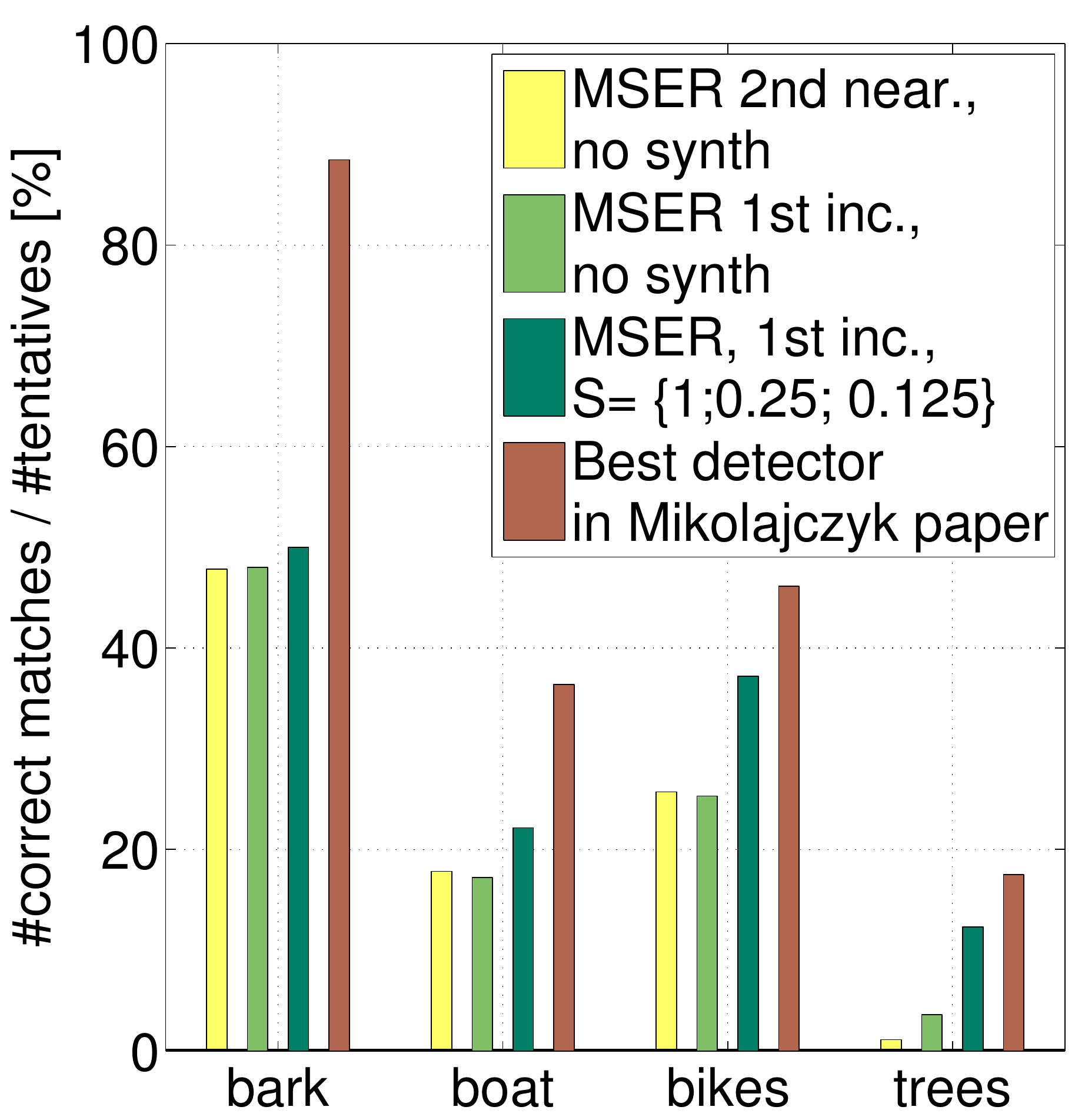} 
\caption{MSER performance with and w/o scale synthesis on the most distorted
pairs (1-6) with scale change and blur from~\cite{Mikolajczyk2005a}. 
Left -- the number of correct SIFT matches. 
Right -- the proportion of correct matches within tentative correspondences.
The best detectors from~\cite{Mikolajczyk2005a}:
{\sc bark}, {\sc boat}, {\sc trees} -- Hessian-Affine, {\sc bikes} -- IBR 
are shown for comparison.}
 \label{fig:mser-myk}
 \end{figure}

Figure~\ref{fig:mser-myk} shows that scale synthesis with 1st geom. inconsistent rule improves MSER performance by 60\% to 1000\%, solving the most common MSER problems -- sensitivity to blur and scale change. The quality of tentative correspondences also increases with the proposed scale synthesis configuration (Figure~\ref{fig:mser-myk}, right). Table~\ref{tabular:mser-scale-time} shows the computation time.  
\begin{table}[hb]
\centering
\caption{MSER matcher runtime on Oxford~\cite{Mikolajczyk2005a} dataset}
\label{tabular:mser-scale-time} 
\small 
\centering 
\begin{tabular}{|l|l|}
\hline
scale synthesis setup& time [s]\\
\hline
$\{S\} = \{1\}$ &56.6\\
$\{S\} = \{1;0.25;0.125\}$ & 61.5\\
\hline
\end{tabular}
\vspace{-1.0em}
\end{table}
\section{Conclusions}
\label{sec:conclusions}
We have introduced view synthesis to two-view wide-baseline matching with
affine-covariant detectors and shown that matching with the Hessian-Affine
or MSER detectors outperforms the state-of-the-art ASIFT.

To address the robustness vs. speed trade-off, we have proposed the Matching On
Demand with view Synthesis algorithm (MODS) that uses progressively more
synthesized images and more (time-consuming) detectors until a reliable estimate
of geometry is obtained. We show experimentally that the MODS algorithm solves
matching problems beyond the state-of-the-art and yet is comparable in
speed to standard wide-baseline matchers on simpler problems.

Minor contributions include an improved method for tentative correspondence
selection, applicable both with and without view synthesis. A modification of
the standard first to second nearest SIFT distance rule  increases the number
of correct matches by 5-20\% at no additional computational cost. Finally,
we found a simple view synthesis set up costing less than 10\% of time that
greatly improves MSER robustness to blur and scale change.

{\footnotesize
\bibliographystyle{ieee}
\bibliography{main.bbl}
}
\newpage
\appendix 
\part*{Appendix}
\section{Tuning view synthesis parameters}
\label{sec:appendixA}
\paragraph{Estimating threshold on the distance ratio.} The well known~\cite{Lowe2004}
matching strategy for SIFT descriptors is based on the distance ratio of
the first to the second closest descriptor. The aim of this experiment is
to set the threshold of the proposed modification --
first to first geometrically inconsistent matching strategy.

To estimate the threshold we used 50 image pairs of the publicly available
datasets~\cite{Mikolajczyk2005a} and~\cite{Cordes2011}, all pairs are
provided with known homography transformation. The detectors -- MSER,
Hessian-Affine, DoG -- were run on all pairs of images and distances between
all descriptors in each pair computed. Then the closest, second closest and
closest geometrically inconsistent descriptors were identified. The cumulative
distributions of the number of correct and incorrect tentative
correspondences as a function of the distance ratio were computed for both
strategies using the ground truth homographies.

\begin{figure*}[ht]
\includegraphics[width=0.50\linewidth]{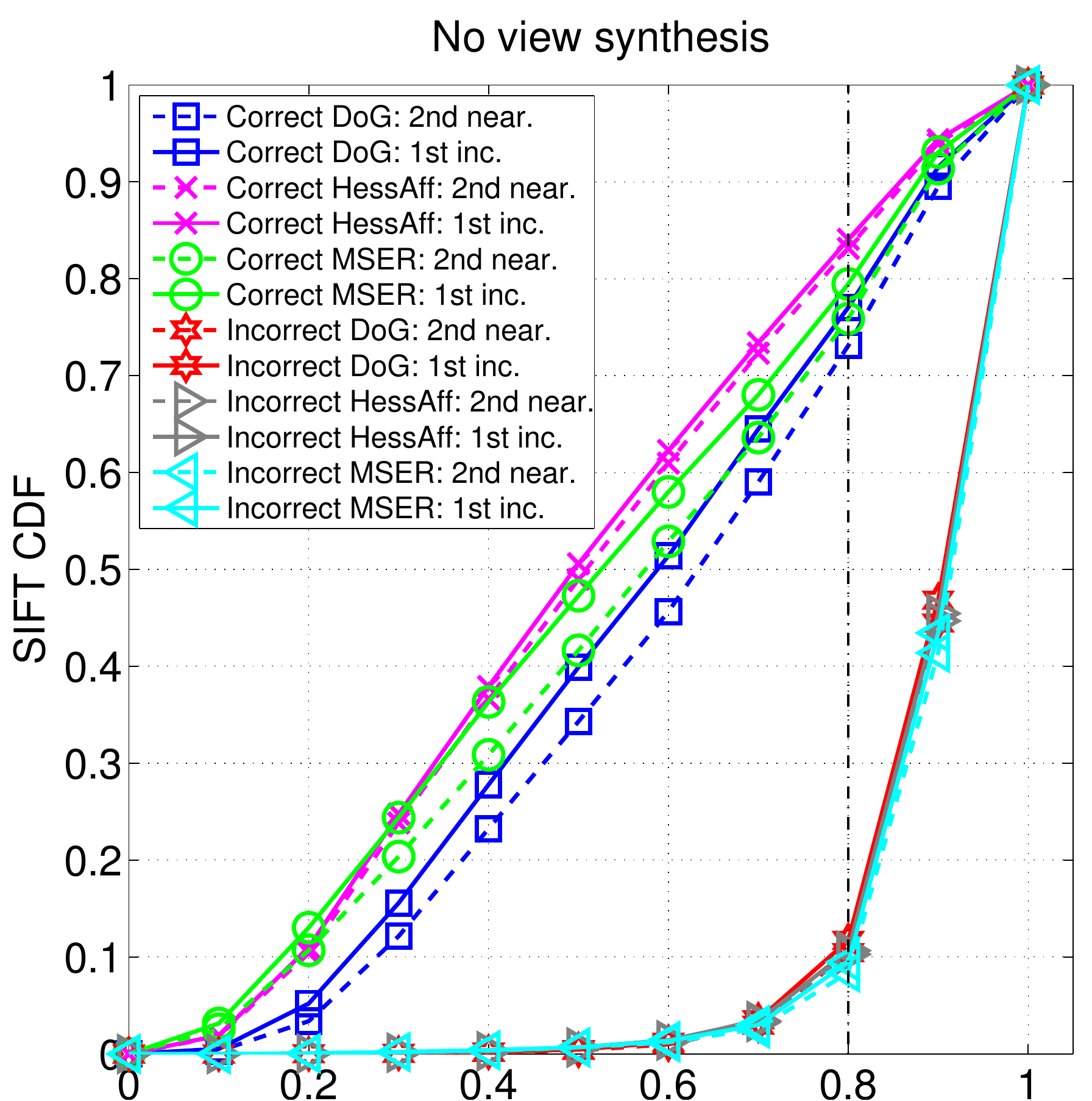}
\includegraphics[width=0.47\linewidth]{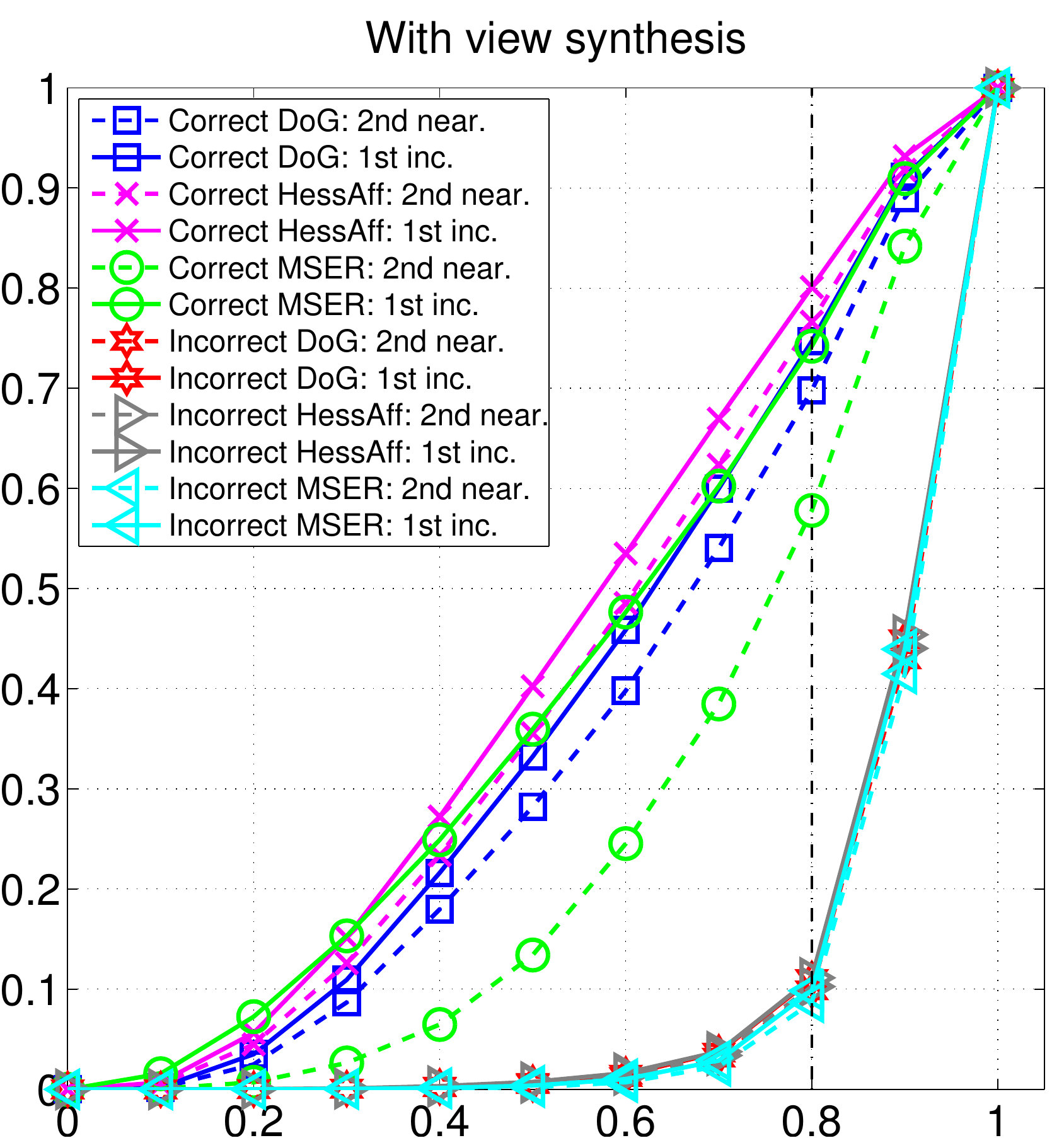}\\
\includegraphics[width=0.50\linewidth]{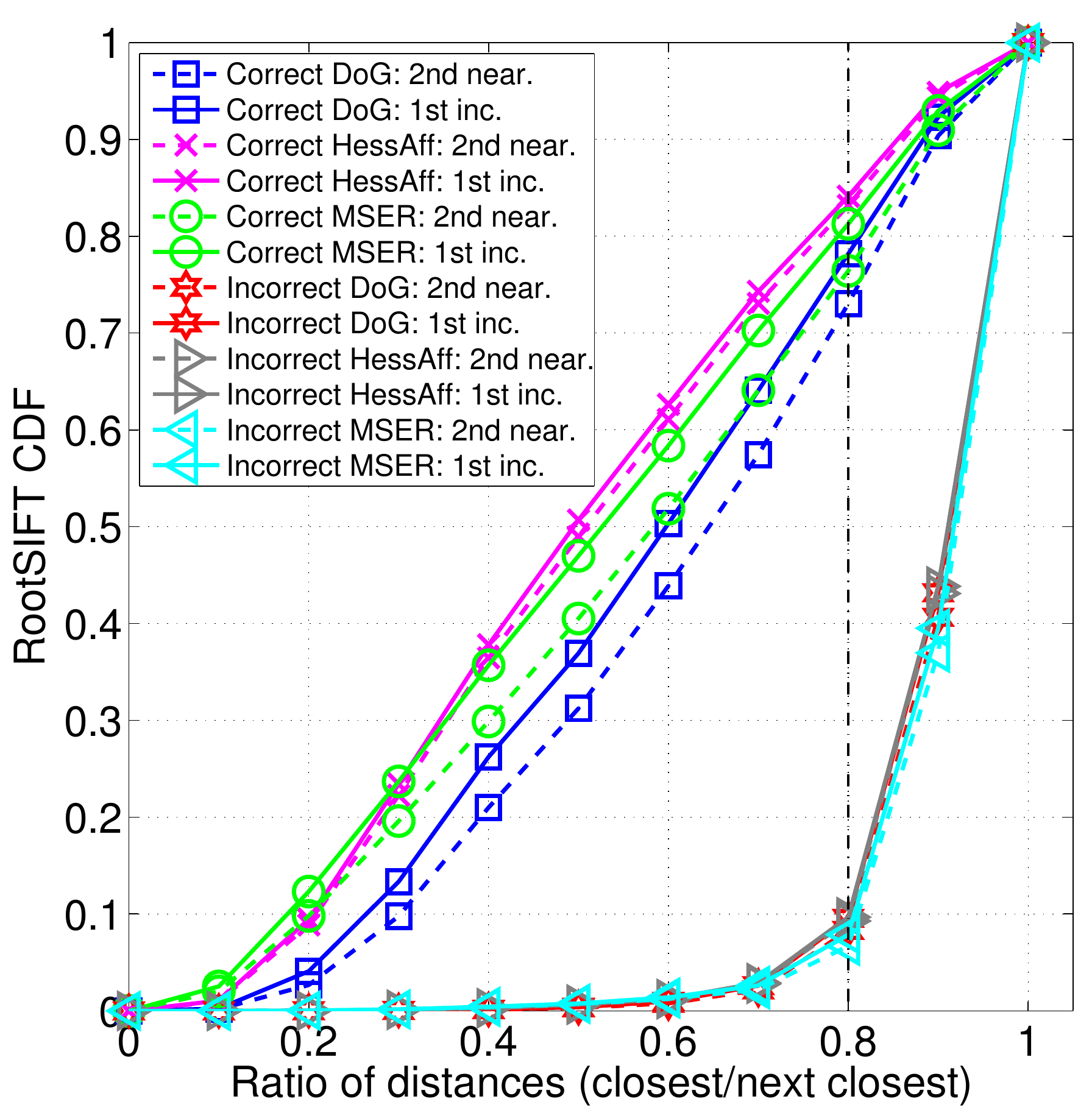}
\includegraphics[width=0.47\linewidth]{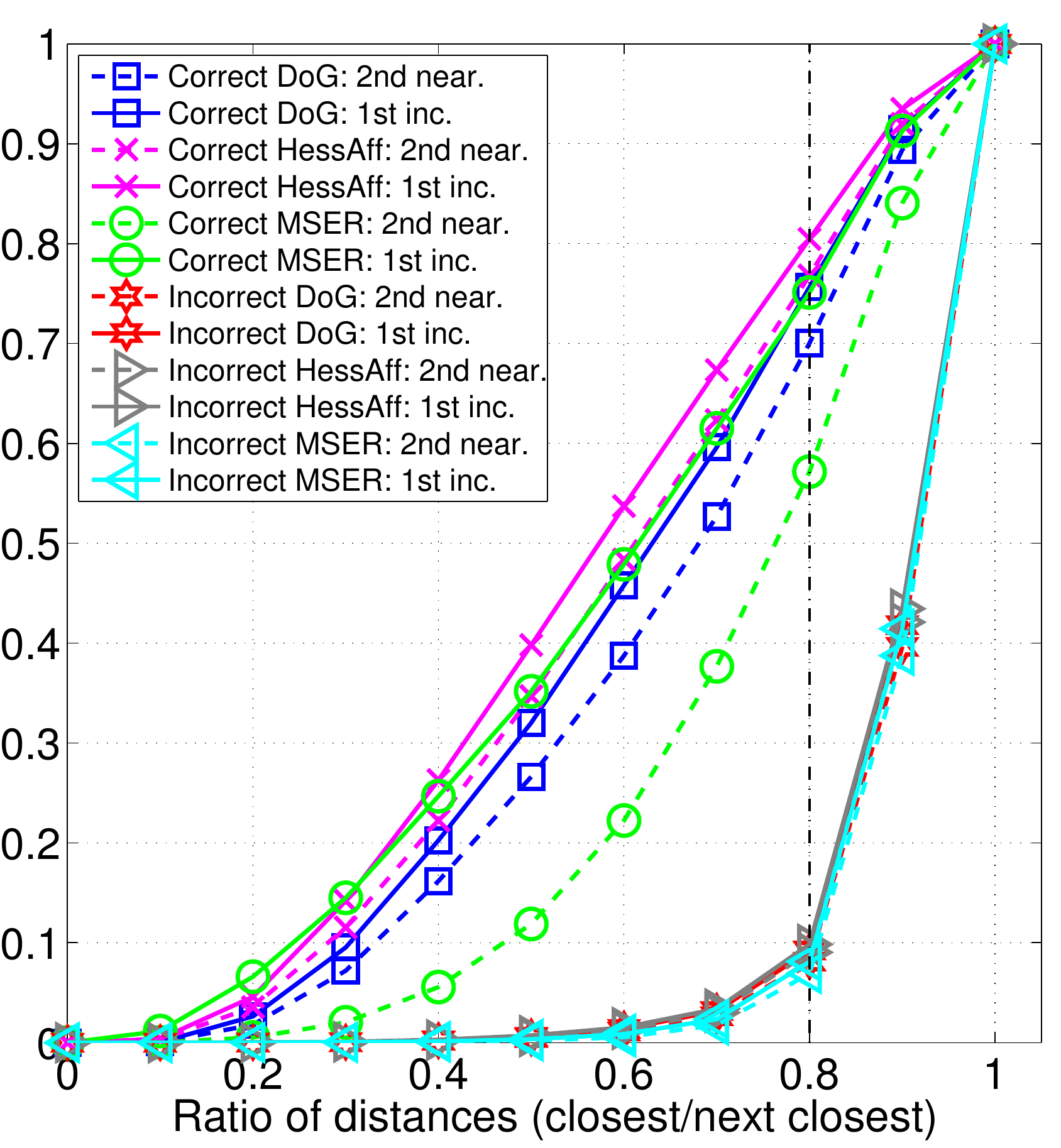}\\
 \caption{CDF. Columns: left -- no view synthesis, right -- with view synthesis.  Rows: upper -- SIFT, lower -- RootSIFT. Average over 50 image pairs from Mikolajczyk~\etal~\cite{Mikolajczyk2005a} and Cordes~\etal~\cite{Cordes2011} datasets. Black dashed line displays standard threshold = 0.8.}
 \label{fig:NN-ratio}
 \end{figure*}

The results for both SIFT and RootSIFT descriptors are shown in
Figure~\ref{fig:NN-ratio}. We see that the DoG and MSER features are
slightly less discriminative than Hessian-Affine. It is also clear from
comparing the left and right columns in Figure~\ref{fig:NN-ratio}, that the
features detected using view synthesis are less distinctive. However, the
distribution of incorrect matches does not change significantly, thus the
thresholds for the new strategy with view synthesis can be kept on the value similar to the
 threshold without view synthesis. The results for the SIFT and
RootSIFT descriptors are also very similar. Therefore, we propose to set
the threshold of the first to first geometrically inconsistent distance
ratio $R$ for the local feature detectors as follows: $R_{\text{MSER}} =
0.85$, $R_{\text{DoG}} = 0.85$ and $R_{\text{HA}} = 0.8$.

\begin{figure*}[p]
\centering
 \includegraphics[width=0.79\linewidth]{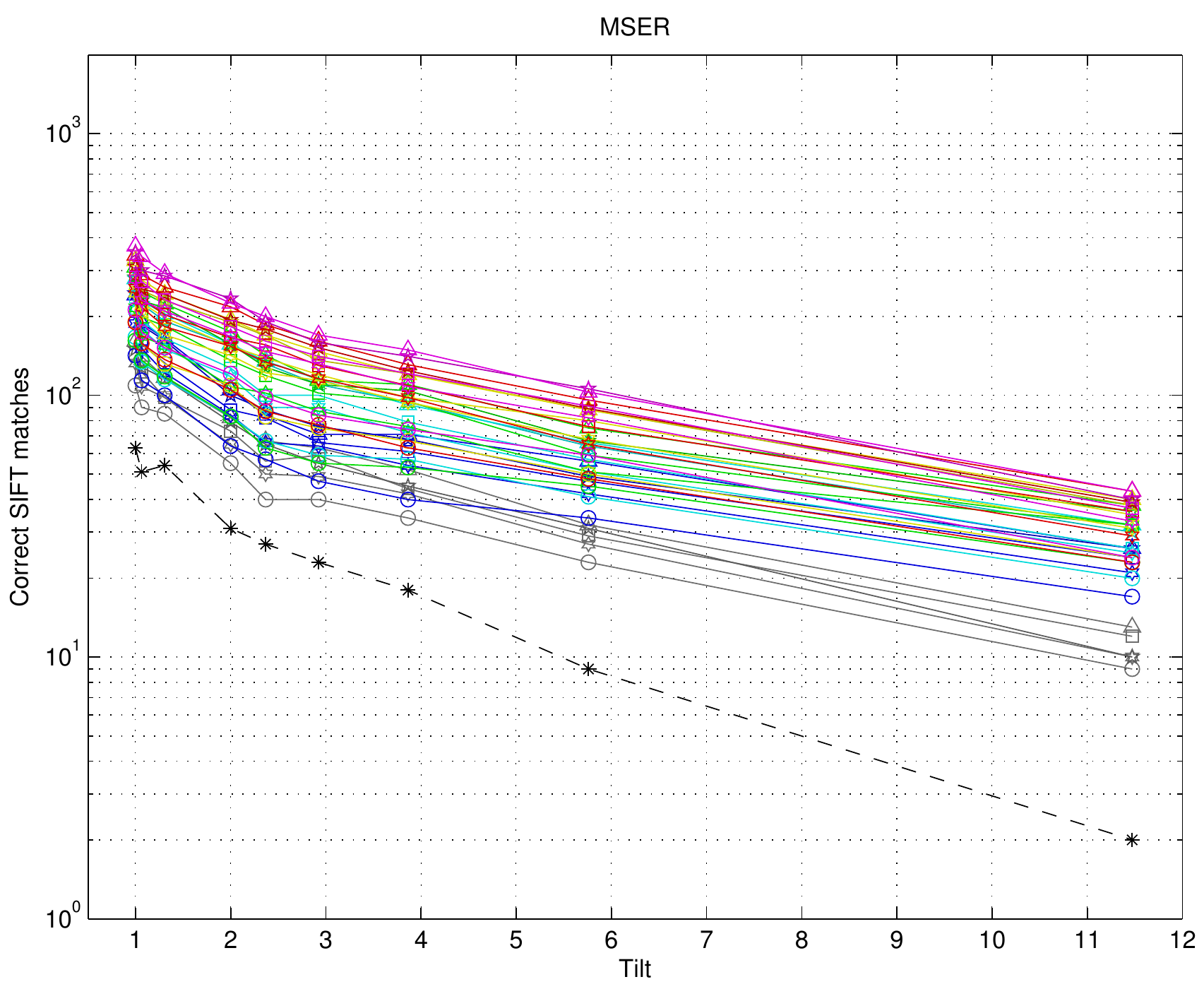}\\
\includegraphics[width=0.79\linewidth]{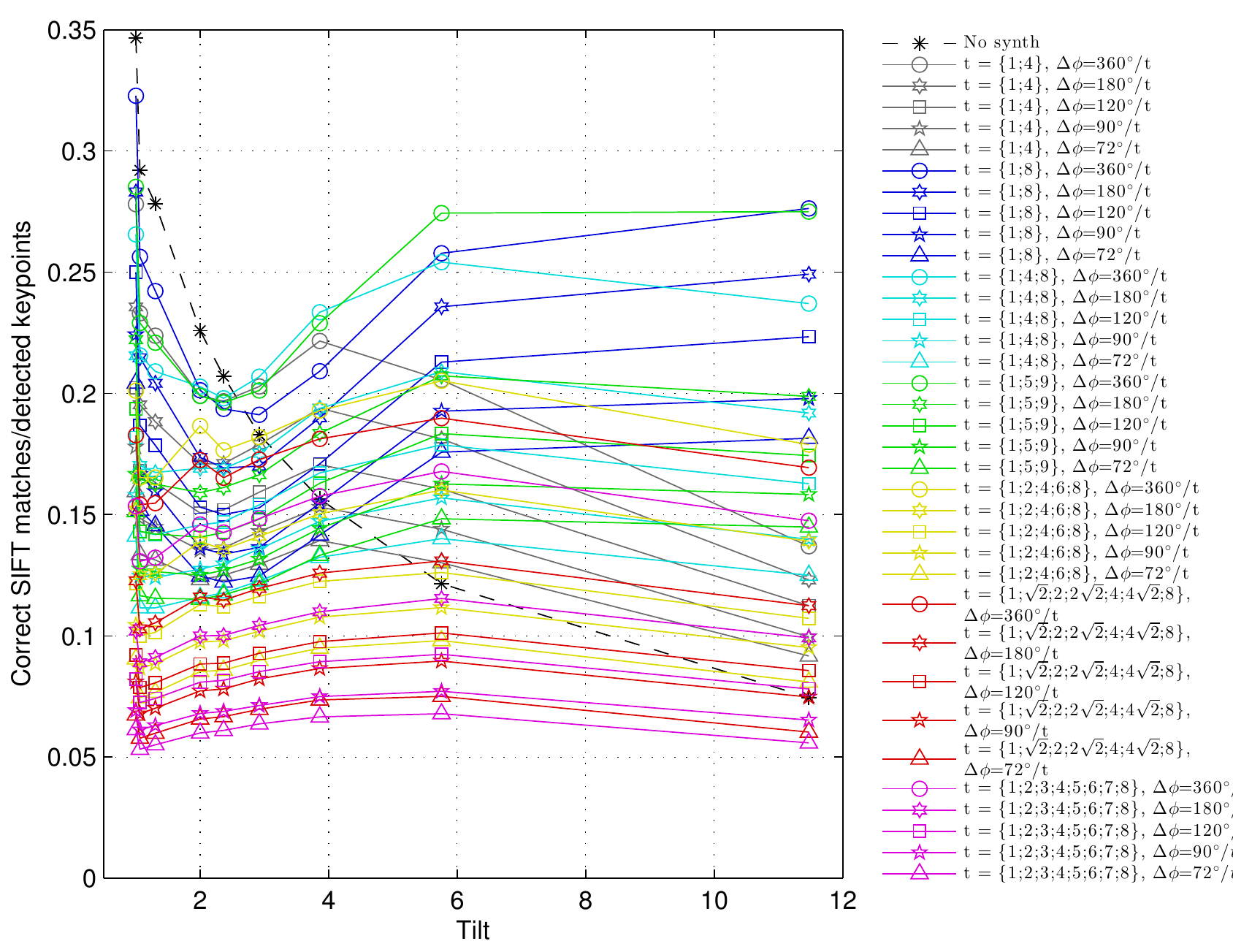}\\
\caption{Comparison of MSER view synthesis configurations on the synthetic dataset.
Upper graph -- the number of correct SIFT matches a robust minimum (value 4\%
quantile) over 100 random images from~\cite{Philbin2007}. Lower graph -- the ratio of the number of correct matches to the number of detected
regions; the mean over 100 random images.}
 \label{fig:tilt-rotation-mser}
 \end{figure*}

\begin{figure*}[p]
\centering
 \includegraphics[width=0.79\linewidth]{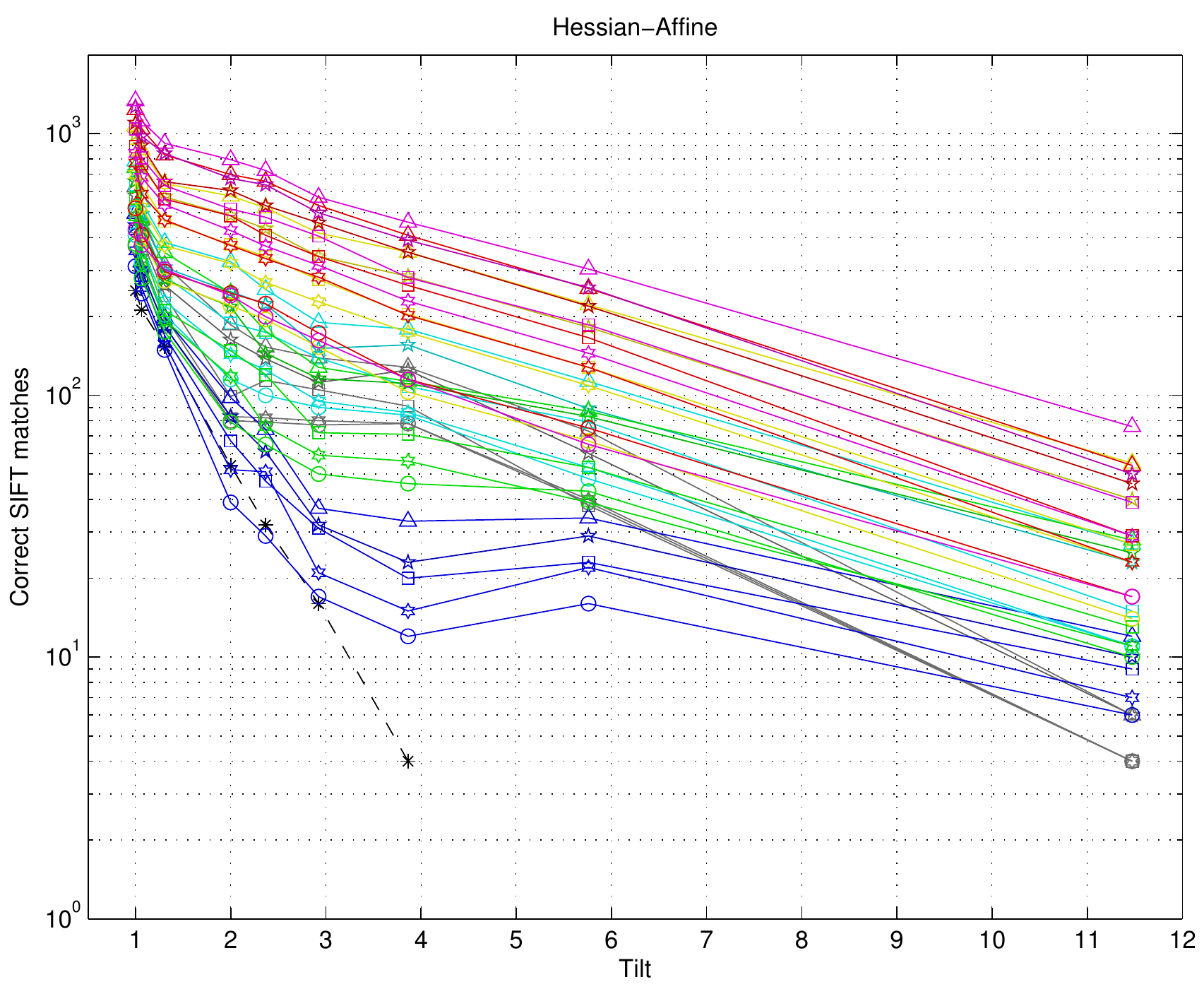}\\
\includegraphics[width=0.79\linewidth]{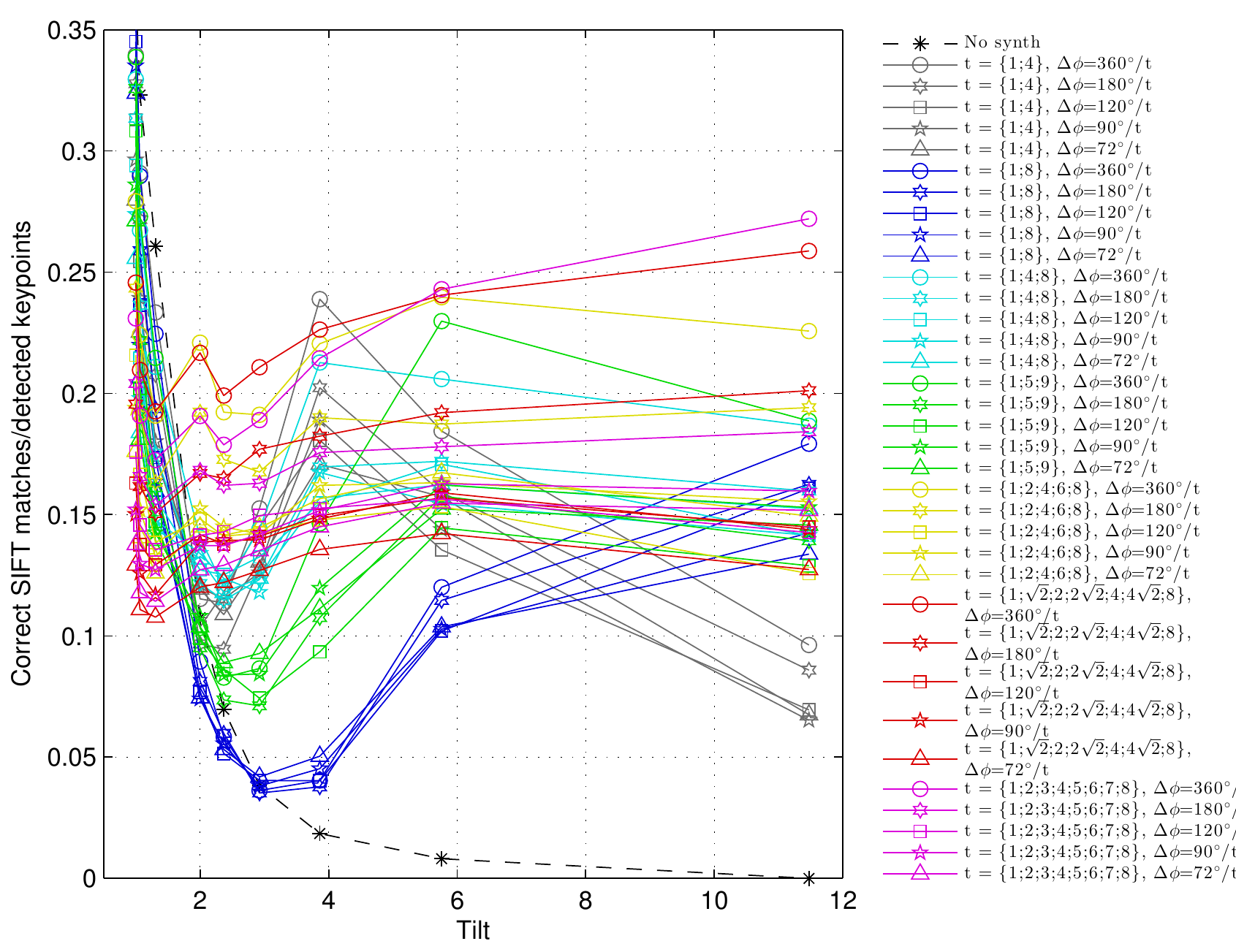}\\
\caption{Comparison of Hessian-Affine view synthesis configurations on the synthetic dataset.
Upper graph -- the number of correct SIFT matches a robust minimum (value 4\%
quantile) over 100 random images from~\cite{Philbin2007}. Lower graph --  the ratio of the number of correct matches to the number of detected
regions; the mean over 100 random images.}
 \label{fig:tilt-rotation-hess}
 \end{figure*}

\begin{figure*}[p]
\centering
 \includegraphics[width=0.79\linewidth]{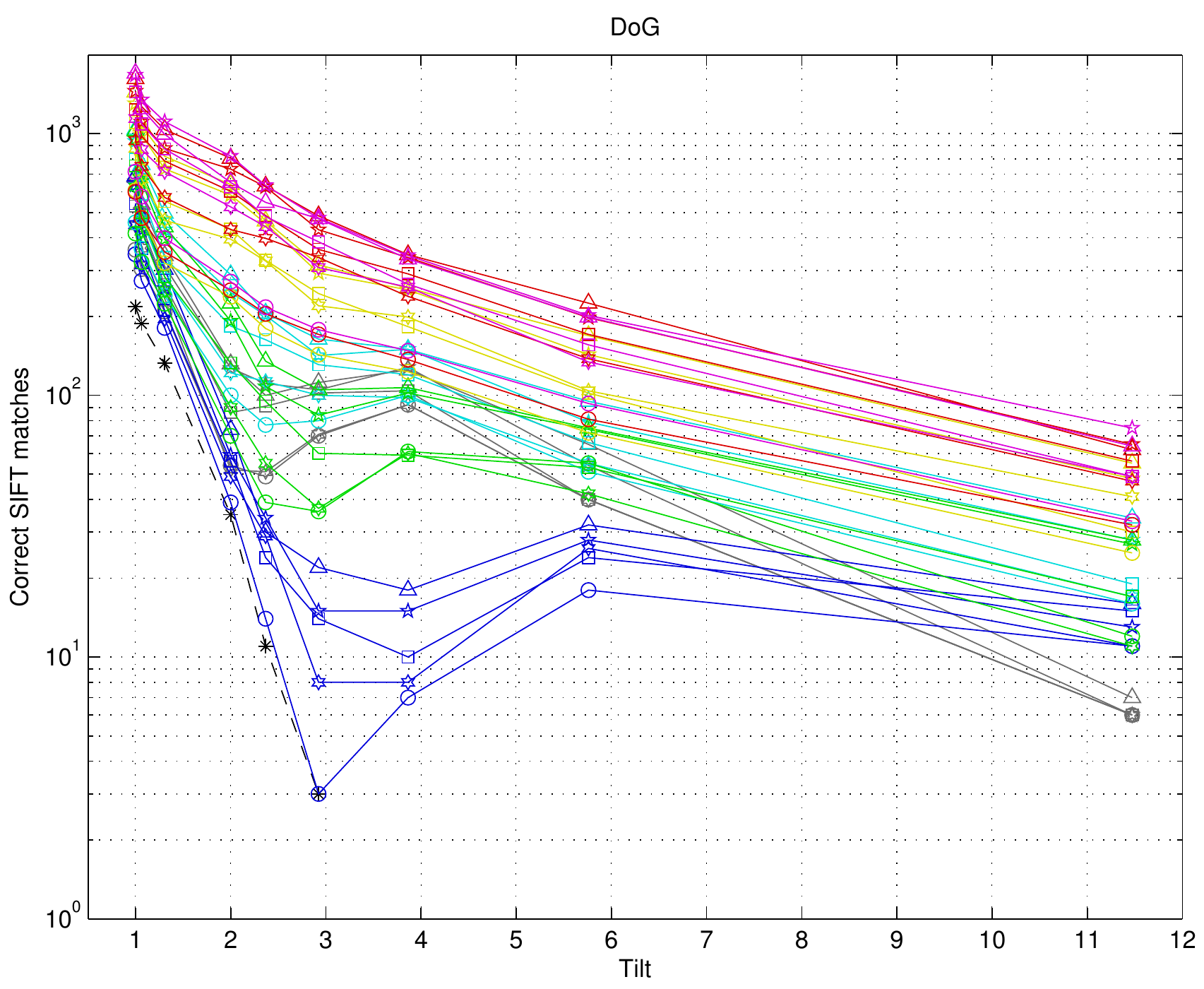}\\
\includegraphics[width=0.79\linewidth]{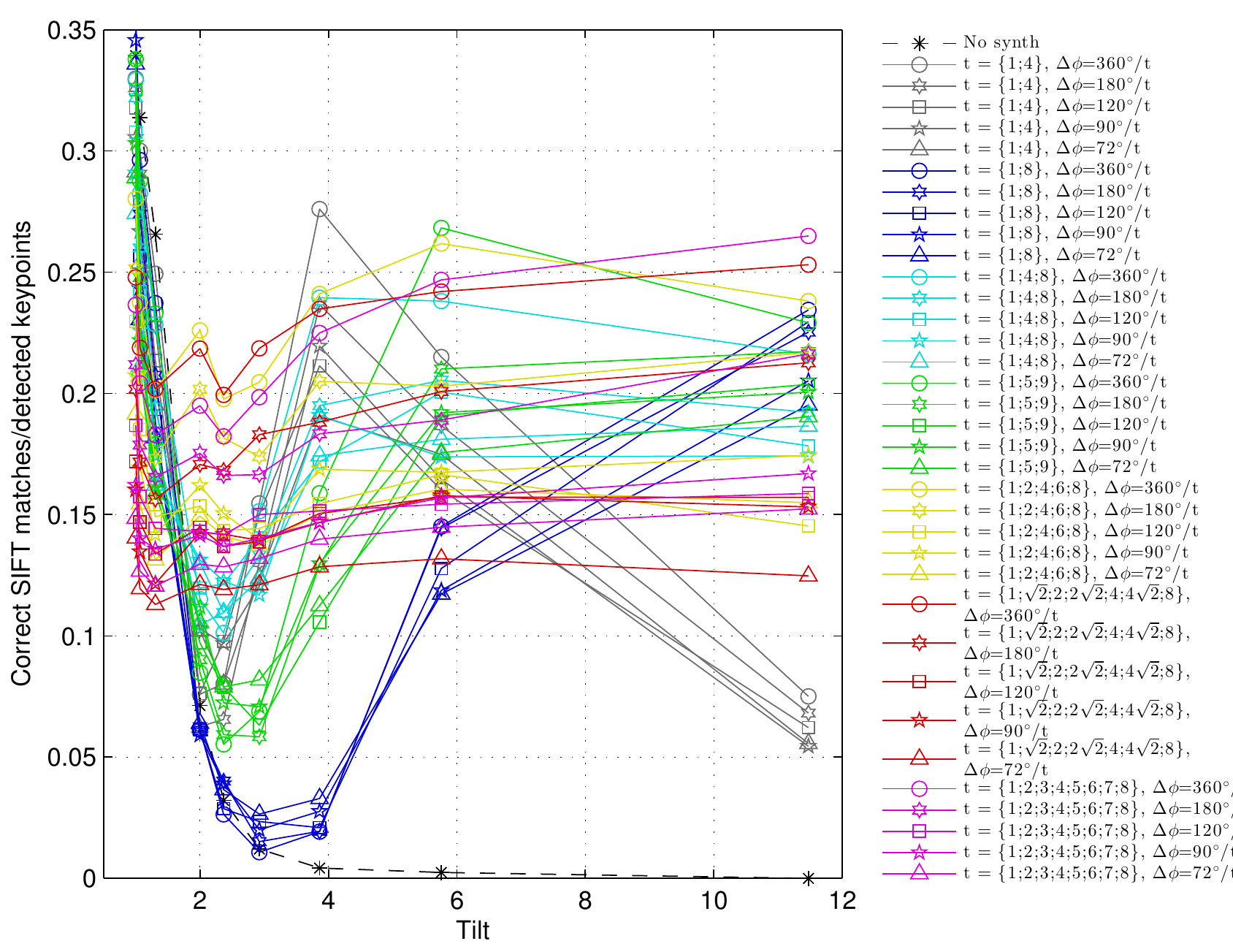}\\
\caption{Comparison of DoG view synthesis configurations on the synthetic dataset. Upper graph -- the number of correct SIFT matches a robust minimum (value 4\%
quantile) over 100 random images from~\cite{Philbin2007}. Lower graph -- the ratio of the number of correct matches to the number of detected
regions; the mean over 100 random images.}
 \label{fig:tilt-rotation-sift}
 \end{figure*}

\paragraph{Tilt set and latitude sampling step.}
The first two parameters of the view synthesis, tilt \{t\} sampling and latitude
step $\Delta\phi_{\text{base}}$, were explored in the following synthetic
experiment.
For each of 100 random images from Oxford Building Dataset\footnote{available at http://www.robots.ox.ac.uk/$\sim$vgg/data/oxbuildings/}~\cite{Philbin2007},
a set of simulated views with latitudes angles $\theta=($0, 20, 40, 60, 65,
70, 75, 80, 85$) \degree$, corresponding to tilt series $t=($1.00, 1.06,
1.30, 2.00, 2.36, 2.92, 3.86,
5.75, 11.47$)$\footnote{assuming that the original image is in the
fronto-parallel view} was generated. The reference image have been convolved with a Gaussian filter with $\sigma_H=0.8$ along horizontal direction and $\sigma_V=0.8t$ along vertical direction and finally shrunk in vertical direction by $t$. The ground truth affine matrix $A$ was
computed for each synthetic view using
equation~(\ref{eq:affine-matrix-decomposition}) and used in the final
verification step of the MODS algorithm. 
The various configurations of the view synthesis were tested and results for the selected configurations are shown in Figures~\ref{fig:tilt-rotation-mser} -- \ref{fig:tilt-rotation-sift}. Note that the view synthesis significantly increases the matching performance, however after reaching some density of the view-sphere sampling additional views does not bring more correspondences. MSER and Hessian-Affine need sparser view-sphere sampling than DoG. 

Two configurations, {\sc Sparse} and {\sc Dense}, were chosen for each detector
(see Table~\ref{tabular:view-synth-configuration}) using the following criteria:
efficiency -- the ratio of correct matches per detected region,  matching
performance -- the number of unique (non-duplicated) matches on the synthetic
image pairs with $85\degree$ out of plane rotation. The {\sc Sparse}
configuration is fast but still able to solve synthetic image pairs with
up to $85\degree$ out of plane rotation. The {\sc Dense} configuration generates sufficient number of correspondences for the most image pairs in the EVD dataset for each detector.

\paragraph{Image pre-smoothing.} 
The early experiments with view synthesis, have shown that the amount of
image smoothing $\sigma_{\text{base}}$ prior to view synthesis affects
matching performance significantly. Values too small fail to prevent
aliasing, values too high oversmooth the image reducing the number of
detected regions.  Unlike MSER, the scale-space based detectors -- DoG,
Hessian-Affine apply pre-smoothing as the initial step of the scale-space
pyramid.

\begin{figure*}[tpb]
\centering
\includegraphics[width=0.51\linewidth]{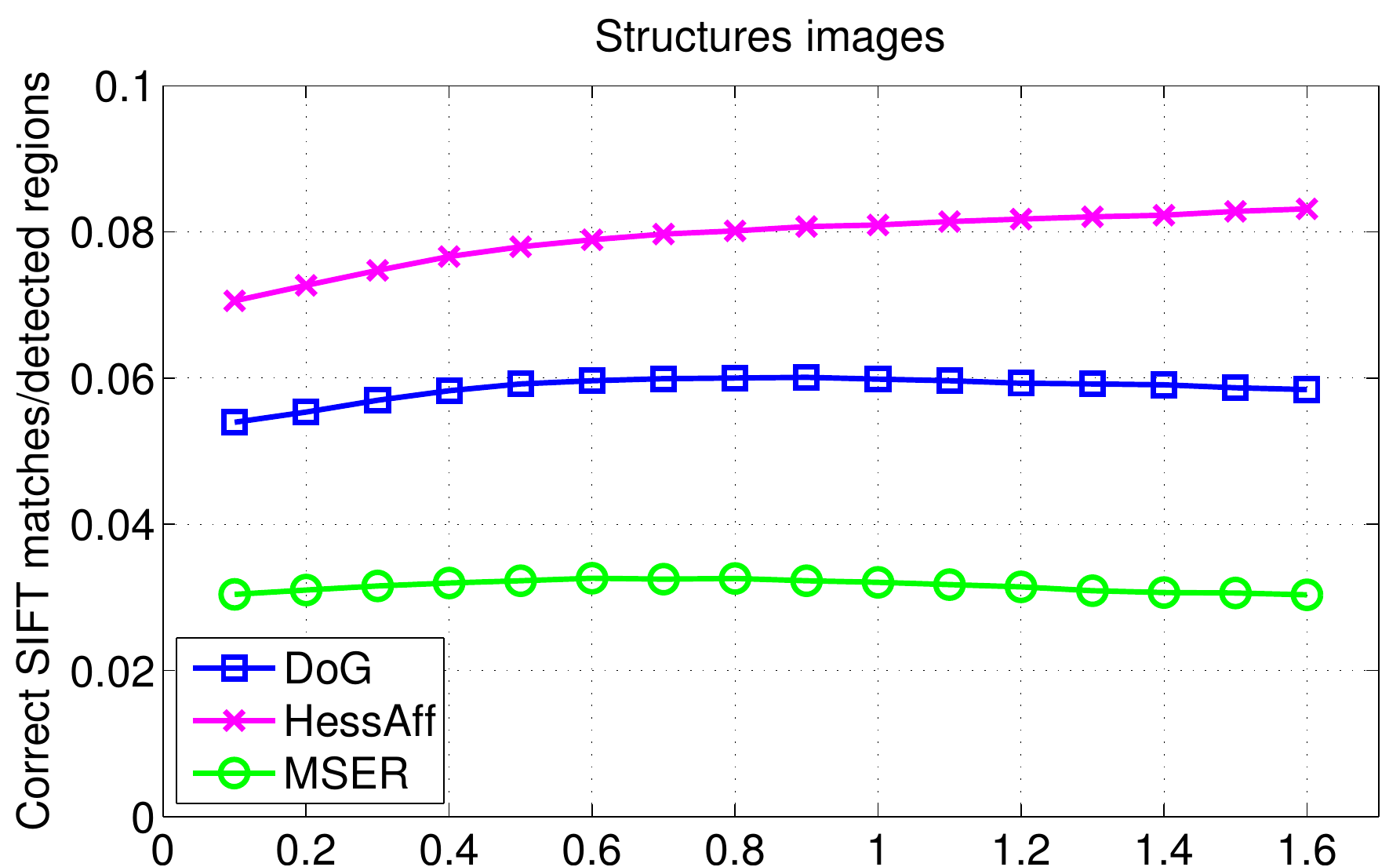}
\includegraphics[width=0.48\linewidth]{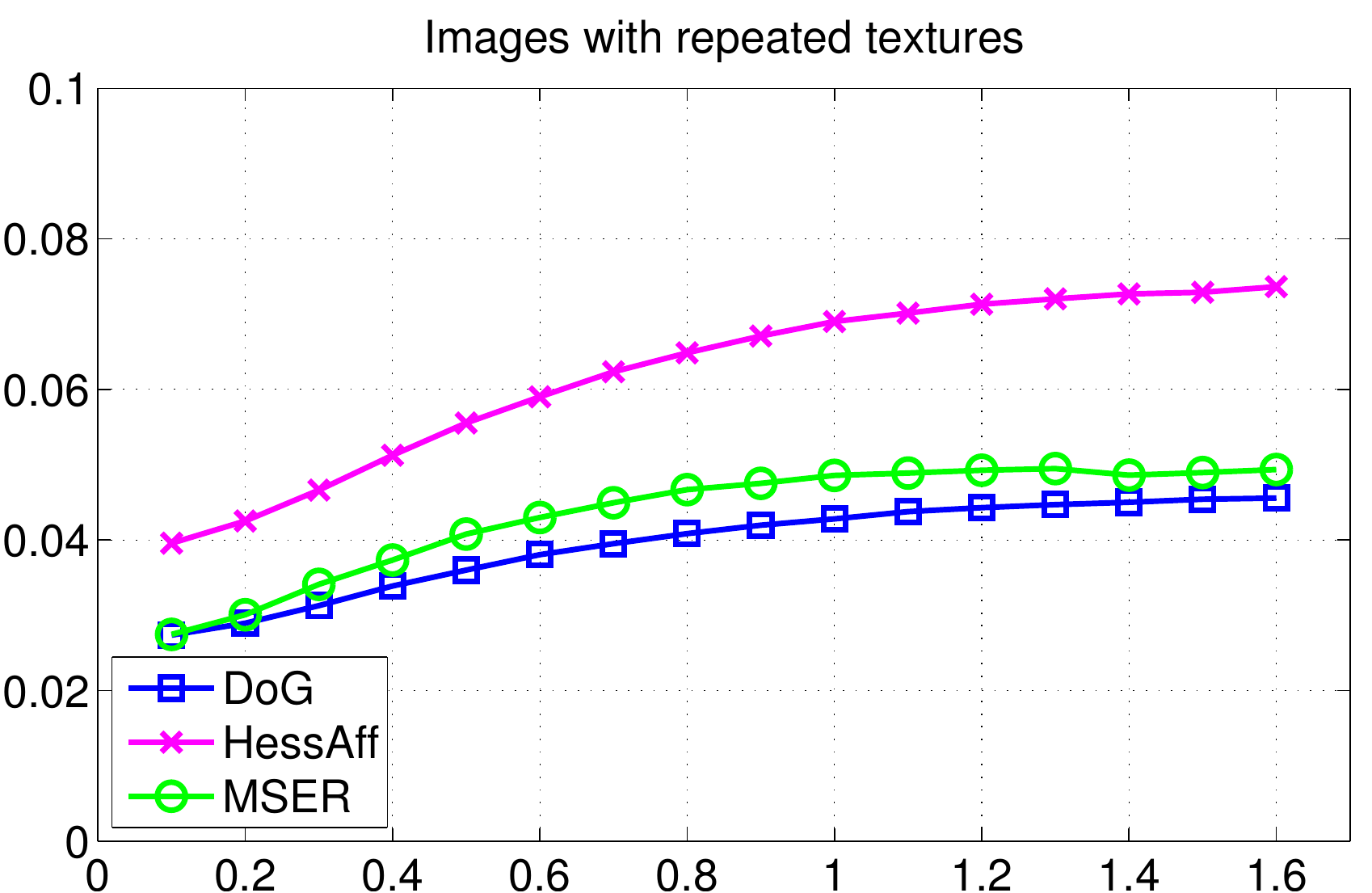}\\
\includegraphics[width=0.51\linewidth]{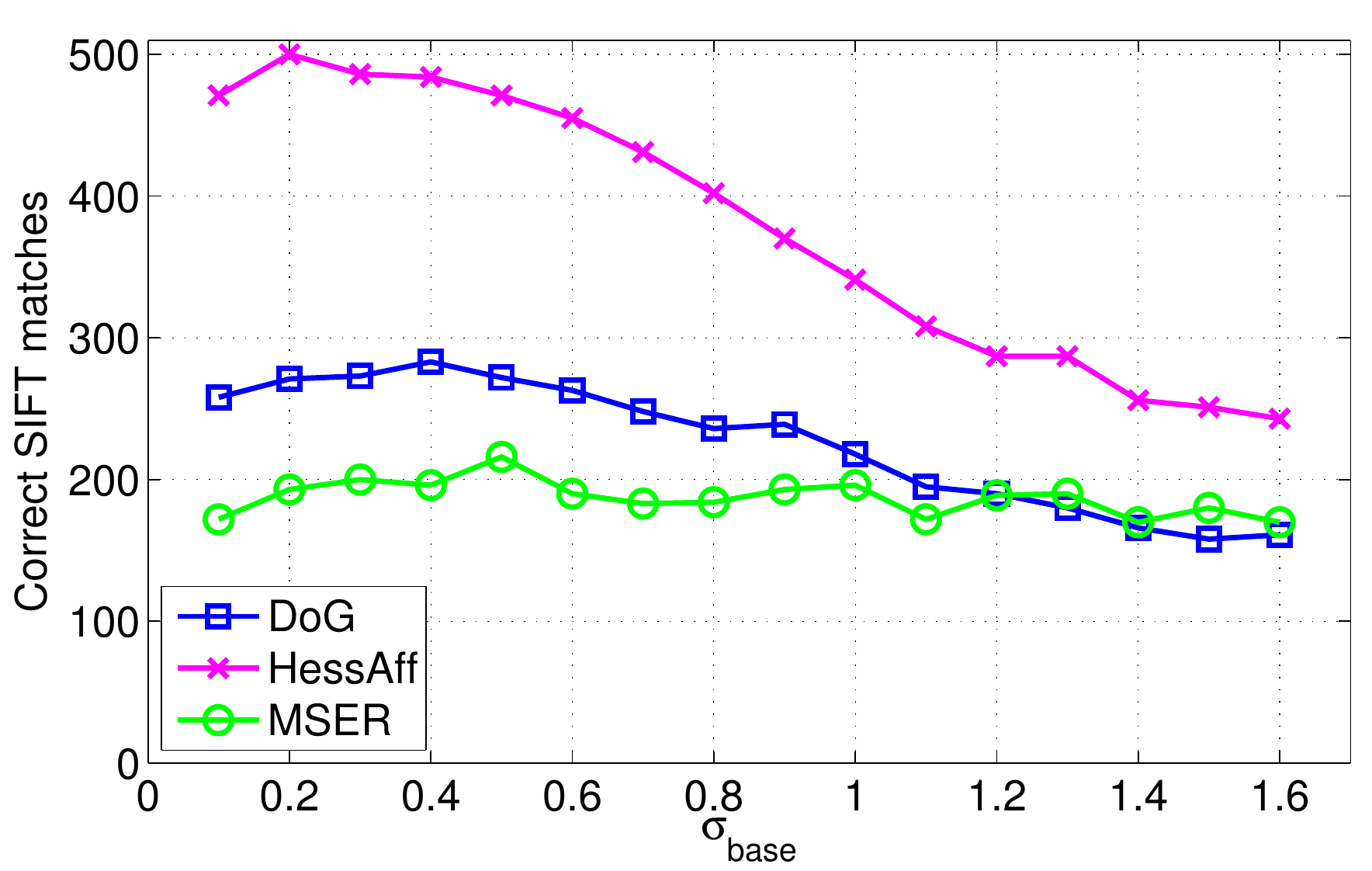} 
\includegraphics[width=0.48\linewidth]{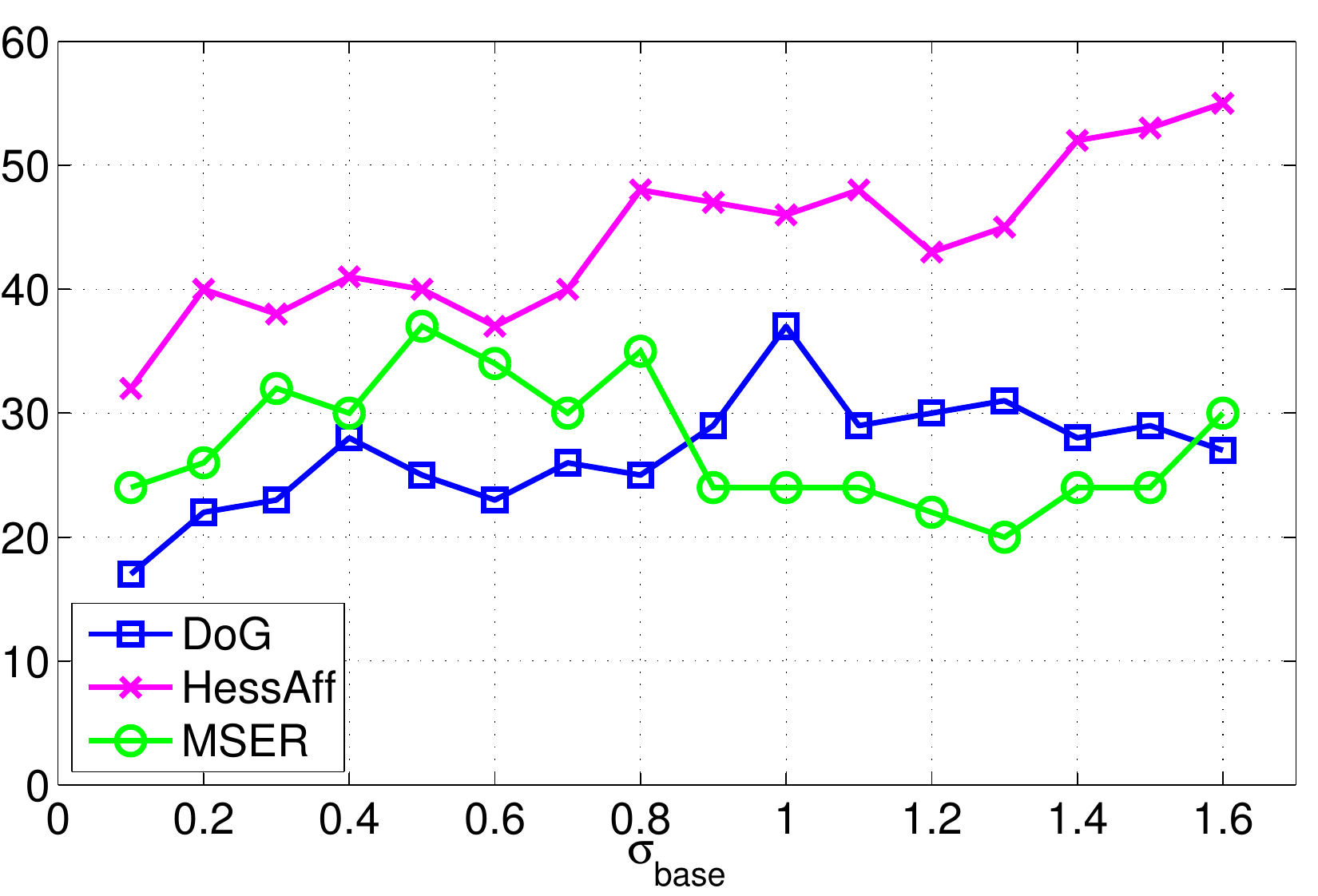} \\
\caption{Matching with view synthesis ({\sc Dense} configuration) using different image pre-smoothing factor $\sigma_{base}$.  Rows: upper -- ratio of correct SIFT matches to number of detected regions, lower -- number of correct SIFT matches -- robust minimum (value 4\% quantile). Columns: left -- structured images, right -- images with repeated patterns.}
 \label{fig:sigma-establishing}
 \end{figure*}

This experiment measures the effect of the pre-smoothing parameter
$\sigma_{\text{base}}$ on the matching performance of different detectors.
The range of values of the $\sigma_{\text{base}}$ were used in matching of 35 image pairs of the publicly available datasets~\cite{Mikolajczyk2005a} and~\cite{Cordes2011}. We have divided all pairs into two sets "Structured images" -- scenes {\sc graf, grace, posters, there, underground} (25 image pairs in total) from~\cite{Cordes2011} and "Images with repeated textures" -- scenes {\sc wall, colors} (10 image pairs in total)~\cite{Cordes2011},~\cite{Mikolajczyk2005a}. The {\sc Dense} configurations of the view synthesis were chosen for each of the detectors (see Table~\ref{tabular:view-synth-configuration}). Based on this experiment (see Figure~\ref{fig:sigma-establishing}), we have set following parameters for image pre-smoothing in the MODS algorithm: $\sigma_{\text{base}}=0.8,0.2$, and 0.4 for the MSER, Hessian-Affine and DoG detectors, respectively.

\section{Full version of the experiments on the EVD dataset}
\label{sec:appendixB}
The full version of experimental evaluation of the matching with view synthesis algorithm on EVD dataset is presented in this section. For this very challenging dataset it is hard to obtain ground truth homographies from the manually selected correspondences. Therefore, the ground truth homography matrices were estimated by running LO-RANSAC on correspondences of all three detectors with the view synthesis configuration $\{t\} = \{1;\sqrt{2};2;2\sqrt{2};4;4\sqrt{2};8\}$,
$\Delta\phi=72\degree{}/t$. The number of inliers for each image pair was $\geq$ 50 and the homographies were manually inspected. For the image pairs {\sc graf}
and {\sc there} precise homographies were provided by Cordes~\etal~\cite{Cordes2011}. The transition tilts were computed using equation~(\ref{eq:affine-matrix-decomposition}) with SVD decomposition of the linearised homography at the center of the first image of the pair.
The configurations of detectors evaluated are listed in Table~\ref{tabular:view-synth-configuration}, additionally, the performance of the MODS and MSER detector with scale synthesis were compared. The configuration for MODS algorithm is shown in Table~\ref{tabular:MODS-configuration}. The MODS algorithm allows to set the minimum desired number of inliers as a stopping criterion. We set the threshold to 15 inliers, since fifteen inliers to a homography (with duplicate matches removed) have very low probability of being accidental and yet allow to demonstrate the speed  gain of the algorithm. 

The results for all configurations for all detectors are shown in Tables~\ref{tabular:test-optimal-nosynth-mser-noscale} -- \ref{tabular:test-MODS-rootsift}. For comparison, ASIFT\footnote{Reference code from http://demo.ipol.im/demo/my\_affine\_sift} results were added. The timing measurements are reported for single, two and four  cores of the Intel i5 CPU @ 2.6GHz processor with 4GB RAM. 

\begin{figure*}[htbp]
\centering
\includegraphics[width=0.50\linewidth]{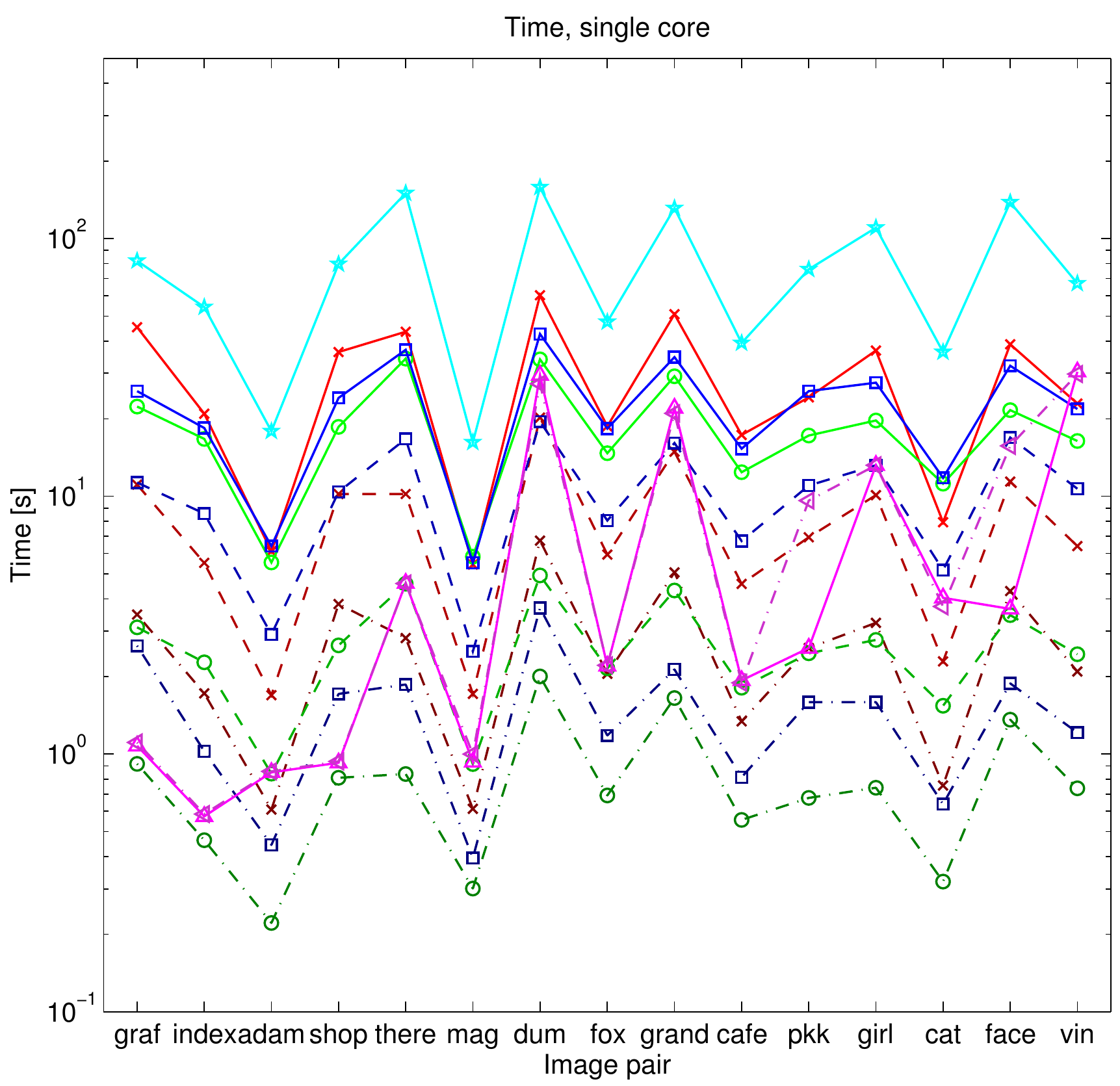}
\includegraphics[width=0.48\linewidth]{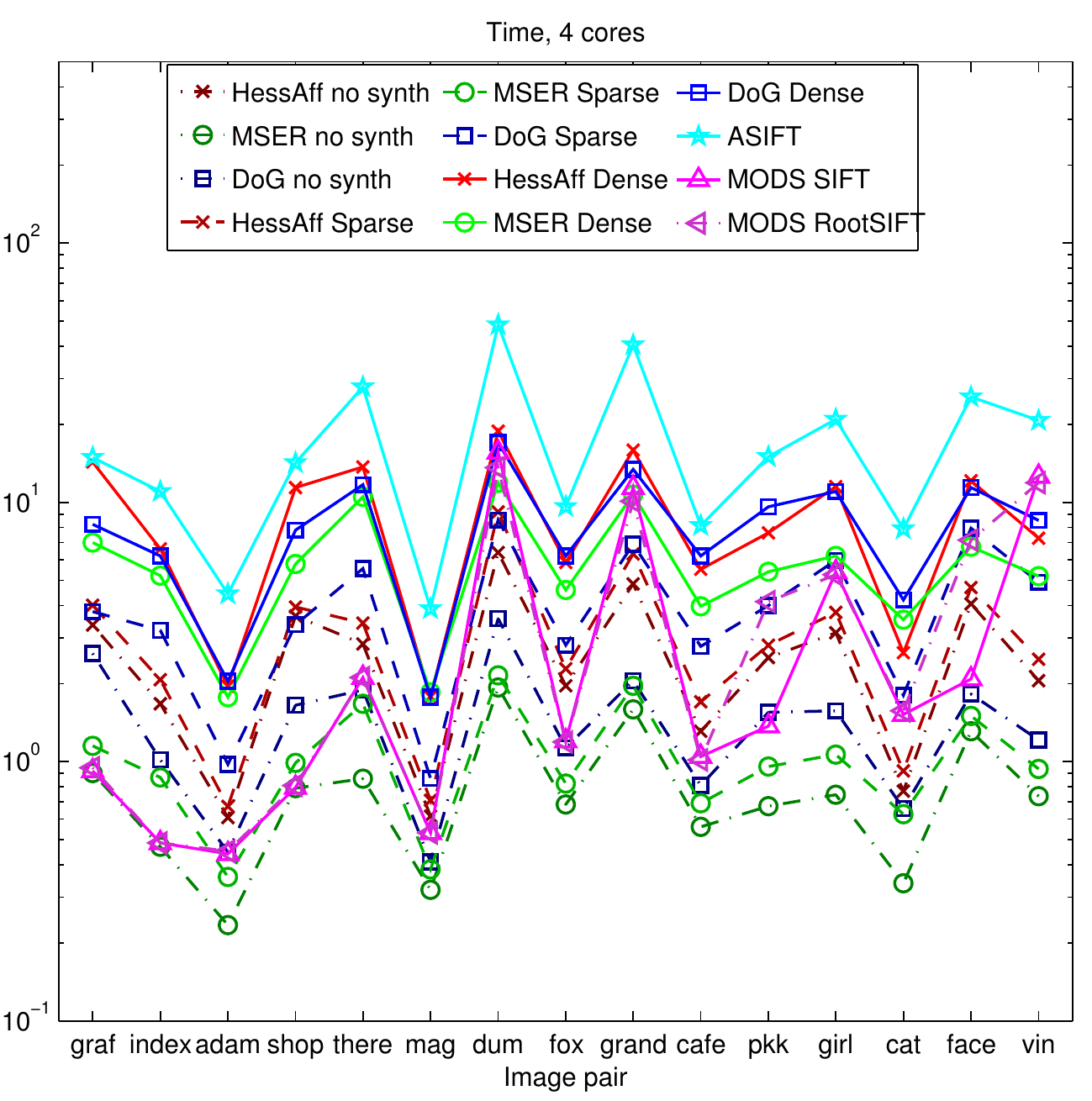}
\caption{Running time of the different view synthesis configurations (Table~\ref{tabular:view-synth-configuration}). 
Left -- 1 core, right -- 4 cores.} 
\label{fig:time-bars}
\end{figure*} 
\clearpage
\renewcommand{\tabcolsep}{3pt}
\begin{table}[htpb]
\centering
\begin{minipage}[p]{0.75\linewidth}
\caption{Performance on the EVD dataset. MSER, no view synthesis. Results with less than 8 correct inliers are in red. }
\label{tabular:test-optimal-nosynth-mser-noscale}
\footnotesize
\centering
\begin{tabular}{|=c|...|=r+r.|=r+r.|=r+r.|=r+r.|=r+r|}
\hline
Image 
& \multicolumn{17}{c|}{MSER, no synths.} \\
\hline
& \multicolumn{3}{c|}{Time}
& \multicolumn{6}{c|}{LO-RANSAC}
& \multicolumn{6}{c|}{Tentatives quality }
& \multicolumn{2}{c|}{Regions}\\
\hline
&
&&& \multicolumn{3}{c|}{SIFT}
& \multicolumn{3}{c|}{RootSIFT}
& \multicolumn{3}{c|}{SIFT}
& \multicolumn{3}{c|}{RootSIFT}
& &\\
&\begin{sideways}1 core $[s]$\end{sideways}
&\begin{sideways}2 cores $[s]$\end{sideways}
&\begin{sideways}4 cores $[s]$\end{sideways}
&\begin{sideways}Correct inliers\end{sideways} 
&\begin{sideways} Inliers\end{sideways}
&\begin{sideways}Correct / all $[\%]$\end{sideways} 
&\begin{sideways}Correct inliers\end{sideways} 
&\begin{sideways} Inliers\end{sideways}
&\begin{sideways}Correct / all $[\%]$\end{sideways} 
&\begin{sideways}Correct matches\end{sideways} 
&\begin{sideways}Tentatives\end{sideways}
&\begin{sideways}Correct  / all $[\%]$\end{sideways}
&\begin{sideways}Correct matches\end{sideways} 
&\begin{sideways}Tentatives\end{sideways}
&\begin{sideways}Correct / all $[\%]$\end{sideways}
&\begin{sideways}Image 1\end{sideways}
&\begin{sideways}Image 2\end{sideways}\\
\hline
 graf&0.8&0.8&0.8&65&68&95.6&67&70&95.7&65&121&53.7&67&110&60.9&804&1432\\ 
index&0.4&0.4&0.4&18&19&94.7&17&18&94.4&18&44&40.9&17&36&47.2&291&172\\ 
shop&0.7&0.7&0.7&18&19&94.7&18&18&100&18&65&27.7&18&40&45&1131&646\\ 
adam&0.1&0.1&0.1&\rowstyle{\color{red}}0&\rowstyle{\color{red}}0&\rowstyle{\color{red}}0&\rowstyle{\color{red}}0&\rowstyle{\color{red}}0&\rowstyle{\color{red}}0&\rowstyle{\color{red}}6&\rowstyle{\color{red}}15&\rowstyle{\color{red}}40&\rowstyle{\color{red}}7&\rowstyle{\color{red}}16&\rowstyle{\color{red}}43.8&118&48\\
\hline
\end{tabular}
\end{minipage}
\end{table}

\begin{table}[b]
\centering
\begin{minipage}[h]{0.8\linewidth}
\caption{Performance on the EVD dataset. MSER, scale view synthesis only. Results with less than 8 correct inliers are in red.}
\label{tabular:test-optimal-nosynth-mser}
\footnotesize
\centering
\begin{tabular}{|=c|...|=r+r.|=r+r.|=r+r.|=r+r.|=r+r|}
\hline
Image 
& \multicolumn{17}{c|}{MSER, 2 scale synths. Total image area  $A_{total}=1.08A_{orig}$} \\
\hline
& \multicolumn{3}{c|}{Time}
& \multicolumn{6}{c|}{LO-RANSAC}
& \multicolumn{6}{c|}{Tentatives quality }
& \multicolumn{2}{c|}{Regions}\\
\hline
&
&&& \multicolumn{3}{c|}{SIFT}
& \multicolumn{3}{c|}{RootSIFT}
& \multicolumn{3}{c|}{SIFT}
& \multicolumn{3}{c|}{RootSIFT}
& &\\
&\begin{sideways}1 core $[s]$\end{sideways}
&\begin{sideways}2 cores $[s]$\end{sideways}
&\begin{sideways}4 cores $[s]$\end{sideways}
&\begin{sideways}Correct inliers\end{sideways} 
&\begin{sideways} Inliers\end{sideways}
&\begin{sideways}Correct / all $[\%]$\end{sideways} 
&\begin{sideways}Correct inliers\end{sideways} 
&\begin{sideways} Inliers\end{sideways}
&\begin{sideways}Correct / all $[\%]$\end{sideways} 
&\begin{sideways}Correct matches\end{sideways} 
&\begin{sideways}Tentatives\end{sideways}
&\begin{sideways}Correct  / all $[\%]$\end{sideways}
&\begin{sideways}Correct matches\end{sideways} 
&\begin{sideways}Tentatives\end{sideways}
&\begin{sideways}Correct / all $[\%]$\end{sideways}
&\begin{sideways}Image 1\end{sideways}
&\begin{sideways}Image 2\end{sideways}\\
\hline
 graf&1&0.8&0.8&81&85&95.3&82&87&94.3&82&160&51.2&83&154&53.9&1018&1674\\ 
index&0.5&0.4&0.4&19&20&95&18&20&90&19&56&33.9&18&42&42.9&411&246\\ 
shop&0.8&0.7&0.7&28&30&93.3&29&31&93.5&28&84&33.3&29&61&47.5&1321&711\\ 
adam&0.2&0.1&0.1&\rowstyle{\color{red}}0&\rowstyle{\color{red}}0&\rowstyle{\color{red}}0&8&8&100&8&18&44.4&9&19&47.4&135&62\\
\hline
\end{tabular}
\end{minipage}
\end{table}

\begin{table*}[p]
\centering
\begin{minipage}[h]{0.75\linewidth}
\caption{Performance on the EVD dataset.Hessian-Affine, no view synthesis. Results with less than 8 correct inliers are in red.}
\label{tabular:test-optimal-nosynth-hess}
\footnotesize
\centering
\begin{tabular}{|=c|...|=r+r.|=r+r.|=r+r.|=r+r.|=r+r|}
\hline
Image 
& \multicolumn{17}{c|}{HessAff, no synths.} \\
\hline
& \multicolumn{3}{c|}{Time}
& \multicolumn{6}{c|}{LO-RANSAC}
& \multicolumn{6}{c|}{Tentatives quality }
& \multicolumn{2}{c|}{Regions}\\
\hline
&
&&& \multicolumn{3}{c|}{SIFT}
& \multicolumn{3}{c|}{RootSIFT}
& \multicolumn{3}{c|}{SIFT}
& \multicolumn{3}{c|}{RootSIFT}
& &\\
&\begin{sideways}1 core $[s]$\end{sideways}
&\begin{sideways}2 cores $[s]$\end{sideways}
&\begin{sideways}4 cores $[s]$\end{sideways}
&\begin{sideways}Correct inliers\end{sideways} 
&\begin{sideways} Inliers\end{sideways}
&\begin{sideways}Correct / all $[\%]$\end{sideways} 
&\begin{sideways}Correct inliers\end{sideways} 
&\begin{sideways} Inliers\end{sideways}
&\begin{sideways}Correct / all $[\%]$\end{sideways} 
&\begin{sideways}Correct matches\end{sideways} 
&\begin{sideways}Tentatives\end{sideways}
&\begin{sideways}Correct  / all $[\%]$\end{sideways}
&\begin{sideways}Correct matches\end{sideways} 
&\begin{sideways}Tentatives\end{sideways}
&\begin{sideways}Correct / all $[\%]$\end{sideways}
&\begin{sideways}Image 1\end{sideways}
&\begin{sideways}Image 2\end{sideways}\\
\hline
 graf&3.4&3.3&3.3&14&17&82.4&16&19&84.2&15&141&10.6&19&97&19.6&3630&4614\\ 
index&1.6&1.6&1.6&\rowstyle{\color{red}}0&\rowstyle{\color{red}}0&\rowstyle{\color{red}}0&\rowstyle{\color{red}}0&\rowstyle{\color{red}}0&\rowstyle{\color{red}}0&\rowstyle{\color{red}}0&\rowstyle{\color{red}}114&\rowstyle{\color{red}}0&\rowstyle{\color{red}}1&\rowstyle{\color{red}}79&\rowstyle{\color{red}}1.3&2188&874\\ 
shop&3.7&3.6&3.5&\rowstyle{\color{red}}0&\rowstyle{\color{red}}0&\rowstyle{\color{red}}0&\rowstyle{\color{red}}0&\rowstyle{\color{red}}0&\rowstyle{\color{red}}0&\rowstyle{\color{red}}0&\rowstyle{\color{red}}78&\rowstyle{\color{red}}0&\rowstyle{\color{red}}0&\rowstyle{\color{red}}39&\rowstyle{\color{red}}0&5675&2657\\
\hline
\end{tabular}
\end{minipage}
\end{table*}

\begin{table*}[p]
\centering
\begin{minipage}[h]{0.7\linewidth}
\caption{Performance on the EVD dataset. DoG, no view synthesis. Results with less than 8 correct inliers are in red.}
\label{tabular:test-optimal-nosynth-sift}
\footnotesize
\centering
\begin{tabular}{|=c|...|=r+r.|=r+r.|=r+r.|=r+r.|=r+r|}
\hline
Image 
& \multicolumn{17}{c|}{DoG, no synths.} \\
\hline
& \multicolumn{3}{c|}{Time}
& \multicolumn{6}{c|}{LO-RANSAC}
& \multicolumn{6}{c|}{Tentatives quality }
& \multicolumn{2}{c|}{Regions}\\
\hline
&
&&& \multicolumn{3}{c|}{SIFT}
& \multicolumn{3}{c|}{RootSIFT}
& \multicolumn{3}{c|}{SIFT}
& \multicolumn{3}{c|}{RootSIFT}
& &\\
&\begin{sideways}1 core $[s]$\end{sideways}
&\begin{sideways}2 cores $[s]$\end{sideways}
&\begin{sideways}4 cores $[s]$\end{sideways}
&\begin{sideways}Correct inliers\end{sideways} 
&\begin{sideways} Inliers\end{sideways}
&\begin{sideways}Correct / all $[\%]$\end{sideways} 
&\begin{sideways}Correct inliers\end{sideways} 
&\begin{sideways} Inliers\end{sideways}
&\begin{sideways}Correct / all $[\%]$\end{sideways} 
&\begin{sideways}Correct matches\end{sideways} 
&\begin{sideways}Tentatives\end{sideways}
&\begin{sideways}Correct  / all $[\%]$\end{sideways}
&\begin{sideways}Correct matches\end{sideways} 
&\begin{sideways}Tentatives\end{sideways}
&\begin{sideways}Correct / all $[\%]$\end{sideways}
&\begin{sideways}Image 1\end{sideways}
&\begin{sideways}Image 2\end{sideways}\\
\hline
 graf&2.5&2.5&2.5&\rowstyle{\color{red}}0&\rowstyle{\color{red}}0&\rowstyle{\color{red}}0&\rowstyle{\color{red}}0&\rowstyle{\color{red}}0&\rowstyle{\color{red}}0&\rowstyle{\color{red}}3&\rowstyle{\color{red}}120&\rowstyle{\color{red}}2.5&\rowstyle{\color{red}}4&\rowstyle{\color{red}}83&\rowstyle{\color{red}}4.8&1682&2419\\ 
index&0.9&0.9&0.9&\rowstyle{\color{red}}0&\rowstyle{\color{red}}0&\rowstyle{\color{red}}0&\rowstyle{\color{red}}0&\rowstyle{\color{red}}0&\rowstyle{\color{red}}0&\rowstyle{\color{red}}0&\rowstyle{\color{red}}106&\rowstyle{\color{red}}0&\rowstyle{\color{red}}0&\rowstyle{\color{red}}80&\rowstyle{\color{red}}0&1171&516\\
\hline
\end{tabular}
\end{minipage}
\end{table*}
\begin{table*}[p]
\centering
\begin{minipage}[h]{0.85\linewidth}
\centering
\caption{Performance on the EVD dataset.MSER,  {\sc Sparse} configuration. Results with less than 8 correct inliers are in red.}
\label{tabular:test-optimal-fast-mser}
\footnotesize
\begin{tabular}{|=c|...|=r+r.|=r+r.|=r+r.|=r+r.|=r+r|}
\hline
Image& \multicolumn{17}{c|}{\shortstack{MSER, 6 tilt synths x (1+ 2 scale synth).  $\Delta\phi=360\degree{}/t$, \\ t = $\{1;5;9\}$. Total image area $A_{total}=2.8A_{orig}$}} \\
\hline
& \multicolumn{3}{c|}{Time}
& \multicolumn{6}{c|}{LO-RANSAC}
& \multicolumn{6}{c|}{Tentatives quality }
& \multicolumn{2}{c|}{Regions}\\
\hline
&
&&& \multicolumn{3}{c|}{SIFT}
& \multicolumn{3}{c|}{RootSIFT}
& \multicolumn{3}{c|}{SIFT}
& \multicolumn{3}{c|}{RootSIFT}
& &\\
&\begin{sideways}1 core $[s]$\end{sideways}
&\begin{sideways}2 cores $[s]$\end{sideways}
&\begin{sideways}4 cores $[s]$\end{sideways}
&\begin{sideways}Correct inliers\end{sideways} 
&\begin{sideways} Inliers\end{sideways}
&\begin{sideways}Correct / all $[\%]$\end{sideways} 
&\begin{sideways}Correct inliers\end{sideways} 
&\begin{sideways} Inliers\end{sideways}
&\begin{sideways}Correct / all $[\%]$\end{sideways} 
&\begin{sideways}Correct matches\end{sideways} 
&\begin{sideways}Tentatives\end{sideways}
&\begin{sideways}Correct  / all $[\%]$\end{sideways}
&\begin{sideways}Correct matches\end{sideways} 
&\begin{sideways}Tentatives\end{sideways}
&\begin{sideways}Correct / all $[\%]$\end{sideways}
&\begin{sideways}Image 1\end{sideways}
&\begin{sideways}Image 2\end{sideways}\\
\hline
 graf&3&1.6&1.1&167&173&96.5&165&169&97.6&175&340&51.5&170&339&50.1&2780&3782\\ 
index&2.2&1.2&0.8&23&32&71.9&24&35&68.6&25&106&23.6&27&103&26.2&1204&736\\ 
shop&2.5&1.4&0.9&67&69&97.1&73&74&98.6&67&172&39&73&163&44.8&2899&1474\\ 
adam&0.7&0.4&0.3&18&20&90&18&21&85.7&20&48&41.7&20&42&47.6&357&164\\
\hline
\end{tabular}
\end{minipage}
\end{table*}
\begin{table*}[p]
\centering
\begin{minipage}[h]{0.85\linewidth}
\caption{Performance on the EVD dataset.Hessian-Affine,  {\sc Sparse} configuration. Results with less than 8 correct inliers are in red.}
\label{tabular:test-optimal-fast-hess}
\footnotesize
\centering
\begin{tabular}{|=c|...|=r+r.|=r+r.|=r+r.|=r+r.|=r+r|}
\hline
Image 
& \multicolumn{17}{c|}{\shortstack{HessAff, 10 synths. $\Delta\phi =360\degree /t$,\\t =$\{1;\sqrt{2};2;2\sqrt{2};4;4\sqrt{2};8\}$. Total image area $A_{total}=4A_{orig}$}} \\
\hline
& \multicolumn{3}{c|}{Time}
& \multicolumn{6}{c|}{LO-RANSAC}
& \multicolumn{6}{c|}{Tentatives quality }
& \multicolumn{2}{c|}{Regions}\\
\hline
&
&&& \multicolumn{3}{c|}{SIFT}
& \multicolumn{3}{c|}{RootSIFT}
& \multicolumn{3}{c|}{SIFT}
& \multicolumn{3}{c|}{RootSIFT}
& &\\
&\begin{sideways}1 core $[s]$\end{sideways}
&\begin{sideways}2 cores $[s]$\end{sideways}
&\begin{sideways}4 cores $[s]$\end{sideways}
&\begin{sideways}Correct inliers\end{sideways} 
&\begin{sideways} Inliers\end{sideways}
&\begin{sideways}Correct / all $[\%]$\end{sideways} 
&\begin{sideways}Correct inliers\end{sideways} 
&\begin{sideways} Inliers\end{sideways}
&\begin{sideways}Correct / all $[\%]$\end{sideways} 
&\begin{sideways}Correct matches\end{sideways} 
&\begin{sideways}Tentatives\end{sideways}
&\begin{sideways}Correct  / all $[\%]$\end{sideways}
&\begin{sideways}Correct matches\end{sideways} 
&\begin{sideways}Tentatives\end{sideways}
&\begin{sideways}Correct / all $[\%]$\end{sideways}
&\begin{sideways}Image 1\end{sideways}
&\begin{sideways}Image 2\end{sideways}\\
\hline
 graf&11&5.9&3.9&371&379&97.9&375&383&97.9&375&799&46.9&384&765&50.2&12519&14144\\ 
index&5.4&3&2&23&40&57.5&34&52&65.4&34&412&8.3&44&284&15.5&7967&3139\\ 
shop&10.1&5.4&3.9&134&143&93.7&133&141&94.3&138&341&40.5&135&257&52.5&16326&8549\\ 
adam&1.6&0.8&0.6&86&93&92.5&86&99&86.9&88&157&56.1&88&151&58.3&2486&635\\
\hline
\end{tabular}
\end{minipage}
\end{table*}
\begin{table*}[p]
\centering
\begin{minipage}[h]{0.85\linewidth}
\caption{Performance on the EVD dataset.DoG, {\sc Sparse} configuration. Results with less than 8 correct inliers are in red.}
\label{tabular:test-optimal-fast-sift}
\footnotesize
\centering
\begin{tabular}{|=c|...|=r+r.|=r+r.|=r+r.|=r+r.|=r+r|}
\hline
Image 
& \multicolumn{17}{c|}{\shortstack{DoG, 30 synths.  $\Delta\phi=120\degree{}/t$,  \\t = $\{1;2;4;6;8\}$,  Total image area $A_{total}=7A_{orig}$}}  \\
\hline
& \multicolumn{3}{c|}{Time}
& \multicolumn{6}{c|}{LO-RANSAC}
& \multicolumn{6}{c|}{Tentatives quality }
& \multicolumn{2}{c|}{Regions}\\
\hline
&
&&& \multicolumn{3}{c|}{SIFT}
& \multicolumn{3}{c|}{RootSIFT}
& \multicolumn{3}{c|}{SIFT}
& \multicolumn{3}{c|}{RootSIFT}
& &\\
&\begin{sideways}1 core $[s]$\end{sideways}
&\begin{sideways}2 cores $[s]$\end{sideways}
&\begin{sideways}4 cores $[s]$\end{sideways}
&\begin{sideways}Correct inliers\end{sideways} 
&\begin{sideways} Inliers\end{sideways}
&\begin{sideways}Correct / all $[\%]$\end{sideways} 
&\begin{sideways}Correct inliers\end{sideways} 
&\begin{sideways} Inliers\end{sideways}
&\begin{sideways}Correct / all $[\%]$\end{sideways} 
&\begin{sideways}Correct matches\end{sideways} 
&\begin{sideways}Tentatives\end{sideways}
&\begin{sideways}Correct  / all $[\%]$\end{sideways}
&\begin{sideways}Correct matches\end{sideways} 
&\begin{sideways}Tentatives\end{sideways}
&\begin{sideways}Correct / all $[\%]$\end{sideways}
&\begin{sideways}Image 1\end{sideways}
&\begin{sideways}Image 2\end{sideways}\\
\hline
 graf&11.2&6&3.7&354&368&96.2&363&376&96.5&368&1121&32.8&379&1019&37.2&12708&17175\\ 
index&8.5&4.8&3.1&52&67&77.6&51&73&69.9&63&930&6.8&67&692&9.7&7986&4770\\ 
shop&10.3&5.5&3.3&62&65&95.4&66&72&91.7&63&575&11&69&374&18.4&15747&7873\\ 
adam&2.8&1.5&0.9&70&72&97.2&69&79&87.3&74&283&26.1&75&241&31.1&3181&899\\
\hline
\end{tabular}
\end{minipage}
\end{table*}
\begin{table*}[p]
\centering
\begin{minipage}[h]{0.85\linewidth}
\caption{Performance on the EVD dataset. MSER,{\sc Dense} configuration. Results with less than 8 correct inliers are in red.}
\label{tabular:test-optimal-rel-mser}
\footnotesize
\centering
\begin{tabular}{|=c|...|=r+r.|=r+r.|=r+r.|=r+r.|=r+r|}
\hline
Image 
& \multicolumn{17}{c|}{\shortstack{MSER, 14 tilt synths x(1+ 2 scale synth).  $\Delta\phi=180\degree{}/t$,\\ t = $\{1;5;9\}$. Total image area  $A_{total}=4.2A_{orig}$}} \\
\hline
& \multicolumn{3}{c|}{Time}
& \multicolumn{6}{c|}{LO-RANSAC}
& \multicolumn{6}{c|}{Tentatives quality }
& \multicolumn{2}{c|}{Regions}\\
\hline
&
&&& \multicolumn{3}{c|}{SIFT}
& \multicolumn{3}{c|}{RootSIFT}
& \multicolumn{3}{c|}{SIFT}
& \multicolumn{3}{c|}{RootSIFT}
& &\\
&\begin{sideways}1 core $[s]$\end{sideways}
&\begin{sideways}2 cores $[s]$\end{sideways}
&\begin{sideways}4 cores $[s]$\end{sideways}
&\begin{sideways}Correct inliers\end{sideways} 
&\begin{sideways} Inliers\end{sideways}
&\begin{sideways}Correct / all $[\%]$\end{sideways} 
&\begin{sideways}Correct inliers\end{sideways} 
&\begin{sideways} Inliers\end{sideways}
&\begin{sideways}Correct / all $[\%]$\end{sideways} 
&\begin{sideways}Correct matches\end{sideways} 
&\begin{sideways}Tentatives\end{sideways}
&\begin{sideways}Correct  / all $[\%]$\end{sideways}
&\begin{sideways}Correct matches\end{sideways} 
&\begin{sideways}Tentatives\end{sideways}
&\begin{sideways}Correct / all $[\%]$\end{sideways}
&\begin{sideways}Image 1\end{sideways}
&\begin{sideways}Image 2\end{sideways}\\
\hline
 graf&22.2&11.7&6.9&330&361&91.4&332&353&94.1&339&1061&32&342&1014&33.7&20454&26177\\ 
index&16.6&8.8&5.1&51&65&78.5&56&76&73.7&60&390&15.4&66&391&16.9&9087&5733\\ 
shop&18.5&9.7&5.7&139&152&91.4&147&161&91.3&145&530&27.4&150&446&33.6&19560&9905\\ 
adam&5.4&2.8&1.7&36&41&87.8&26&38&68.4&38&114&33.3&34&104&32.7&2161&1201\\
\hline
\end{tabular}
\end{minipage}
\end{table*}
\begin{table*}[p]
\centering
\begin{minipage}[h]{0.9\linewidth}
\caption{Performance on the EVD dataset.  Hessian-Affine, {\sc Dense} configuration. Results with less than 8 correct inliers are in red.}
\label{tabular:test-optimal-rel-hess}
\footnotesize 
\centering
\begin{tabular}{|=c|...|=r+r.|=r+r.|=r+r.|=r+r.|=r+r|}
\hline
Image 
& \multicolumn{17}{c|}{\shortstack{HessAff, 50 synths. $\Delta\phi =72\degree /t$,\\t = $\{1;2;4;6;8\}$. Total image area  $A_{total}=11A_{orig}$}} \\
\hline
& \multicolumn{3}{c|}{Time}
& \multicolumn{6}{c|}{LO-RANSAC}
& \multicolumn{6}{c|}{Tentatives quality }
& \multicolumn{2}{c|}{Regions}\\
\hline
&
&&& \multicolumn{3}{c|}{SIFT}
& \multicolumn{3}{c|}{RootSIFT}
& \multicolumn{3}{c|}{SIFT}
& \multicolumn{3}{c|}{RootSIFT}
& &\\
&\begin{sideways}1 core $[s]$\end{sideways}
&\begin{sideways}2 cores $[s]$\end{sideways}
&\begin{sideways}4 cores $[s]$\end{sideways}
&\begin{sideways}Correct inliers\end{sideways} 
&\begin{sideways} Inliers\end{sideways}
&\begin{sideways}Correct / all $[\%]$\end{sideways} 
&\begin{sideways}Correct inliers\end{sideways} 
&\begin{sideways} Inliers\end{sideways}
&\begin{sideways}Correct / all $[\%]$\end{sideways} 
&\begin{sideways}Correct matches\end{sideways} 
&\begin{sideways}Tentatives\end{sideways}
&\begin{sideways}Correct  / all $[\%]$\end{sideways}
&\begin{sideways}Correct matches\end{sideways} 
&\begin{sideways}Tentatives\end{sideways}
&\begin{sideways}Correct / all $[\%]$\end{sideways}
&\begin{sideways}Image 1\end{sideways}
&\begin{sideways}Image 2\end{sideways}\\
\hline
 graf&45.2&23.8&14.2&1214&1260&96.3&1235&1274&96.9&1249&2958&42.2&1267&2887&43.9&46997&58020\\ 
index&20.8&10.9&6.5&270&297&90.9&264&302&87.4&312&1698&18.4&326&1297&25.1&27910&11497\\ 
shop&36.2&19.2&11.3&303&315&96.2&311&322&96.6&311&847&36.7&326&676&48.2&55508&25538\\ 
adam&6&3.1&1.9&205&239&85.8&214&231&92.6&242&525&46.1&239&497&48.1&7616&2310\\
\hline
\end{tabular}
\end{minipage}
\end{table*}
\begin{table*}[p]
\centering
\begin{minipage}[h]{0.9\linewidth}
\caption{Performance on the EVD dataset. DoG,{\sc Dense} configuration. Results with less than 8 correct inliers are in red.}
\label{tabular:test-optimal-rel-sift}
\footnotesize
\centering
\begin{tabular}{|=c|...|=r+r.|=r+r.|=r+r.|=r+r.|=r+r|}
\hline
Image 
& \multicolumn{17}{c|}{\shortstack{DoG, 59 synths.  $\Delta\phi=72\degree{}/t$,  \\t =$\{1;\sqrt{2};2;2\sqrt{2};4;4\sqrt{2};8\}$. Total image area $A_{total}=16A_{orig}$}}  \\
\hline
& \multicolumn{3}{c|}{Time}
& \multicolumn{6}{c|}{LO-RANSAC}
& \multicolumn{6}{c|}{Tentatives quality }
& \multicolumn{2}{c|}{Regions}\\
\hline
&
&&& \multicolumn{3}{c|}{SIFT}
& \multicolumn{3}{c|}{RootSIFT}
& \multicolumn{3}{c|}{SIFT}
& \multicolumn{3}{c|}{RootSIFT}
& &\\
&\begin{sideways}1 core $[s]$\end{sideways}
&\begin{sideways}2 cores $[s]$\end{sideways}
&\begin{sideways}4 cores $[s]$\end{sideways}
&\begin{sideways}Correct inliers\end{sideways} 
&\begin{sideways} Inliers\end{sideways}
&\begin{sideways}Correct / all $[\%]$\end{sideways} 
&\begin{sideways}Correct inliers\end{sideways} 
&\begin{sideways} Inliers\end{sideways}
&\begin{sideways}Correct / all $[\%]$\end{sideways} 
&\begin{sideways}Correct matches\end{sideways} 
&\begin{sideways}Tentatives\end{sideways}
&\begin{sideways}Correct  / all $[\%]$\end{sideways}
&\begin{sideways}Correct matches\end{sideways} 
&\begin{sideways}Tentatives\end{sideways}
&\begin{sideways}Correct / all $[\%]$\end{sideways}
&\begin{sideways}Image 1\end{sideways}
&\begin{sideways}Image 2\end{sideways}\\
\hline
 graf&25.5&13.9&8.1&637&659&96.7&653&675&96.7&665&2284&29.1&675&2041&33.1&28836&38361\\ 
index&18.3&9.9&6.1&138&167&82.6&143&164&87.2&161&1901&8.5&157&1385&11.3&18477&10549\\ 
shop&24&12.8&7.7&118&122&96.7&130&134&97&124&1238&10&132&845&15.6&37205&18541\\ 
adam&6.3&3.3&1.9&129&135&95.6&125&138&90.6&136&549&24.8&145&434&33.4&7462&2062\\
\hline
\end{tabular}
\end{minipage}
\end{table*}
\begin{table*}[p]
\centering
\begin{minipage}[h]{0.6\linewidth}
\caption{Performance on the EVD dataset.ASIFT. Results with less than 8 correct inliers are in red.}
\label{tabular:test-asift}
\footnotesize
\centering
\begin{tabular}{|=l|...|=r+r.|=r+r.|=r+r|}
\hline
Image 
& \multicolumn{11}{c|}{\shortstack{ASIFT, 59 synths.  $\Delta\phi=72\degree{}/t$, \\t =$\{1;\sqrt{2};2;2\sqrt{2};4;4\sqrt{2};8\}$. Total image area  $A_{total}=16A_{orig}$}} \\
\hline
& \multicolumn{3}{c|}{Time}
& \multicolumn{3}{c|}{ORSA}
& \multicolumn{3}{c|}{Tentatives quality }
& \multicolumn{2}{c|}{Regions}\\
\hline
&&
&& \multicolumn{3}{c|}{SIFT}
& \multicolumn{3}{c|}{SIFT}
& &\\
&\begin{sideways}1 core $[s]$\end{sideways}
&\begin{sideways}2 cores $[s]$\end{sideways}
&\begin{sideways}4 cores $[s]$\end{sideways}
&\begin{sideways}Correct inliers\end{sideways} 
&\begin{sideways} Inliers\end{sideways}
&\begin{sideways}Correct / all $[\%]$\end{sideways} 
&\begin{sideways}Correct matches\end{sideways} 
&\begin{sideways}Tentatives\end{sideways}
&\begin{sideways}Correct / all $[\%]$\end{sideways}
&\begin{sideways}Image 1\end{sideways}
&\begin{sideways}Image 2\end{sideways}\\
\hline
 graf&81.8&26.5&14.8&322&531&60.6&325&582&55.8&31199&38677\\ 
index&54.1&18.3&10.9&23&94&24.5&23&178&12.9&20349&10115\\ 
shop&79.5&25&14.1&17&34&50&18&76&23.6&41984&25270\\ 
adam&17.8&6&4.3&24&63&38.1&25&92&27.1&7572&3295\\
\hline
\end{tabular}
\end{minipage}
\end{table*}
\begin{table*}[b]
\centering
\begin{minipage}[h]{0.65\linewidth}
\caption{Performance on the EVD dataset. MODS ($\theta_m$ = 15), SIFT. Results with less than 8 correct inliers are in red.}
\label{tabular:test-MODS}
\footnotesize
\centering
\begin{tabular}{|=l|=c...|=r+r.|=r+r.|=r+r|}
\hline
Image 
& \multicolumn{12}{c|}{\shortstack{MODS, 4 steps. \\1. MSER Scale only. 2. MSER {\sc Sparse}. \\3. HessAff {\sc Sparse}. 4. HessAff  {\sc Dense} }} \\
\hline
& \multicolumn{4}{c|}{Time}
& \multicolumn{3}{c|}{LO-RANSAC}
& \multicolumn{3}{c|}{Tentatives quality }
& \multicolumn{2}{c|}{Regions}\\
\hline
&&&&& \multicolumn{3}{c|}{SIFT}
& \multicolumn{3}{c|}{SIFT}
& &\\
&\begin{sideways}Step\end{sideways}
&\begin{sideways}1 core $[s]$\end{sideways}
&\begin{sideways}2 cores $[s]$\end{sideways}
&\begin{sideways}4 cores $[s]$\end{sideways}
&\begin{sideways}Correct inliers\end{sideways} 
&\begin{sideways} Inliers\end{sideways}
&\begin{sideways}Correct / all $[\%]$\end{sideways} 
&\begin{sideways}Correct matches\end{sideways} 
&\begin{sideways}Tentatives\end{sideways}
&\begin{sideways}Correct  / all $[\%]$\end{sideways}
&\begin{sideways}Image 1\end{sideways}
&\begin{sideways}Image 2\end{sideways}\\
\hline
 graf&1&1&0.8&0.8&81&85&95.3&82&160&51.2&1018&1674\\ 
index&1&0.5&0.4&0.4&19&20&95&19&56&33.9&411&246\\ 
shop&1&0.8&0.7&0.7&28&30&93.3&28&84&33.3&1321&711\\ 
adam&2&0.8&0.5&0.3&19&22&86.4&21&48&43.8&357&164\\ 
there&2&4.5&2.8&2&10&18&55.6&15&66&22.7&571&2833\\ 
mag&2&0.8&0.5&0.4&30&31&96.8&30&52&57.7&393&509\\ 
dum&3&29.4&18.9&15.5&24&29&82.8&30&1136&2.6&33342&23711\\ 
grand&3&21.9&13.7&11.3&17&25&68&21&754&2.8&24731&20297\\ 
fox&2&2.1&1.4&1.1&16&19&84.2&19&76&25&1717&1011\\ 
cafe&2&1.8&1.2&0.9&18&20&90&18&142&12.7&1402&1319\\ 
girl&3&13.1&7.4&5.3&35&46&76.1&38&549&6.9&10460&16105\\ 
pkk&2&2.5&1.6&1.3&\rowstyle{\color{red}}7&\rowstyle{\color{red}}15&\rowstyle{\color{red}}46.7&10&81&12.3&1229&1267\\ 
cat&3&3.9&2&1.4&35&38&92.1&35&143&24.5&1262&3279\\ 
face&2&3.6&2.4&2&9&15&60&11&118&9.3&3411&2371\\ 
vin&4&30.3&17.8&12.5&18&38&47.4&22&657&3.4&18956&31984\\ 

\hline
\end{tabular}
\end{minipage}
\end{table*}
\begin{table*}[p]
\centering
\begin{minipage}[h]{0.6\linewidth}
\centering
\caption{Performance on the EVD dataset.MODS ($\theta_m$ = 15), RootSIFT. Results with less than 8 correct inliers are in red.}
\label{tabular:test-MODS-rootsift}
\footnotesize
\centering
\begin{tabular}{|=l|=c...|=r+r.|=r+r.|=r+r|}
\hline
Image 
& \multicolumn{12}{c|}{\shortstack{MODS, 4 steps. \\1. MSER Scale only. 2. MSER {\sc Sparse}.\\ 3. HessAff {\sc Sparse}. 4. HessAff  {\sc Dense} }} \\
\hline
& \multicolumn{4}{c|}{Time}
& \multicolumn{3}{c|}{LO-RANSAC}
& \multicolumn{3}{c|}{Tentatives quality }
& \multicolumn{2}{c|}{Regions}\\
\hline
&&&&& \multicolumn{3}{c|}{RootSIFT}
& \multicolumn{3}{c|}{RootSIFT}
& &\\
&\begin{sideways}Step\end{sideways}
&\begin{sideways}1 core $[s]$\end{sideways}
&\begin{sideways}2 cores $[s]$\end{sideways}
&\begin{sideways}4 cores $[s]$\end{sideways}
&\begin{sideways}Correct inliers\end{sideways} 
&\begin{sideways} Inliers\end{sideways}
&\begin{sideways}Correct / all $[\%]$\end{sideways} 
&\begin{sideways}Correct matches\end{sideways} 
&\begin{sideways}Tentatives\end{sideways}
&\begin{sideways}Correct  / all $[\%]$\end{sideways}
&\begin{sideways}Image 1\end{sideways}
&\begin{sideways}Image 2\end{sideways}\\
\hline
 graf&1&1&0.8&0.8&82&87&94.3&83&154&53.9&1018&1674\\ 
index&1&0.5&0.4&0.4&18&20&90&18&42&42.9&411&246\\ 
shop&1&0.8&0.7&0.7&29&31&93.5&29&61&47.5&1321&711\\ 
adam&2&0.8&0.5&0.4&20&23&87&22&47&46.8&357&164\\ 
there&2&4.5&2.8&2&14&17&82.4&16&60&26.7&571&2833\\ 
mag&2&0.9&0.5&0.4&31&31&100&31&44&70.5&393&509\\ 
dum&3&27.2&16.9&13.5&25&32&78.1&29&850&3.4&33342&23711\\ 
grand&3&20.9&12.5&10&14&24&58.3&19&468&4.1&24731&20297\\ 
fox&2&2.1&1.4&1.1&19&20&95&20&62&32.3&1717&1011\\ 
cafe&2&1.8&1.2&0.9&17&21&81&18&117&15.4&1402&1319\\ 
girl&3&13.1&7.3&5.2&34&44&77.3&38&436&8.7&10460&16105\\ 
pkk&3&9.5&5.3&4&27&37&73&33&344&9.6&10686&7085\\ 
cat&3&3.6&2.1&1.5&25&34&73.5&30&149&20.1&1262&3279\\ 
face&3&15.6&8.9&7.1&39&44&88.6&42&534&7.9&18857&13271\\ 
vin&4&29.7&17.1&11.8&19&32&59.4&21&455&4.6&18956&31984\\ 

\hline
\end{tabular}
\end{minipage}
\end{table*}
\clearpage


\end{document}